%% file: aaai2026.tex
\documentclass[letterpaper]{article} % DO NOT CHANGE THIS
\usepackage{aaai2026}  % DO NOT CHANGE THIS
\usepackage{times}  % DO NOT CHANGE THIS
\usepackage{helvet}  % DO NOT CHANGE THIS
\usepackage{courier}  % DO NOT CHANGE THIS
\usepackage[hyphens]{url}  % DO NOT CHANGE THIS
\usepackage{graphicx} % DO NOT CHANGE THIS
\usepackage{natbib}  % DO NOT CHANGE THIS AND DO NOT ADD ANY OPTIONS TO IT
\usepackage{caption} % DO NOT CHANGE THIS AND DO NOT ADD ANY OPTIONS TO IT
\usepackage{xcolor}

\usepackage{algorithm}
\usepackage{graphics} % for pdf, bitmapped graphics files
\usepackage{epsfig} % for postscript graphics files
\usepackage{mathptmx} % assumes new font selection scheme installed
\usepackage{amsmath} % assumes amsmath package installed
\usepackage{amssymb}  % assumes amsmath package installed
\usepackage{microtype}
\usepackage{graphicx}
\usepackage{booktabs} % for professional tables
\usepackage{algorithm}
\usepackage{amsthm}
\usepackage{amsfonts}
\usepackage[noend]{algpseudocode}
\usepackage{array}
\usepackage{adjustbox}
\usepackage{subfig}

\usepackage{newfloat}
\usepackage{listings}
\DeclareCaptionStyle{ruled}{labelfont=normalfont,labelsep=colon,strut=off} % DO NOT CHANGE THIS
\floatstyle{ruled}
\newfloat{listing}{tb}{lst}{}
\floatname{listing}{Listing}
\title{CRISP: Curriculum-Inducing Primitive Informed Subgoal Prediction \\ for Boosting Hierarchical Reinforcement Learning}
\author {
    Utsav Singh\textsuperscript{\rm 1},
    Vinay P. Namboodiri\textsuperscript{\rm 2}
}
\affiliations {
    \textsuperscript{\rm 1}Department of Computer Science and Engineering, IIT Kanpur, India\\
    \textsuperscript{\rm 2}Department of Computer Science, University of Bath, UK\\
    utsavz@iitk.ac.in, vpn22@bath.ac.uk
}
\usepackage{bibentry}
\begin{document}

\maketitle

\input{sections/abstract}
\input{sections/intro}
\input{sections/related_work}

\input{sections/background}
\input{sections/method}
\input{sections/experiments}
\input{sections/discussion}

\bibliography{aaai2026}

\newpage
\appendix
\input{sections/appendix}
\end{document}

%% file: sections/abstract.tex
\begin{abstract}
Hierarchical reinforcement learning (HRL) leverages temporal abstraction to efficiently tackle complex long-horizon tasks. However, HRL often collapses because the low-level primitive’s continual updates make earlier sub-goals issued by the high-level policy obsolete, introducing non-stationarity that destabilizes training. We propose CRISP, a curriculum-driven framework that tackles this instability with three key ingredients: (1) primitive-informed parsing (PIP), which adaptively re-labels a handful of expert demonstrations to always generate reachable subgoals by the current low-level primitive; (2) an inverse-reinforcement-learning regularizer that steers the high-level policy toward the expert-induced subgoal distribution and stabilizes learning; and (3) a unified training loop that leverages these components to boost sample efficiency. Across six sparse-reward robotic navigation and manipulation benchmarks, CRISP improves success rates by more than 40\% over strong hierarchical and flat baselines and successfully transfers to real-world tasks, demonstrating the promise of curriculum-based HRL for practical scenarios.
\end{abstract}

%% file: sections/intro.tex
\section{Introduction}
\label{sec:introduction}

While reinforcement learning (RL) has demonstrated remarkable successes in continuous control domains such as robotic manipulation~\cite{DBLP:journals/corr/LevineFDA15, DBLP:journals/corr/VecerikHSWPPHRL17}, tackling long-horizon continuous control tasks characterized by sparse rewards, inefficient exploration, and difficult credit assignment~\cite{nachum2019does, DBLP:journals/corr/KulkarniNST16, Andrychowicz2017HindsightER} remains a significant hurdle. Hierarchical reinforcement learning (HRL) offers a promising paradigm to address these challenges by decomposing tasks into subtasks via temporal abstraction, enabling efficient exploration~\cite{Dayan:1992:FRL:645753.668239, SUTTON1999181, NIPS1997_5ca3e9b1, nachum2019does}. Several goal-conditioned HRL frameworks employ a high-level policy that proposes subgoals and a lower-level primitive that executes actions to achieve those subgoals~\cite{DBLP:journals/corr/abs-1805-08296, DBLP:journals/corr/VezhnevetsOSHJS17, DBLP:journals/corr/abs-1712-00948}. 

\textbf{HRL Challenges.} Despite these benefits, HRL approaches suffer from the issue of non-stationarity. Specifically, as the higher-level policy predicts subgoals, the lower-level primitive behavior evolves during training, causing the higher-level state transition dynamics and reward functions to shift over time. This causes the higher-level policy to continually adjust to a moving target, thus destabilizing learning and slowing convergence. Additionally, the higher-level policy may generate subgoals that are currently unachievable by the lower primitive, further impeding effective learning. Therefore, the higher-level policy should consistently predict subgoals that are achievable given the current lower-level primitive.

To address these challenges, we introduce a novel framework that seamlessly integrates reinforcement learning (RL) and imitation learning (IL) to mitigate HRL’s issues of non-stationarity and infeasible subgoal generation. At the core of our approach is \emph{Primitive-Informed Parsing} (PIP), a mechanism that periodically segments expert demonstration trajectories to construct a subgoal transition dataset tailored for the higher-level policy. By continually updating this dataset, PIP adaptively identifies \emph{achievable} subgoals that align with the evolving capabilities of the current lower-level primitive.

Leveraging this adaptive subgoal transition dataset, we employ an inverse reinforcement learning (IRL) objective to regularize the higher-level policy, ensuring it consistently generates subgoals that are achievable by the current lower-level primitive. By grounding subgoal generation in the evolving capabilities of the lower primitive, this approach naturally induces a subgoal curriculum that mitigates non-stationarity in hierarchical reinforcement learning, thereby stabilizing training and improving overall performance.

CRISP efficiently integrates RL and IL by jointly optimizing the RL objective (which promotes efficient autonomous exploration) with an IRL regularization that mitigates non-stationarity and stabilizes hierarchical training. This unified framework enables the agent to explore effectively while leveraging a curriculum of achievable subgoals, leading to more stable learning and enhanced sample efficiency.

\input{figures_tex/main_figure}
We evaluate CRISP on six challenging simulated domains: maze navigation, pick-and-place, bin, hollow, rope manipulation, and the franka-kitchen suite, demonstrating that our approach achieves over 40\% higher success rates than strong hierarchical and flat baselines, while consistently delivering superior sample efficiency and stable hierarchical learning. Further, we perform experiments on real-world pick-and-place, bin, and rope-manipulation tasks, where CRISP is able to significantly outperform the baselines. Figure~\ref{fig:explain_method} provides an overview of the PIP approach for segmenting expert demonstrations.
\\
\\
\noindent Our main contributions are as follows:
\begin{itemize}
  \item \textbf{Adaptive subgoal dataset generation via PIP.} CRISP employs PIP to periodically generate a dataset of achievable subgoals for the lower primitive.
  \item \textbf{Curriculum via IRL regularization.} CRISP employs an IRL based objective to guide the high-level policy to predict a curriculum of achievable subgoals, thereby mitigating non-stationarity in HRL.
  \item \textbf{Extensive simulated benchmarks.} Across six sparse-reward navigation and manipulation tasks, CRISP achieves more than 40\% higher success rates and faster convergence than prior methods.
  \item \textbf{Real-robot validation.} The policies trained with CRISP transfer directly to physical pick-and-place, bin, and rope-manipulation scenarios without additional fine-tuning, outperforming competing approaches.
\end{itemize}

%% file: figures_tex/main_figure.tex
\begin{figure*}[t]
\vspace{5pt}
\centering
% \captionsetup{font=footnotesize}
\includegraphics[scale=0.32]{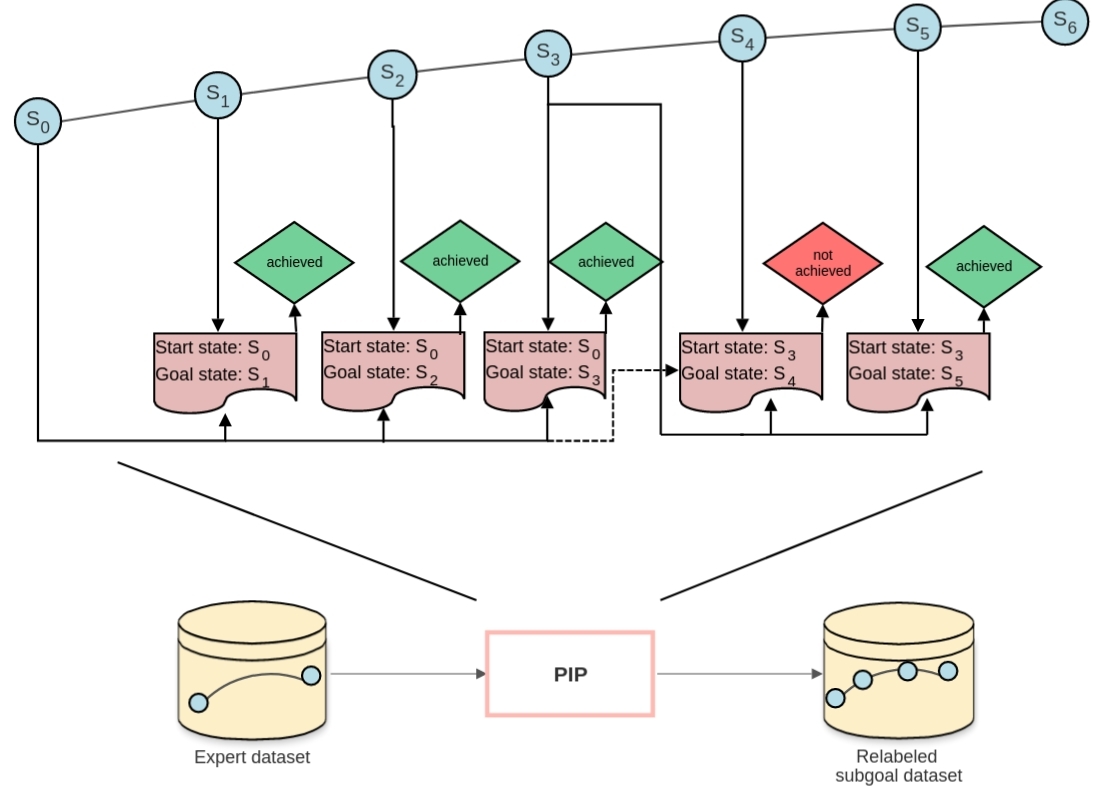}
\caption{\textbf{PIP Overview}: PIP segments expert demonstrations by consecutively passing demonstration states as subgoals ($s_i$: for $i=1$ to $6$). If the lower primitive is unable to achieve state $i$ in $c$ timesteps (here $i=4$th timestep), $s_{i-1}$ (here $s_3$) is selected as subgoal for initial state $s_0$, since $s_{i-1}$ (here $s_{3}$) was the last reachable subgoal. The process is repeated with $s_{i-1}$ (here $s_3$) as next initial state.}
\label{fig:explain_method}
\end{figure*}

%% file: sections/related_work.tex
\section{Related Work}
\label{sec:related_work}

\subsection{Hierarchical Reinforcement Learning.}
Learning effective policy hierarchies is a long-standing goal in RL~\cite{Barto03recentadvances,SUTTON1999181,NIPS1997_5ca3e9b1,DBLP:journals/corr/cs-LG-9905014}.  
\emph{Options} frameworks extend primitive actions in time and learn termination conditions~\cite{SUTTON1999181,DBLP:journals/corr/BaconHP16,DBLP:journals/corr/abs-1709-04571,DBLP:journals/corr/abs-1711-03817,DBLP:journals/corr/abs-1902-09996,DBLP:journals/corr/abs-1712-00004}, but can collapse to degenerate solutions without additional regularization.  
Goal-conditioned control constrains exploration to specific targets~\cite{Kaelbling93learningto,fd-ssvf-02} and has been lifted to hierarchical settings through two-level architectures~\cite{DBLP:journals/corr/abs-1906-11228,DBLP:journals/corr/abs-2007-15588,NEURIPS2019_c8d3a760}.  
To curb non-stationarity arising from an evolving low-level policy, HIRO~\cite{DBLP:journals/corr/abs-1805-08296} and HAC~\cite{DBLP:journals/corr/abs-1712-00948} relabel past subgoals in the replay buffer. CRISP instead imposes an IRL regularizer that steers the high-level policy toward \emph{reachable} subgoals predicted by primitive-informed parsing (PIP), yielding greater stability.  
Reset-controller work focuses on learning policies that return the agent to safe states inside simulation~\cite{salimans2018learning,florensa2017reverse,peng2018deepmimic}; CRISP uses environment resets only to \emph{parse} expert demonstrations and is agnostic to resets during training.  
Skill-prior methods pre-train on related tasks before fine-tuning~\cite{DBLP:journals/corr/abs-2010-11944,DBLP:journals/corr/abs-2011-10024}, but often struggle under distribution shift or sub-optimal demonstrations. Hand-crafted primitive libraries~\cite{DBLP:journals/corr/abs-2110-15360,DBLP:journals/corr/abs-2110-03655} avoid learning low-level skills at the cost of extensive domain engineering; CRISP learns both hierarchy levels jointly, eliminating this manual burden.

\subsection{Learning from demonstrations} Prior works~\cite{DBLP:journals/corr/abs-1709-10089,DBLP:journals/corr/abs-1709-10087,DBLP:journals/corr/HesterVPLSPSDOA17} that leverage expert demonstrations to solve complex tasks have demonstrated impressive results. Expert demonstrations have been used to bootstrap option learning~\cite{krishnan2017ddco, fox2017multilevel, Shankar2020LearningRS, kipf2019compile}.  Other approaches use imitation learning to bootstrap hierarchical approaches in complex task domains~\cite{pmlr-v80-shiarlis18a,DBLP:journals/corr/abs-1710-05421,doi:10.1177/0278364918784350,kipf2019compile}. A particular line of work~\cite{DBLP:journals/corr/abs-1910-11956} uses fixed window based approach for parsing expert demonstrations to generate subgoal transition dataset for training higher level policy using imitation learning. However, such methods might produce sub-optimal subgoals for the lower primitive. In contrast, our relabeling approach (PIP) segments expert demonstration trajectories into \textit{meaningful} subtasks by always predicting reachable subgoals and thus balancing the task-split between hierarchical levels.

\subsection{Curriculum Learning} Our approach is inspired from curriculum learning~\cite{10.1145/1553374.1553380}, where task difficulty gradually increases in complexity, propelling the policy to achieve incrementally harder subgoals. A genetic curriculum based approach~\cite{song2022robust} identifies unsolved scenarios to automatically generate an associated curriculum via adversarial training. ACCEL~\cite{parker2022evolving} proposes a regret based curriculum approach that keeps a record of previous scenarios, and selects the ones with highest regret. Prior work generates subgoals while considering the lower primitive performance, by factoring in the task success rate~\cite{fournier2018accuracy,florensa2018automatic,racaniere2019automated}, value function~\cite{ren2019exploration,sharma2021autonomous}, achieved state density~\cite{pitis2020maximum}, and value uncertainty~\cite{kim2023variational}. CRISP differs by deriving curricula directly from expert trajectories via IRL-regularized subgoal prediction, providing an implicit difficulty schedule tied to the ability of lower primitive.

%% file: sections/background.tex
\section{Background}
\label{sec:background}

\begin{algorithm}[t]
\caption{PIP: Primitive Informed Parsing}
\label{alg:algo_pip}
%  \hspace*{\algorithmicindent} \textbf{Input:} 
%  $env \:E$, Expert Demonstrations $D$ , \\
%  \hspace*{\algorithmicindent}
%  \hspace*{\algorithmicindent}\hspace*{\algorithmicindent}Lower level policy $\pi_{l}$ , Upper level policy $\pi_{u}$\\  
%  \hspace*{\algorithmicindent} \textbf{Output:} Upper level expert demonstrations
\begin{algorithmic}
% \Procedure{Careful Parsing}{}
    \State Initialize $D_g = \{\}$
    \State (final goal) $s_f = g $ %s^e_{T-1}
    \For{each trajectory $e=(s^e_0, s^e_1,., s^e_{T-1})$ in $D$}
        \State (initial state) $s_{in} \leftarrow s^e_0$
        \State Initialize list of subgoals $D^e_g = \{\}$
        \For{i = $1$ \textbf{to} $T-1$}
            \State Reset to initial state $s_{in}$
            \State Pass $s^e_i$ as the current goal to $\pi_{L}$
            \If{$s^e_i$ is not achieved by $\pi_{L}$ in $c$ time-steps}
            \State Add $(s_{in}, s^e_{i-1}, s_f)$ to $D^e_g$
            \State $s_{in} \leftarrow s^e_{i-1}$
            \EndIf
        \EndFor
        \State $D_g \leftarrow D_g \cup D^e_g$
    \EndFor
\end{algorithmic}
\end{algorithm}

We consider \textit{Universal Markov Decision Process} (UMDP) \cite{pmlr-v37-schaul15} setting, where Markov Decision processes (MDP) are augmented with the goal space $G$. UMDPs are represented as a 6-tuple $(S,A,P,R,\gamma,G)$, where $S$ is the state space, $A$ is the action space, $P(s^{'}|s,a)=\mathbb{P}(s_{t+1}=s^{'}|s_{t}=s,a_{t}=a)$ is the transition function that describes the probability of reaching state $s^{'}$, when the agent takes action $a$ in the current state $s$. The reward function $R$ generates rewards $r$ at every timestep. $\gamma$ is the discount factor, and $G$ is the goal space. In the UMDP setting, a fixed goal $g$ is selected for an episode, and $\pi(a|s,g)$ denotes the goal-conditioned policy. The discounted future state distribution is represented as $d^{\pi}(s)=(1-\gamma)\sum_{t=0}^{T}\gamma^{t}P(s_t=s|\pi)$, and the c-step future state distribution for policy $\pi$ is represented as $d^{\pi}_{c}(s)=(1-\gamma^c)\sum_{t=0}^{T}\gamma^{tc}P(s_{tc}=s|\pi)$. The overall objective is to learn policy $\pi(a|s,g)$ which maximizes the expected future discounted reward objective $ J = (1-\gamma)^{-1}\mathbb{E}_{s \sim d^{\pi}, a \sim \pi(a|s,g), g \sim G}\left[r(s_t,a_t,g)\right]$

\subsection{Problem Formulation.} Let $s$ be the current state and $g$ be the final goal for the current episode. In our goal-conditioned hierarchical RL setup, the overall policy $\pi$ is divided into multi-level policies. The higher level policy $\pi^{H}(s_g|s, g)$ predicts subgoals~\cite{Dayan:1992:FRL:645753.668239} $s_g$ for the lower level primitive $\pi^{L}(a | s, s_g)$, which in turn executes primitive actions $a$ directly on the environment. The lower primitive $\pi^{L}$ tries to achieve subgoal $s_g$ within $c$ timesteps, by maximizing intrinsic rewards $r_{in}$ provided by the higher level policy. The higher level policy $\pi^{H}$ gets extrinsic reward $r_{ex}$ from the environment, and predicts the next subgoal $s_g$ for the lower primitive. The process is continued until either the final goal $g$ is achieved, or the episode terminates. We consider sparse reward setting where the lower primitive is sparsely rewarded intrinsic reward $0$ if the agent reaches within $\delta^{L}$ distance of the predicted subgoal $s_g$ and $-1$ otherwise: $r_{in}=-1(\|s_t-s_g\|_2 > \delta^{L})$, and the higher level policy is sparsely rewarded extrinsic reward $0$ if the achieved goal is within $\delta^{H}$ distance of the final goal $g$, and $-1$ otherwise: $r_{ex}=-1(\|s_t-g\|_2 > \delta^{H})$. The expert demonstrations are represented as $D=\{e^i\}_{i=1}^N$, where $e^i=(s^e_0, s^e_1, \ldots, s^e_{T-1})$. 
% We only assume access to demonstration states $s^e_i$ (and not demonstration actions) which is a reasonable assumption in most robotic control tasks.

% \begin{algorithm}[tb]
% \caption{CRISP}\label{alg:algo_crisp}
% % \centering
% \begin{algorithmic}
% % \Procedure{Careful Parsing}{}
%     \Require $D$ (expert demonstrations)
%     \State $p$ (population hyperparameter)
%     \State Initialize higher level subgoal transition dataset $D_g = \{\}$ 
%     \For{epoch $i = 1 \ldots N $}
%         \If{$i \% p==0$}
%             \State Clear $D_g$
%             \State Populate $D_g$ by relabeling $D$ using PIP
%         \EndIf
%         \For{$j$ = $1$ \textbf{to} $T-1$}
%             \State Collect off policy experience using $\pi_{H}$ and $\pi_{L}$
%         \EndFor
%         \State Update lower primitive via SAC and IRL (Eq \ref{eqn:joint_update_lower})
%         \State Sample transitions from $D_g$
%         \State Update higher policy via SAC and IRL (Eq \ref{eqn:joint_update_upper})
%     \EndFor
% \end{algorithmic}
% \end{algorithm}

\begin{algorithm}[t]
\caption{CRISP}\label{alg:algo_crisp}
% \centering
\begin{algorithmic}
% \Procedure{Careful Parsing}{}
    \Require $D$ (expert demonstrations)
    \State $p$ (population hyperparameter)
    \State Initialize higher level subgoal transition dataset $D_g = \{\}$ 
    \For{epoch $i = 1 \ldots N $}
        \If{$i \% p==0$}
            \State Clear $D_g$
            \State Populate $D_g$ by relabeling $D$ using PIP
        \EndIf
        \For{$j$ = $1$ \textbf{to} $T-1$}
            \State Collect experience using $\pi_{H}$ and $\pi_{L}$
        \EndFor
        \State Update lower policy via RL and IRL (Eq \ref{eqn:joint_update_lower})
        \State Sample transitions from $D_g$
        \State Update higher policy via RL and IRL (Eq \ref{eqn:joint_update_upper})
    \EndFor
\end{algorithmic}
\end{algorithm}
% \end{minipage}

\subsection{Limitations of existing approaches to HRL} HRL promises the advantages of temporal abstraction and improved exploration~\cite{nachum2019does}. However it suffers from non-stationarity due to unstable lower primitive behavior. This hinders applying HRL advances to complex tasks, especially in sparse reward scenarios. The primary motivation of this work is to devise a hierarchical curriculum learning based approach to mitigate non-stationarity in HRL.

%% file: sections/method.tex
\section{Methodology}
\label{sec:method}

This section comprises of overview on (i) primitive-informed parsing (PIP) that adaptively builds the subgoal dataset according to the current lower primitive, (ii) IRL regularization that conditions the higher level policy to predict achievable sugoals, and (iii) joint SAC optimisation that unifies RL and IRL updates.

\subsection{Primitive Informed Parsing (PIP)}
\label{subsec:parsing}
PIP leverages the \emph{current} low-level policy $\pi_L$ to re-segment the expert, state-only dataset $D$ into a buffer of \emph{reachable} sub-goal transitions $D_g$ (overview in Fig.~\ref{fig:explain_method}).  
\par Notably, PIP assumes ability to reset the environment to any state in $D$. Although this seems impracticable in real world robotic scenarios, this becomes feasible in our setup since we first learn good policies in simulation, and then deploy them in real world robotic scenarios. This follows the underlying assumption that with enough training in simulation, the policy becomes general enough to perform well in real world tasks. We perform extensive experiments to support this claim in Experiments section and discuss various ways to relax this assumption in Discussion section.

% \subsection{Inverse RL regularization}
% Here, we explain CRISP uses $D_g$, the subgoal transition dataset generated using PIP to learn an IRL regularizer, which regularizes the higher level policy into predicting a curriculum of achievable goals for the lower primitive. We devise IRL objective as a GAIL~\cite{DBLP:journals/corr/HoE16} like objective implemented using LSGAN~\cite{DBLP:journals/corr/MaoLXLW16}. Let $(s^e, s^e_g, s^e_{next}) \sim D_g$ be higher level subgoal transition from an expert trajectory where $s^e$ is current state, $s^e_{next}$ is next state, $g^e$ is final goal and $s^e_g$ is subgoal supervision. Let $s_g$ be the subgoal predicted by the high level policy $\pi_{\theta_H}^{H}(\cdot|s^e, g^e)$ with parameters $\theta_H$, and $\mathbb{D}_{\epsilon_{H}}^H$ be the higher level discriminator with parameters $\epsilon_{H}$. $J_D^{H}$ represents upper level IRL objective, which depends on parameters $(\theta_{H},\epsilon_{H})$. We bootstrap the learning of higher level policy by optimizing:

\subsection{Inverse-RL Regularisation}
\label{subsec:irl}

The sub-goal buffer $D_g$ produced by PIP serves as expert data for an IRL regulariser that nudges the
\emph{high-level} policy to propose \emph{reachable} goals for the
lower primitive.  We follow the GAIL framework
\cite{DBLP:journals/corr/HoE16} with the stabilising least-squares GAN
loss (LSGAN) \cite{DBLP:journals/corr/MaoLXLW16}.

Let $(s^{e},s^{e}_{g},s^{e}_{\text{next}})\sim D_{g}$ denote an subgoal
transition where $s^{e}$ is the current state, $s^{e}_{g}$ the
(supervised) sub-goal, $s^{e}_{\text{next}}$ the next state, and
$g^{e}$ the final task goal.  The high-level policy
$\pi^{H}_{\theta_{H}}(\,\cdot\,|s^{e},g^{e})$ predicts a sub-goal
$s_{g}$, while a discriminator
$D^{H}_{\epsilon_{H}}$ with parameters $\epsilon_{H}$ tries to distinguish expert sub-goals from
policy-generated ones. $J_D^{H}$ represents upper level IRL objective, which depends on parameters $(\theta_{H},\epsilon_{H})$. We bootstrap the learning of higher level policy by optimizing:

% \begin{equation}
% \label{eq:irl_high}
% \resizebox{\linewidth}{!}{$
% \displaystyle
% \max_{\theta_{H}}\min_{\epsilon_{H}}\;
% J^{H}_{D}(\theta_{H},\epsilon_{H})=
% \frac12\,
% \mathbb{E}_{(s^{e},s^{e}_{g})\sim D_{g}}
% \!\bigl[(D^{H}_{\epsilon_{H}}(s^{e}_{g})-1)^{2}\bigr]
% +\frac12\,
% \mathbb{E}_{s^{e}\sim D_{g},\;\hat{s}_{g}\sim\pi^{H}_{\theta_{H}}}
% \!\bigl[D^{H}_{\epsilon_{H}}(\hat{s}_{g})^{2}\bigr].
% $}
% \end{equation}

% \end{align*}
\begin{equation}
\label{eqn:irl_update}
    \begin{split}
    & \max_{\theta_{H}}\min_{\epsilon_{H}} J_D^{H}(\theta_{H}, \epsilon_{H}) = \max_{\theta_{H}}\min_{\epsilon_{H}} \frac{1}{2}\mathbb{E}_{(s^e, s^e_g, \cdot) \sim D_g} [\mathbb{D}_{\epsilon_{H}}^H(s^e_g) - 1]^2
    \\ & \hspace{0.9cm} + \frac{1}{2}\mathbb{E}_{(s^e,\cdot,\cdot)\sim D_g, s_g \sim \pi_{\theta_H}^{H}(\cdot|s^e, g^e)} [\mathbb{D}_{\epsilon_{H}}^H(\pi_{\theta_H}^{H}(\cdot|s^e, g^e)) - 0]^2.
    \end{split}
\end{equation}

Minimising \eqref{eqn:irl_update} with respect to $\epsilon_{H}$ and
maximising it with respect to $\theta_{H}$ pushes
$\pi^{H}_{\theta_{H}}$ to generate sub-goals that are
indistinguishable from those in $D_{g}$, thus inducing an automatic
curriculum whose difficulty evolves according to the
low-level policy.

% This objective forces the higher policy subgoal predictions to be close to subgoal predictions of the dataset $D_g$. The discriminator $\mathbb{D}_{\epsilon_{H}}^H$ creates a natural curriculum for regularizing higher level policy by assigning the value $1$ to the predicted subgoals that are closer to the subgoals from dataset $D_g$, and $0$ otherwise. $\mathbb{D}_{\epsilon_{H}}^H$ improves with training, and regularizes the higher policy to predict achievable subgoals for the lower primitive.
\par Similarly for lower level primitive, let $(s^f, a^f, s^f_{next}) \sim D_g^L$ be lower level expert transition where $s^f$ is current state, $s^f_{next}$ is next state, $g^f$ is final goal, $a$ is the primitive action predicted by lower policy $\pi_{\theta_L}^{L}(\cdot|s^f, s^e_g)$ with parameters $\theta_L$, and $\mathbb{D}_{\epsilon_{L}}^L$ be the lower level discriminator with parameters $\epsilon_{L}$. Let $J_D^{L}$ represent lower level IRL objective, which depends on parameters $(\theta_{L},\epsilon_{L})$. The lower level IRL objective is thus:
\begin{equation}
\label{eqn:irl_update_lower}
    \begin{split}
    & \max_{\theta_{L}}\min_{\epsilon_{L}} J_D^{L}(\theta_{L}, \epsilon_{L}) = \max_{\theta_{L}}\min_{\epsilon_{L}} \frac{1}{2}\mathbb{E}_{(s^f, a^f, \cdot) \sim D_g^L} [\mathbb{D}_{\epsilon_{L}}^L(a^f) - 1]^2
    \\ & \hspace{1.1cm} + \frac{1}{2}\mathbb{E}_{(s^f,\cdot,\cdot)\sim D_g^L, a \sim \pi_{\theta_L}^{L}(\cdot|s^f, s^e_g)} [\mathbb{D}_{\epsilon_{L}}^L(\pi_{\theta_L}^{L}(\cdot|s^f, s^e_g)) - 0]^2.
    \end{split}
\end{equation}

\subsection{Joint optimization}
% Finally, the higher level policy is trained to produce subgoals, which when fed into the lower level primitive, maximize the sum of future discounted rewards for our task using off-policy reinforcement learning.

% Here $T$ is the task horizon and $g$ is the sampled goal for the current episode. For brevity, we can refer to this objective function as $J^{H}_{\theta_{H}}$ and $J^{L}_{\theta_{L}}$ for upper and lower level policies. We use the IRL objective to regularize the high level off-policy RL objective:
% \begin{equation}
% \label{eqn:joint_update_upper}
%     \max_{\theta_{H}}(J^{H}_{\theta_{H}} + \psi * (\min_{\epsilon_{H}} J_D^{H}(\theta_{H}, \epsilon_{H}))),
% \end{equation}
% whereas the lower level primitive is trained by optimizing:
% \begin{equation}
% \label{eqn:joint_update_lower}
%     \max_{\theta_{L}} (J^{L}_{\theta_{L}} + \psi * (\min_{\epsilon_{L}} J_D^{L}(\theta_{L}, \epsilon_{L}))).
% \end{equation}

% The lower policy is regularized using expert demonstration dataset, and the upper level is optimized using subgoal transition dataset populated using PIP. $\psi$ is the regularization weight for the IRL objective. When $\psi=0$, the method reduces to HRL policy with no higher level policy regularization, and large values of $\psi$ may lead to  overfitting to the expert demonstrations. Refer to Algorithm~\ref{alg:algo_crisp} for CRISP algorithm.

To train both hierarchical levels stably, we combine off-policy RL with IRL regularization. While the off-policy RL objective enables agents to autonomously explore and learn from their interactions with the environment, the IRL objective allows them to leverage expert demonstrations for more sample-efficient and guided skill acquisition. The high-level policy is optimized to generate subgoals which, when provided to the lower-level primitive, maximize the expected sum of discounted rewards for each task.

\noindent Let $T$ denote the episode horizon and $g$ the sampled episodic goal. We denote the standard RL objectives for high and low levels as $J^{H}_{\theta_{H}}$ and $J^{L}_{\theta_{L}}$, respectively. Both objectives are augmented with their respective IRL losses, weighted by a regularization factor $\psi$. The joint optimization objectives are:

\begin{equation}
\label{eqn:joint_update_upper}
    \max_{\theta_{H}}\, 
    \Big[ J^{H}_{\theta_{H}} + \psi\, \min_{\epsilon_{H}} J_D^{H}(\theta_{H}, \epsilon_{H}) \Big],
\end{equation}
\begin{equation}
\label{eqn:joint_update_lower}
    \max_{\theta_{L}}\,
    \Big[ J^{L}_{\theta_{L}} + \psi\, \min_{\epsilon_{L}} J_D^{L}(\theta_{L}, \epsilon_{L}) \Big].
\end{equation}

The lower-level policy is regularized using expert demonstration data, while the upper-level policy leverages the subgoal transition dataset generated by PIP. $\psi$ trades off between pure task reward and IRL-guided imitation: $\psi=0$ yields a vanilla HRL agent with no regularization; high values emphasize imitation and may risk overfitting to demonstrations.

\smallskip
See Algorithm~\ref{alg:algo_crisp} for the full training pseudocode. This joint framework enables continual adaptation at both levels: the lower-level primitive explores via RL to reach higher-level subgoals, while its IRL regularizer keeps its behavior close to expert demonstrations, ensuring reliable skill learning. Simultaneously, the higher-level policy uses IRL guidance to produce subgoals attainable by the evolving primitive, promoting task progress without inducing non-stationarity and thereby stabilizing hierarchical training.

%% file: sections/experiments.tex
\section{Experiments}
\label{sec:experiment1}

\input{figures_tex/success_rate_comparison.tex}

\par In this section, we perform experiments to answer the following questions: 
\begin{itemize}
    \item \textbf{Q1.} Does CRISP's primitive-informed parsing outperform fixed-window parsing approaches?
    \item \textbf{Q2.} Does CRISP improve sample efficiency and stability over state-of-the-art HRL baselines? 
    \item \textbf{Q3.} How well does CRISP mitigate HRL non-stationarity? 
    \item \textbf{Q4.} Does IRL regularization generate a curriculum of achievable subgoals?
    \item \textbf{Q5.} Can CRISP policies transfer to real-world robots?  
    \item \textbf{Q6.} What is the impact of each design choice?
\end{itemize}
\noindent \textbf{Environment and Implementation details.} We perform experiments on six complex robotic environments with continuous state and action spaces that require long term planning: $(i)$ maze navigation, $(ii)$ pick and place, $(iii)$ bin, $(iv)$ hollow, $(v)$ rope manipulation, and $(vi)$ franka kitchen. We use the off-policy Soft Actor Critic~\cite{DBLP:journals/corr/abs-1801-01290} as RL algorithm with Adam~\cite{kingma2014method} optimizer. Extensive implementation and environment details are provided in Appendix Sections~\ref{appendix:implementation_details} and~\ref{sec:environment_details}. We collect $28$ expert demonstrations in kitchen and $100$ demonstrations in all other tasks. The datasets are collected such that they cover a large enough distribution of initial state
conditions. For each baseline, we conducted a comprehensive hyper-parameter grid search within the ranges recommended by the original publications to ensure optimal performance. We provide the hyper-parameter list in Appendix Section~\ref{appendix:hyperparameters}, and demonstrations generation approach in Appendix Section~\ref{sec:appendix_expert_demos}.
\\
\\
\noindent \textbf{Sparse Reward and Goal-Conditioning Complexity.} While some tasks may appear simple, all evaluation tasks employ \emph{sparse rewards}, where the agent has to extensively explore the environment before coming across any rewards, thus significantly increasing task complexity. For example, in the Franka Kitchen task, the agents gets a reward only on completing the final task (e.g. opening the microwave and turning on the gas knob). We consider a goal-conditioned RL setting, where the final goals are randomly generated, which further increases task complexity. These design choices elevate task complexity and highlight the advantages of learning efficient hierarchical policies with curriculum generation, as naive or flat baseline methods struggle to make progress.
\\
\\
\noindent \textbf{Evaluating two CRISP regularizers.} Along with our RL objective, We evaluate CRISP using two imitation learning regularizers: IRL regularization (CRISP-IRL) and BC regularization (CRISP-BC). Evaluating both BC (behavior cloning) and IRL (inverse-RL) variants offers complementary insights due to their fundamentally different approaches to imitation learning (IL). BC directly learns a mapping from states to expert actions, making it sample-efficient and effective when demonstrations are abundant, high-quality, and cover the relevant state distribution. However, BC is prone to compounding errors and struggles with distributional shift, which can degrade performance in complex or noisy environments. IRL, in contrast, infers the underlying reward function guiding the expert’s behavior, enabling the agent to generalize beyond demonstrations by optimizing policies that reflect the expert’s intent, even in unseen states. This often leads to better robustness and the ability to surpass imperfect experts but can require more complex training. By separately implementing both IRL and BC variants, we gain a clearer understanding of their relative strengths and weaknesses across different tasks, and also renders insights into which IL approach is better suited for specific robotic control tasks.

In Figure \ref{fig:success_rate_comparison} we report the success rates of CRISP alongside all baselines, averaged over five independent random seeds. Training conditions (network architecture, optimizer, batch size, replay buffer size, etc.) were kept identical across methods unless stated otherwise; each baseline was re-implemented from scratch and its hyper-parameters grid-searched within the ranges recommended by the original papers to guarantee a fair comparison.

\noindent The subsections that follow present empirical results that systematically address each of the questions listed above.
\\
\\
\noindent \textbf{Q1: Does CRISP's adaptive primitive-informed parsing outperform fixed-window parsing approaches?}
\input{figures_tex/non_stationarity}
\\
\\
In Figure~\ref{fig:success_rate_comparison}, we compare CRISP with \textit{RPL} (Relay Policy Learning)~\cite{DBLP:journals/corr/abs-1910-11956}, to demonstrate the efficacy of primitive-informed parsing compared to fixed window based parsing. RPL first uses supervised pre-training from undirected demonstrations, and then fine-tunes the policy using RL. To ensure fair comparisons, we use a variant of \textit{RPL} which does not use this pre-training. Thus, the only difference between CRISP and RPL is that CRISP uses primitive-informed parsing to select subgoals, and RPL uses fixed window based parsing.

As illustrated in Figure~\ref{fig:success_rate_comparison}, CRISP consistently outperforms \textit{RPL} across all evaluated tasks, highlighting the effectiveness of primitive-informed parsing over fixed window-based approaches. While both methods leverage a combination of RL and IL, their key difference is in how subgoals are assigned. CRISP’s primitive-informed parsing dynamically selects subgoals that are achievable by the current lower-level primitive, thus automatically adjusting the subgoal difficulty without the need for manual tuning.

This adaptability proves especially valuable when perturbations or random exploration cause the agent to deviate from the demonstration trajectory. In such cases, CRISP’s high-level policy responds by generating future subgoals that steer the primitive back toward success, enabling recovery from off-manifold states where RPL typically fails. For example, in rope manipulation experiments, agents using fixed window parsing often become stuck after a poor poke, unable to recover. In contrast, CRISP promptly proposes a reachable intermediate rope configuration, allowing the agent to proceed and ultimately succeed.

We also elucidate the importance of primitive-informed parsing by considering with a variant of CRISP that uses the subgoal dataset collected using fixed window based parsing. We compare this variant (CRISP-RPL) with our approach (CRISP-IRL) in Appendix Figure~\ref{fig:rpl_rpl_ablation}. (CRISP-IRL) consistently outperforms this baseline, showing that primitive-informed parsing is crucial for improved performance. 
\\
\\
\noindent \textbf{Q2: Does CRISP improve sample efficiency and stability over state-of-the-art HRL baselines?}
\input{figures_tex/curriculum_figure}
\\
\\
We benchmark CRISP against diverse hierarchical and flat baselines to gauge sample efficiency and training stability. HAC~\cite{DBLP:journals/corr/abs-1712-00948} assumes an optimal low-level primitive, which is often untrue in practice. Thus, HAC may produce sub-optimal subgoals. As Figure \ref{fig:success_rate_comparison} shows, HAC fares well on the simple maze task but fails on harder tasks. In contrast, CRISP’s primitive-informed parsing dynamically aligns the subgoal difficulty with the evolving capabilities of the lower-level primitive, yielding stable HRL learning and higher success rates performance. SAGA~\cite{wang2023state} uses a discriminator generate subgoals but still suffers when those goals remain unreachable. CRISP consistently surpasses SAGA on harder tasks due to primitive-informed parsing and IRL regularization, which ensure subgoals remain achievable and mitigate non-stationarity.

In \textit{RAPS}~\cite{DBLP:journals/corr/abs-2110-15360}, hand-crafted action primitives drive the low-level policy, leaving the high level to choose their sequence. Designing these primitives is labour-intensive, and except for the  maze navigation task, RAPS performs poorly. We attribute this to navigation reducing the high-level choice to a simple direction selection, whereas other domains demand more complex primitives. Further, our sparse-reward re-implementation of tasks accounts for RAPS's poor performance. In contrast, CRISP outperforms RAPS on all tested manipulation tasks without requiring any domain-specific primitives. We do not evaluate RAPS on the rope task because creating suitable primitives there proved impractical.

We also compare our approach with two other hierarchical baselines: \textit{HIER} and \textit{HIER-NEG}, which are hierarchical off-policy SAC~\cite{DBLP:journals/corr/abs-1801-01290} based baselines that do not leverage expert demonstrations. In \textit{HIER-NEG}, the higher level policy is negatively rewarded if the lower primitive is unable to reach the predicted subgoal. CRISP achieves notably higher performance compared to these hierarchical baselines, indicating that improvements stem not only from hierarchical abstraction but from our primitive-informed parsing and primitive regularization, that directly mitigates non-stationarity and enforces subgoal reachability.

Finally, we compare CRISP against flat baselines, including (DAC) Discriminator Actor Critic (which leverages expert demonstrations), single-level RL (FLAT), and behavior cloning (BC). These baselines fail to reliably solve the tasks tested, reinforcing the critical importance of both hierarchical abstraction and curriculum-based subgoal adaptation. Thus, our comprehensive evaluation demonstrates that CRISP significantly improves sample efficiency and stabilizes learning compared to several existing hierarchical and flat baselines.
\\
\\
\noindent \textbf{Q3: How well does CRISP mitigate non-stationarity?}
\\
\\
We illustrate CRISP’s ability to mitigate non-stationarity in HRL in Figure~\ref{fig:non_stationarity}. To quantify this, we compare CRISP to the \textit{HIER} baseline by measuring the average distance during several stages of training (Initial: when training begins, Mid: half-way during training, Final: when training ends, e.g. since maze navigation is trained for 4.7\textbf{E}6 timesteps, the values are Initial: iteration 1, Mid: iteration 2.35\textbf{E}6, and Final: iteration 4.7\textbf{E}6) between the subgoals issued by the higher-level policy and the actual states reached by the lower-level primitive. Lower distance values indicate that the higher-level policy is generating subgoals that are well-aligned with the capabilities of the lower-level primitive, which leads to more consistent goal achievement and reduces non-stationarity within the hierarchical learning process. As shown, CRISP consistently produces achievable subgoals throughout training, resulting in superior stability and effective non-stationarity mitigation.
\\
\\
\noindent \textbf{Q4: Does PIP and IRL regularization generate a curriculum of achievable subgoals?}
\\
\\
Primitive-Informed Parsing (PIP) and IRL regularization plays a crucial role in generating a curriculum of subgoals that dynamically adapt to the lower-level primitive’s evolving capabilities. PIP periodically segments expert demonstrations by identifying subgoals that the current low-level controller can realistically achieve within a given horizon, effectively ensuring that each subgoal is feasible. Subsequently, IRL regularization regularizes the higher level policy to predict such achievable subgoals according to the current lower level primitive. This adaptive parsing leads to an implicit curriculum: early in training, subgoals are simpler and localized, while later stages naturally involve more complex and distant targets as the primitive improves. This is evident in Figure~\ref{fig:env_curriculum}, where early on, subgoals cluster close to the agent (Row 1), enabling quick wins; mid-training (Row 2) they move outward and diversify; by the final phase (Row 3) they span the full task horizon and often coincide with, or lie just before, the final goal. The increasing subgoal difficulty across rows demonstrates CRISP’s ability to generate a curriculum of achievable subgoals for the lower primitive.
\\
\\
\noindent \textbf{Q5: Can CRISP transfer to real-world robots?}
\label{real_subsection}
\\
\\
To evaluate real-world deployability, we conducted experiments on pick-and-place, bin packing, and rope manipulation tasks using a four-axis Dobot Magician desktop arm (Appendix Figure~\ref{fig:dobot_real}). The robot was controlled via the manufacturer’s Python SDK at a fixed 20 Hz command rate ($\delta t = 50$ ms). Each episode lasted approximately 40 seconds for pick-and-place and bin tasks, and 60 seconds for rope manipulation—allowing sufficient time for joint-space commands to achieve the goal or time out. After each episode, a “Home” macro returned the arm to its calibrated start pose, followed by a brief scripted motion to position the end-effector above the workspace. This automatic reset took roughly 10 seconds on average, minimizing downtime and ensuring consistent initial conditions. We used a Realsense D435 depth camera to track the robot, block, bin, and rope cylinder positions. Due to the difficulty of precisely estimating linear and angular velocities in real settings, we assigned them small fixed values, which proved effective in practice. We performed five sets of ten trials per task and report the average success rates.

In real-robot experiments, the ability to structure the task into feasible subgoals is essential: both CRISP-IRL and CRISP-BC successfully decompose complex tasks into reachable stages, leading to notably higher real-world success rates. Further, by ensuring that subgoals correspond to attainable states, these methods inherently promote safer operation, as the robot avoids attempting unsafe or unreachable maneuvers. CRISP-IRL achieves an accuracy of $0.6$, $0.6$ and $0.5$, whereas CRISP-BC achieves accuracy of $0.8$, $0.3$, $0.3$ on pick and place, bin and rope tasks respectively. We also deployed the next best performing baseline \textit{RPL} on the tasks, however it is unable to generate good subgoals and the agent gets stuck, thus failing to show any progress in the tasks. 
\\
\\
\noindent \textbf{Q6: What is the impact of each design choice?}
\\
\\
We also perform ablations to select the population hyperparameter $p$ (Appendix Figure~\ref{fig:p_ablation}), learning rate $\psi$ (Appendix Figure~\ref{fig:psi_ablation}), RPL window size ablation (Appendix Figure~\ref{fig:rpl_ablation}), and the optimal number of expert demos (Appendix Figure~\ref{fig:demos_ablation}) in Appendix Section~\ref{sec:appendix_ablations}. Extensive ablation studies show that CRISP maintains strong performance across a wide range of hyper-parameter settings, indicating that the method is not unduly sensitive to precise hyper-parameter tuning. We empirically found that very large values of $p$ are unable to generate good curriculum of subgoals. Further, when $\psi$ is too high, the method might overfit to expert data, whereas if $\psi$ is too small, CRISP is unable to utilize IL regularization. We perform ablations to deduce the optimal number of expert demos required for each task. If the number of expert demos is too small, the policy may overfit. Although the number of expert demos are subject to availability, we increase the number until there is no significant improvement in performance. We analyse the effect of varying the quality of expert data, and found direct co-relation quality of expert data and performance. The best value of window size hyper-parameter $k$ on RPL experiments is chosen. Finally, we provide qualitative visualizations for all tasks in Appendix Section~\ref{sec:appendix_qualitative_viz}.

%% file: figures_tex/success_rate_comparison.tex
\begin{figure*}[t]
\vspace{1pt}
\centering
% \captionsetup{font=footnotesize,labelfont=scriptsize,textfont=scriptsize}
\subfloat[][Maze navigation]{\includegraphics[scale=0.27]{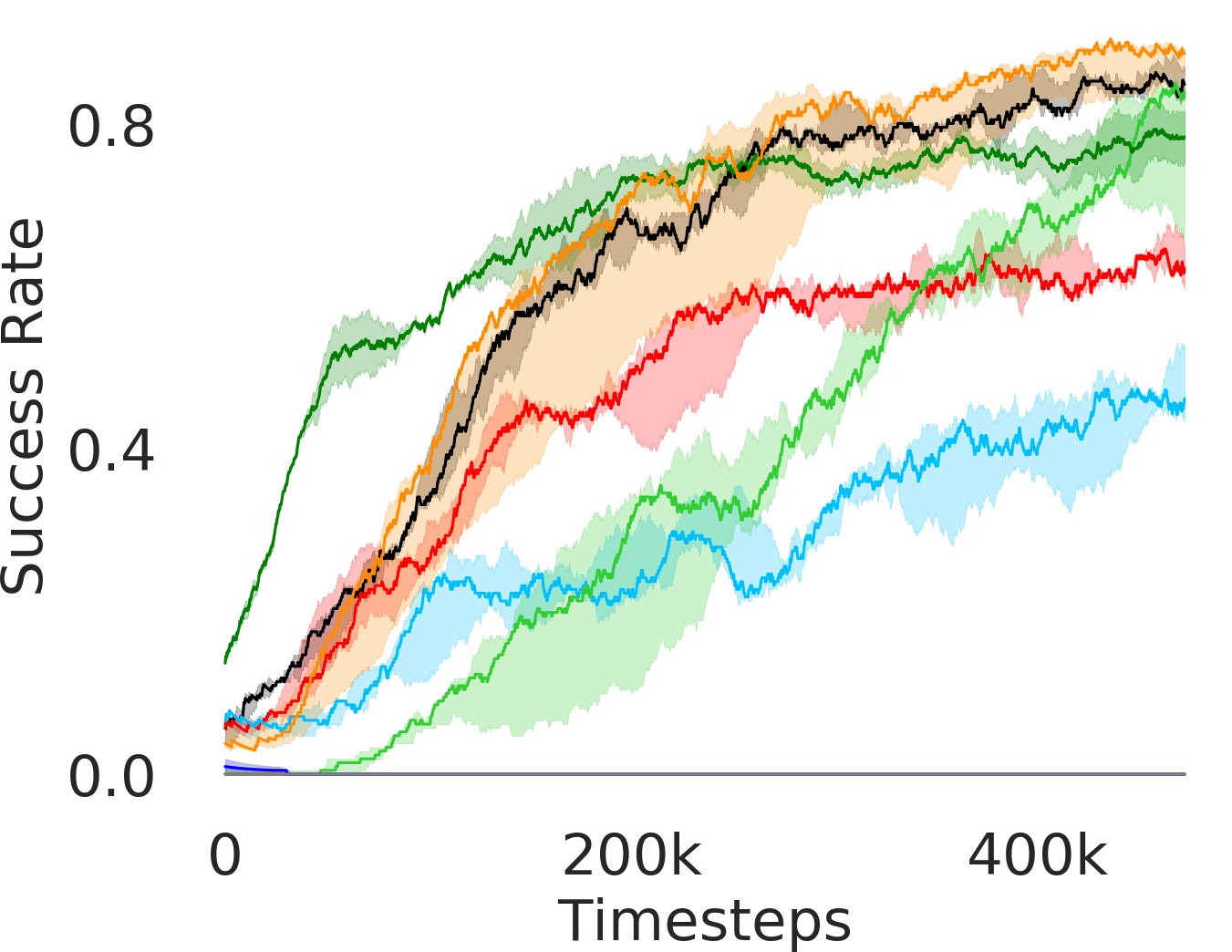}}
\subfloat[][Pick and place]{\includegraphics[scale=0.27]{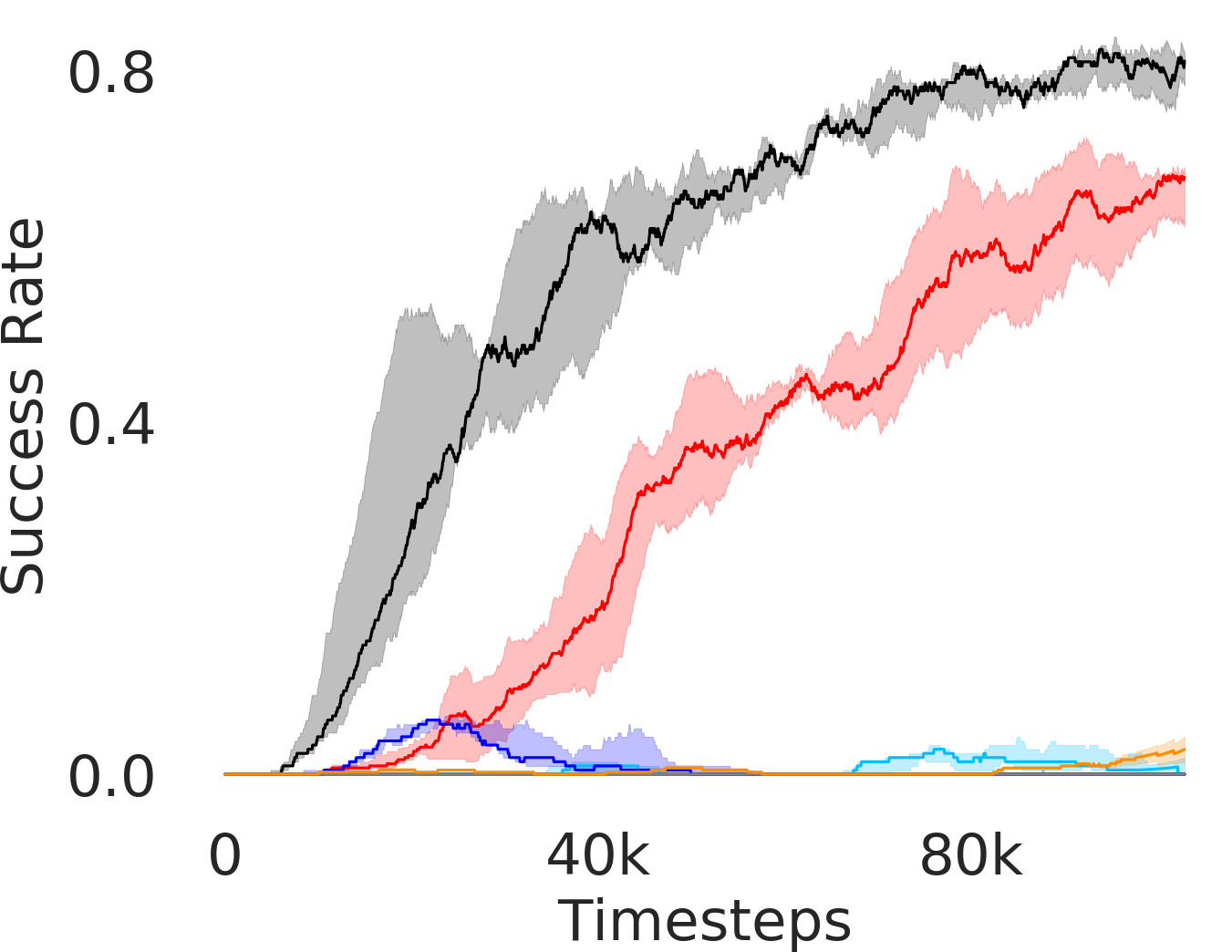}}
\subfloat[][Bin]{\includegraphics[scale=0.27]{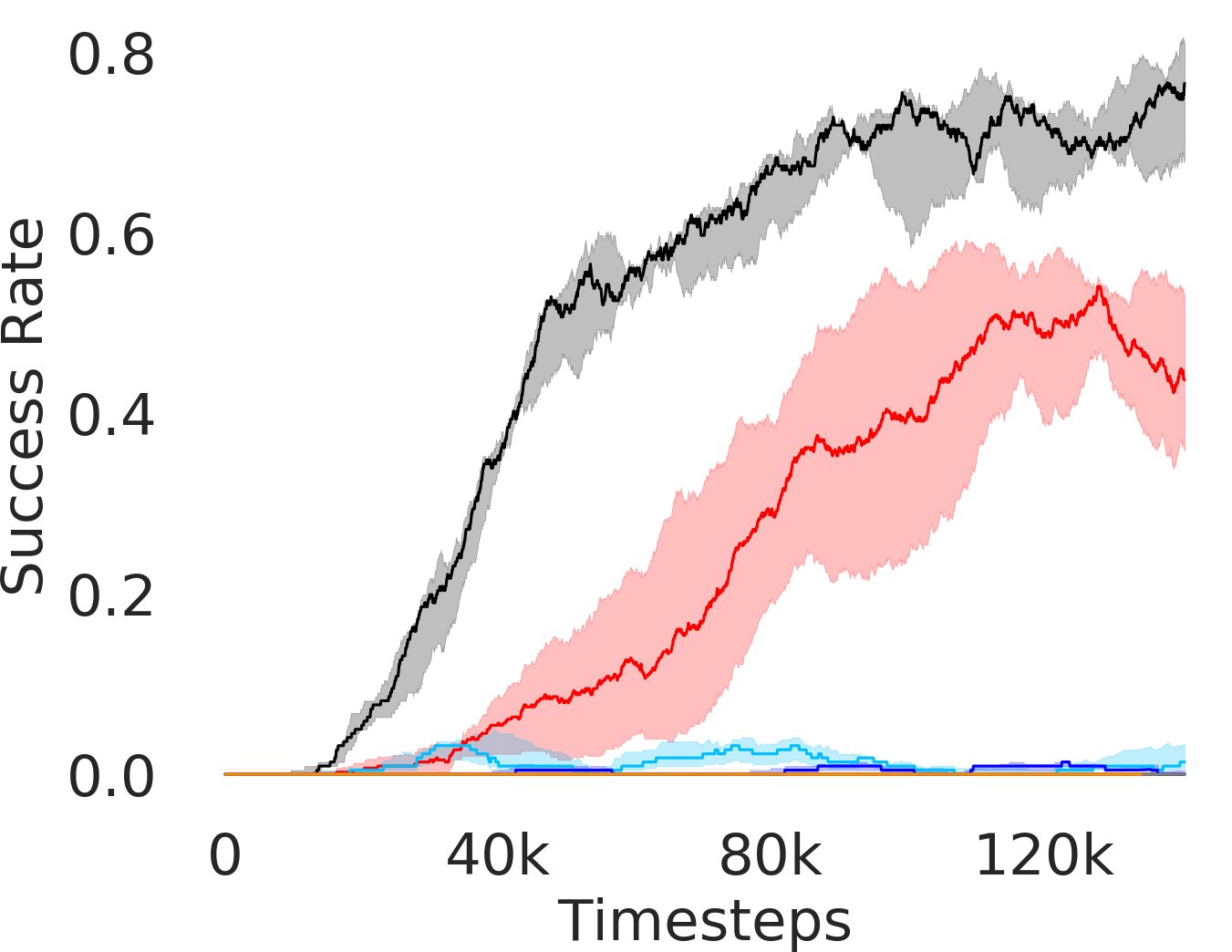}}
\\
\subfloat[][Hollow]{\includegraphics[scale=0.27]{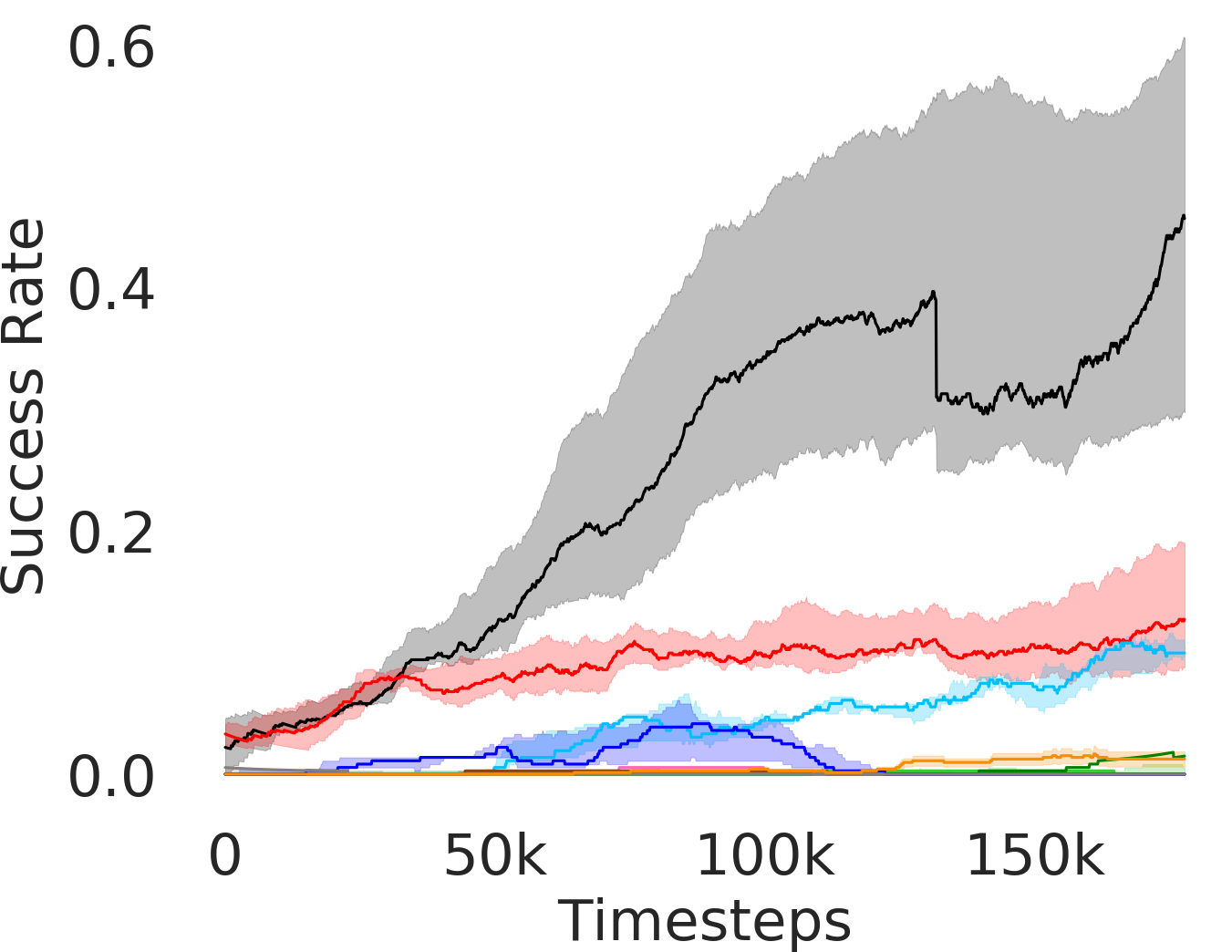}}
\subfloat[][Rope]{\includegraphics[scale=0.27]{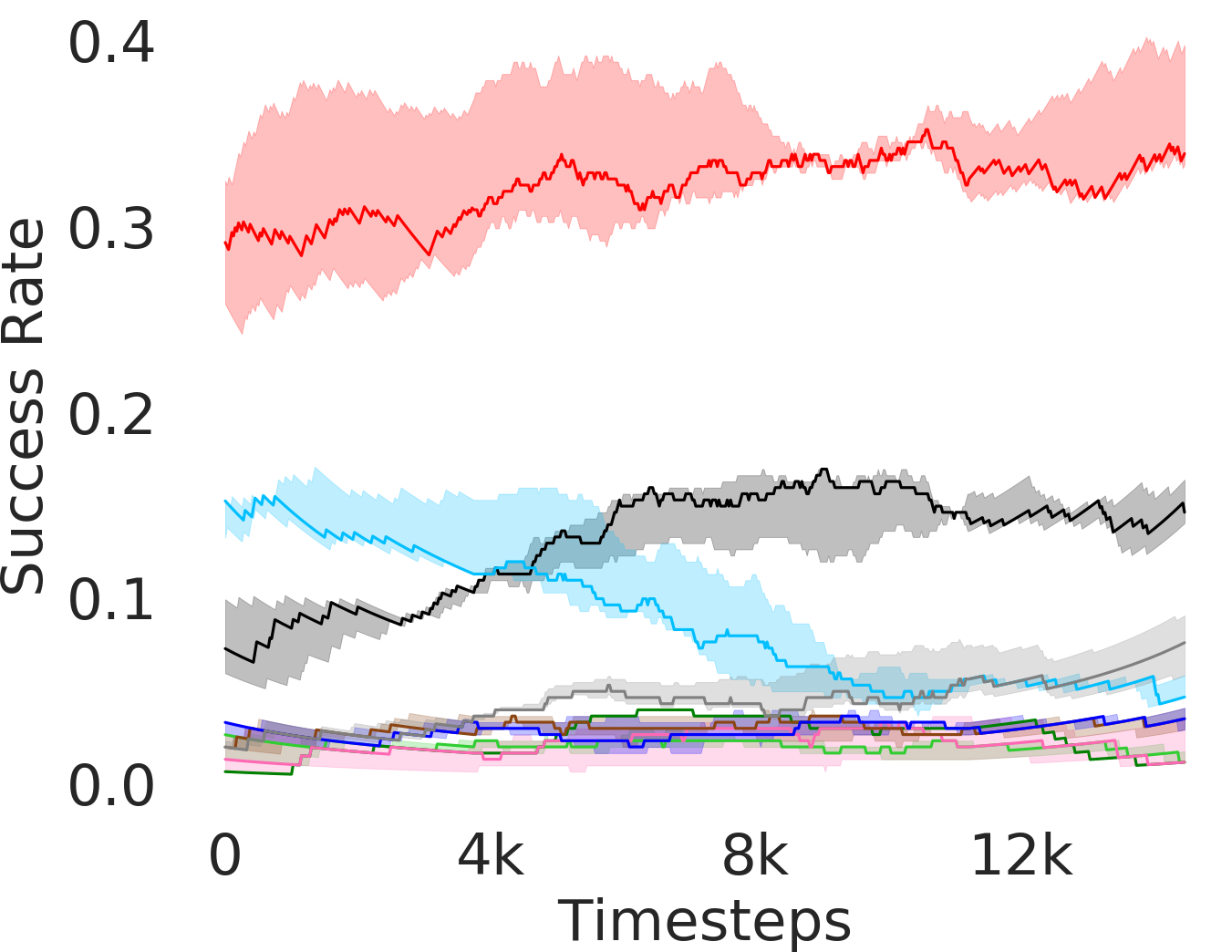}}
\subfloat[][Franka kitchen]{\includegraphics[scale=0.27]{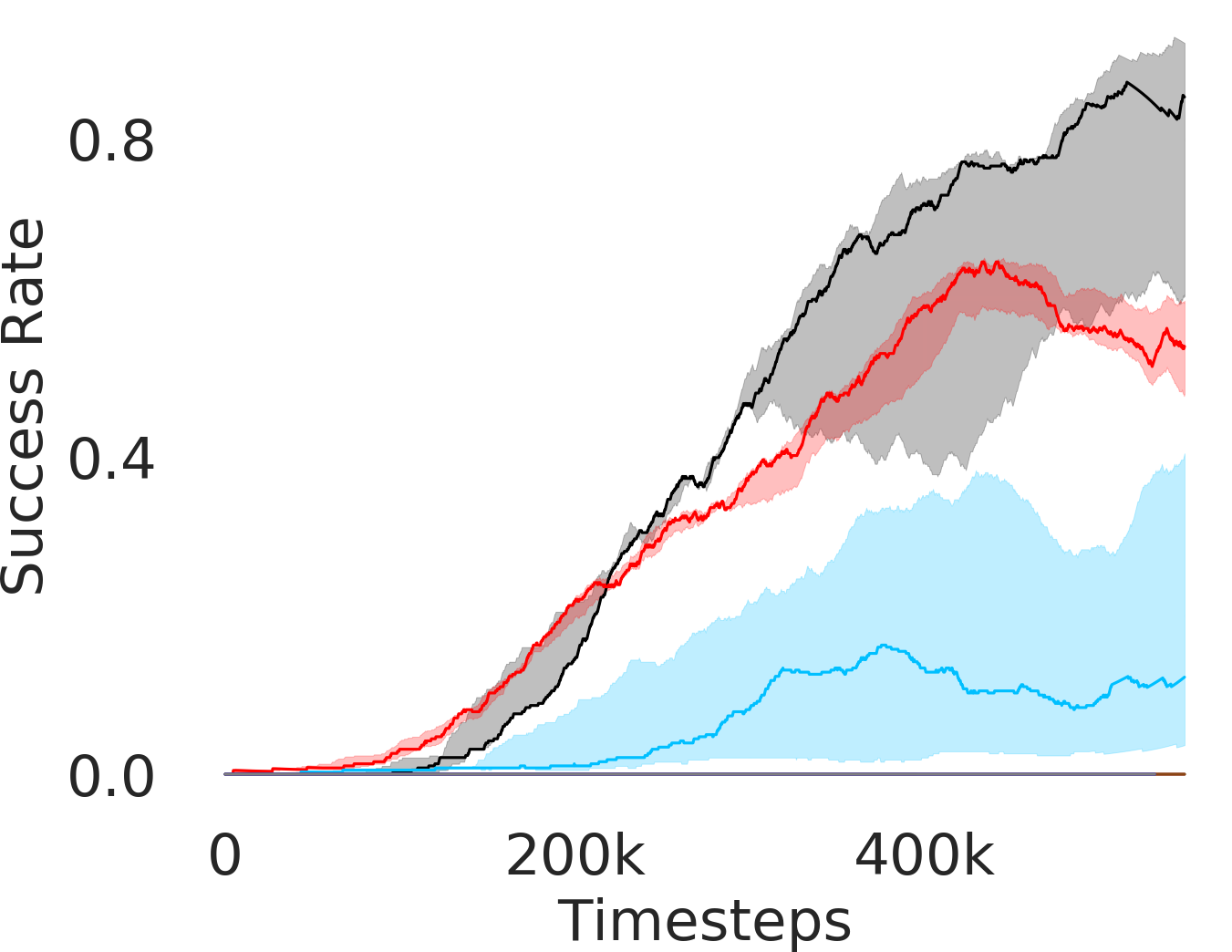}}
% \subfloat[][Maze]{\includegraphics[height=2cm,width=2cm]{figures/pick_distance.png}}
% \subfloat[][Maze]{\includegraphics[height=2cm,width=2cm]{figures/rope_distance.png}}
% \includegraphics[height=4cm,width=12cm]{figures/train_test_val_collage_final.png}
% \includegraphics[scale=0.27]{figures/eight_room_comparison.png}
\\
{\includegraphics[scale=0.6]{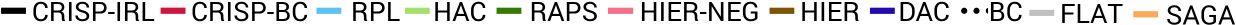}}
\caption{\textbf{Success rate comparison} This figure compares the methods on six sparse robotic environments. The solid line represents the mean and shaded region represents the standard deviation across 5 seeds. We compare CRISP-IRL and CRISP-BC against multiple baselines: RPL (Relay Policy learning~\cite{DBLP:journals/corr/abs-1910-11956}), HAC (Hierarchical Actor Critic~\cite{DBLP:journals/corr/abs-1712-00948}, SAGA~\cite{wang2023state}, RAPS~\cite{DBLP:journals/corr/abs-2110-15360}, HIER-NEG (HRL implementation using SAC~\cite{DBLP:journals/corr/abs-1801-01290} where higher policy is negatively rewarded when lower primitive fails to achieve the predited subgoal), HIER (HRL implementation using SAC), DAC~\cite{kostrikov2018discriminator}, and FLAT (Single-level RL policy using SAC). As seen from the plots, CRISP achieves more than 40\% higher success rates, shows better convergence speed and training stability over the baselines.}
\label{fig:success_rate_comparison}
\end{figure*}

% \begin{figure}[t]
% \vspace{5pt}
% \centering
% \subfloat[][Maze]{\includegraphics[scale=0.27, scale=0.27]{figures/success_rate_four_room_sep.png}}
% \subfloat[][Maze]{\includegraphics[scale=0.27, scale=0.27]{figures/success_rate_pick_sep.png}}
% \subfloat[][Maze]{\includegraphics[scale=0.27, scale=0.27]{figures/success_rate_rope_sep.png}}
% \subfloat[][Maze]{\includegraphics[scale=0.27, scale=0.27]{figures/flat_distance.png}}
% \subfloat[][Maze]{\includegraphics[scale=0.27, scale=0.27]{figures/pick_distance.png}}
% \subfloat[][Maze]{\includegraphics[scale=0.27, scale=0.27]{figures/rope_distance.png}}
% % \includegraphics[scale=0.27, height=4cm,width=12cm]{figures/train_test_val_collage_final.png}
% % \includegraphics[scale=0.27]{figures/eight_room_comparison.png}
% \caption{In row 1, we compare the success rates between our method and baselines in maze navigation (Col. 1), pick and place (Col. 2), and rope manipulation (Col. 3) versus number of training epochs. In row 2, the methods are compared via distance metric (average distance between achieved goal and final goal in 100 episodic rollouts) in maze navigation (Col. 1), pick and place (Col. 2), and rope manipulation environment (Col. 3).}
% \label{fig:success_rate_comparison}
% \end{figure}

%% file: figures_tex/non_stationarity.tex
\begin{figure*}[t]
% \begin{minipage}{0.59\textwidth}
\centering
% \captionsetup{font=footnotesize,labelfont=scriptsize,textfont=scriptsize}
\subfloat[][Maze navigation]{\includegraphics[scale=0.27]{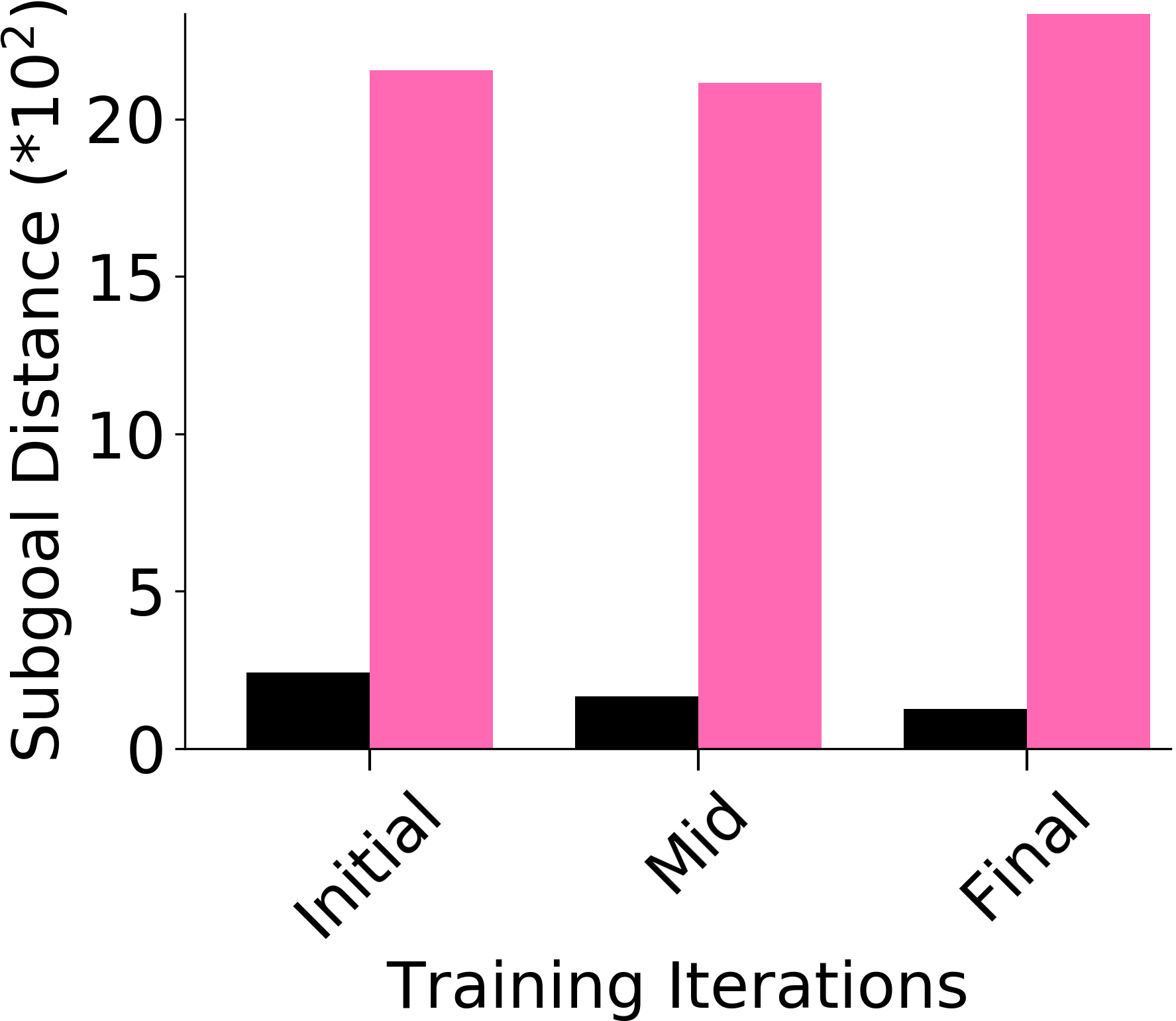}}\hspace{0.3cm}
\subfloat[][Pick and place]{\includegraphics[scale=0.27]{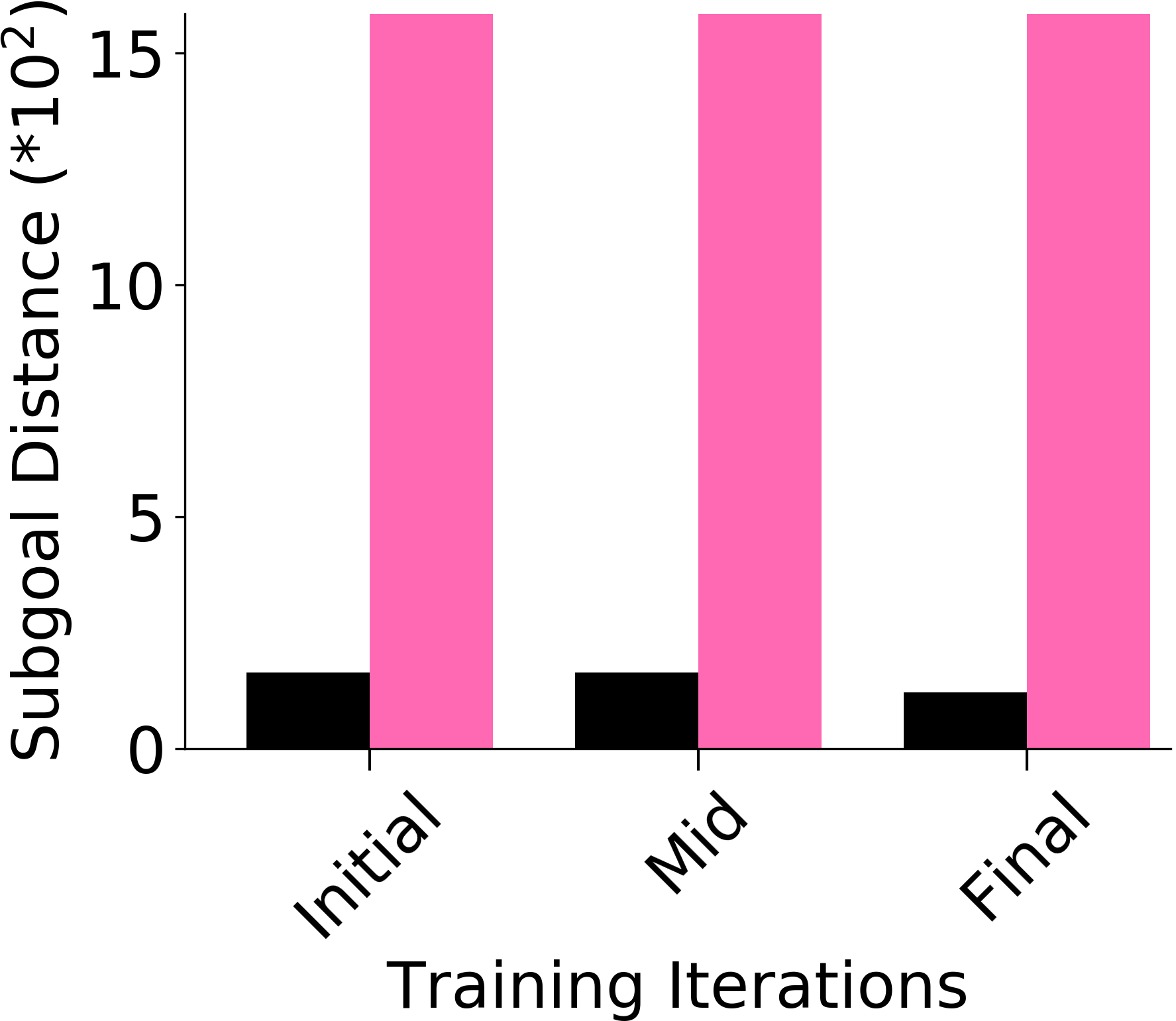}}\hspace{0.3cm}
\subfloat[][Bin]{\includegraphics[scale=0.27]{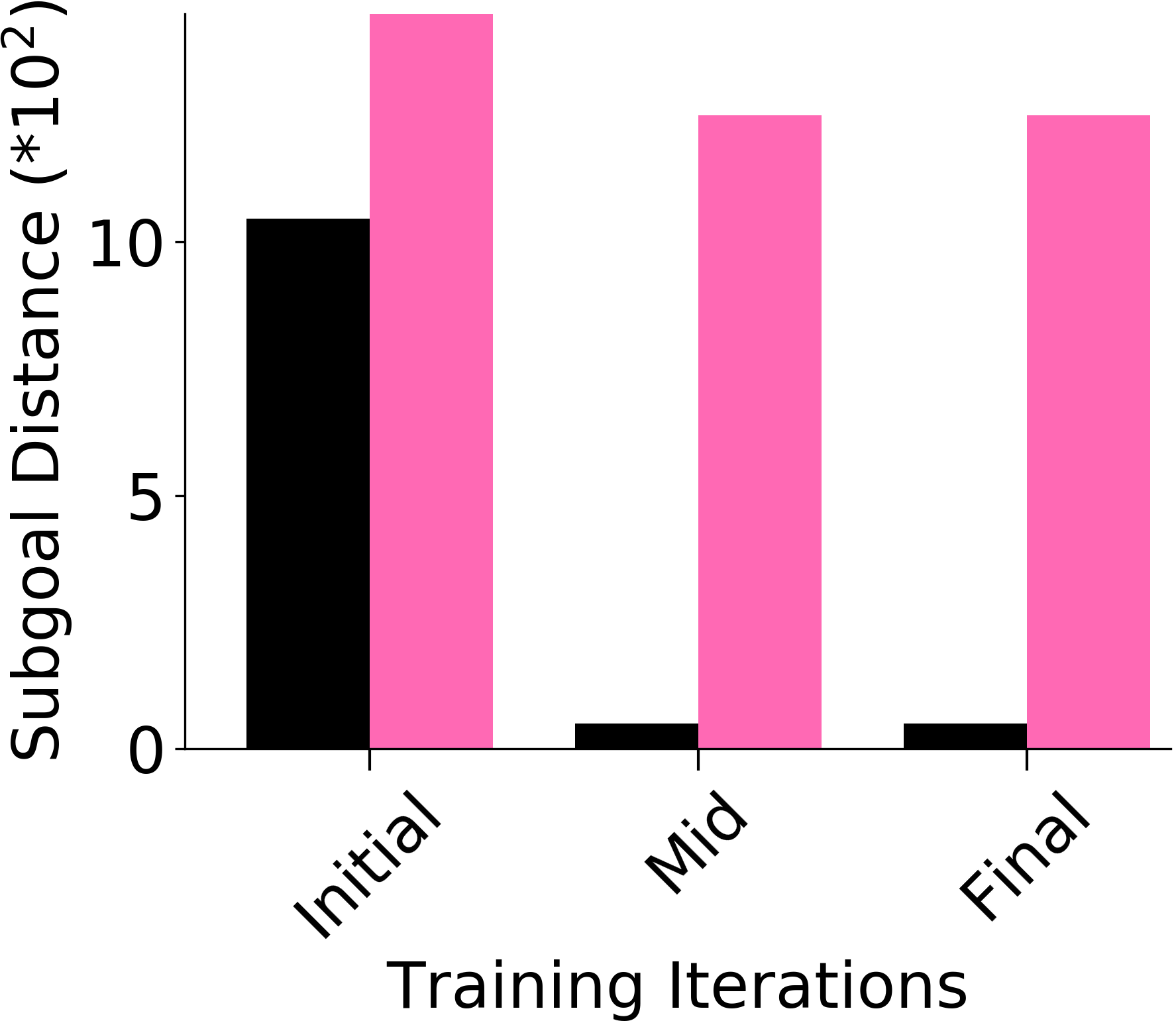}} \\
\subfloat[][Hollow]{\includegraphics[scale=0.27]{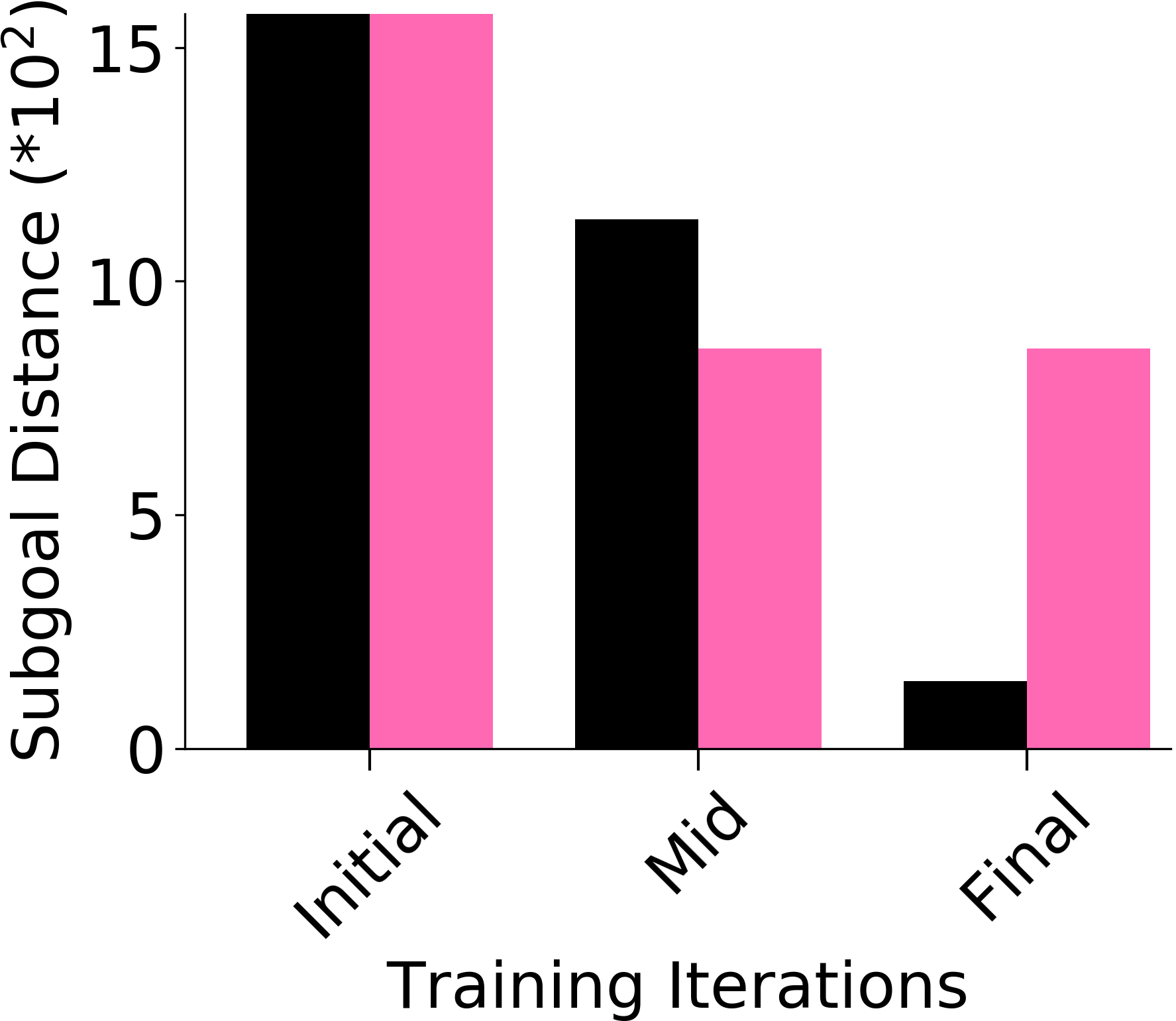}}\hspace{0.3cm}
\subfloat[][Rope]{\includegraphics[scale=0.27]{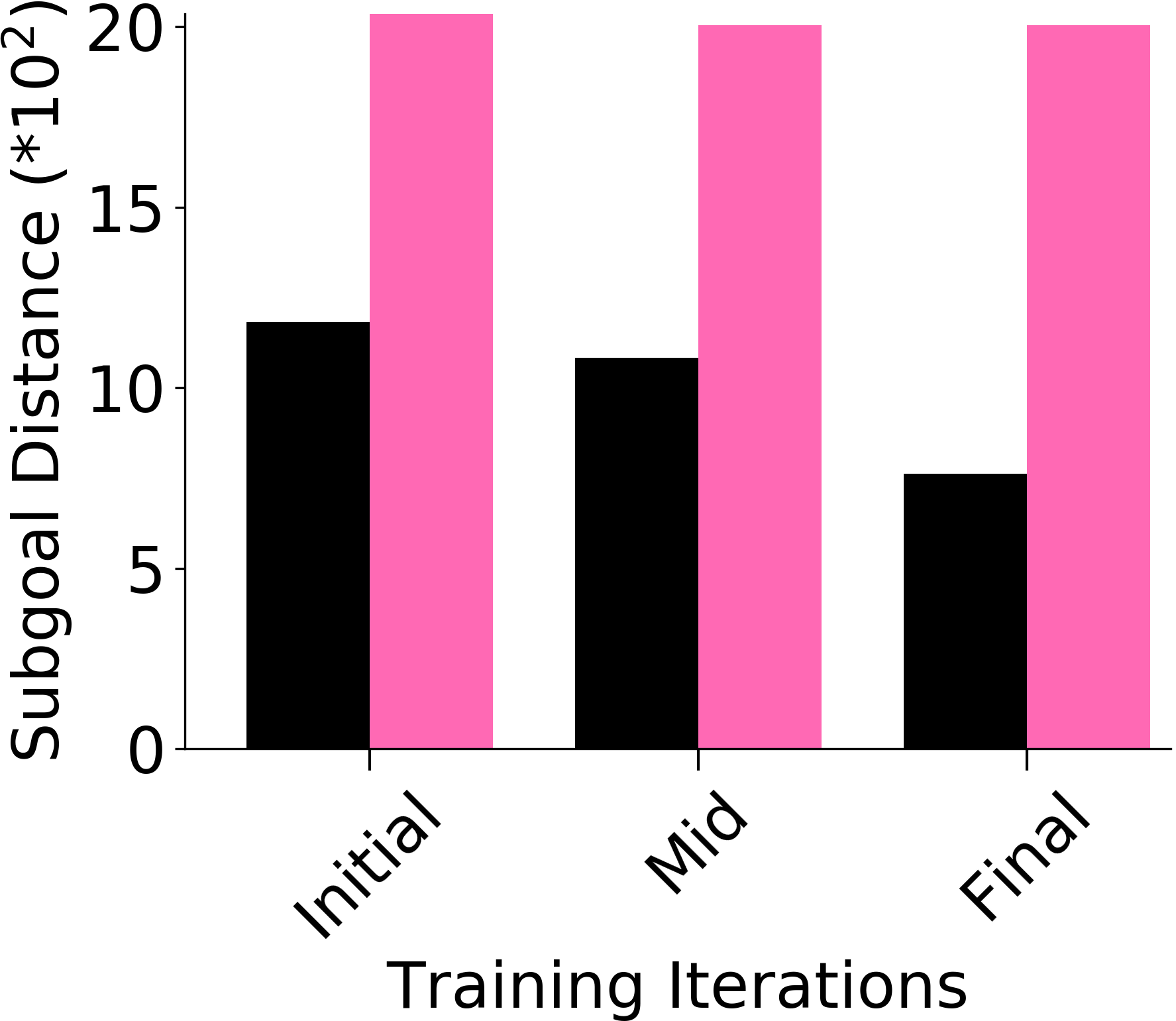}}\hspace{0.3cm}
\subfloat[][Franka kitchen]{\includegraphics[scale=0.27]{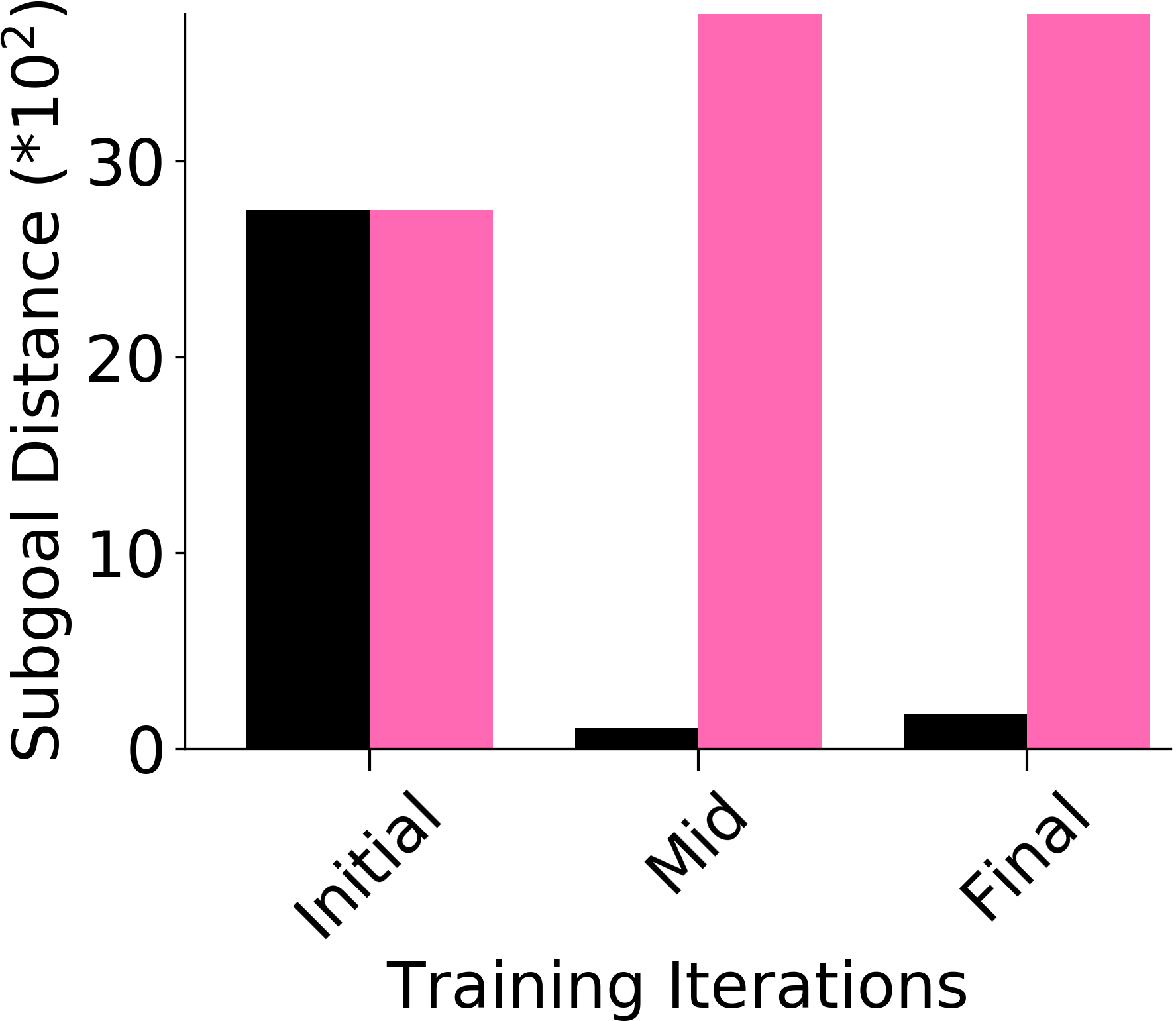}}
\\
{\includegraphics[scale=0.65]{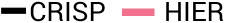}}
\caption{\textbf{Non-stationarity metric comparison} This figure compares the average distance metric between the subgoals predicted by the higher level policy and the states achieved by the lower level policy during training. (The columns represent Initial: when training begins, Mid: half-way during training, Final: when training ends, e.g. since maze navigation is trained for 4.7\textbf{E}6 timesteps, the values are Initial: iteration 1, Mid: iteration 2.35\textbf{E}6, and Final: iteration 4.7\textbf{E}6). As seen in figure, CRISP consistently produces efficient subgoals leading to low distances between the predicted and achieved subgoals throughout the training process. This mitigates non-stationarity in HRL.}
\label{fig:non_stationarity}
\end{figure*}

%% file: figures_tex/curriculum_figure.tex
\begin{figure*}[t]
\vspace{5pt}
\centering
% \captionsetup{font=footnotesize}
\includegraphics[scale=0.12]{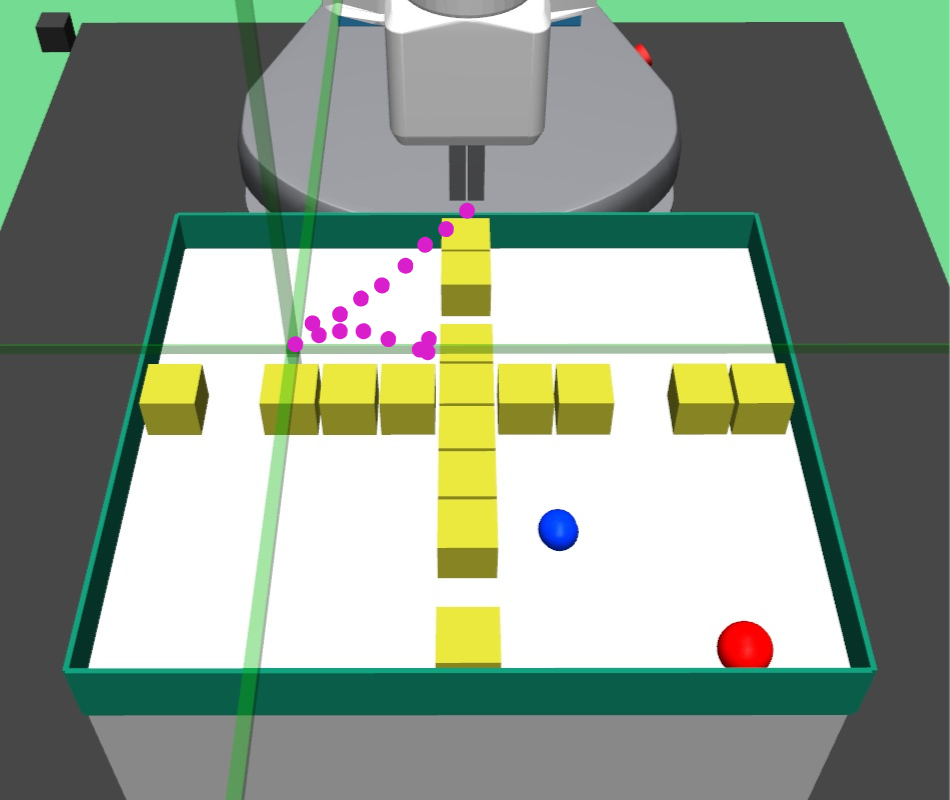}
\includegraphics[scale=0.12]{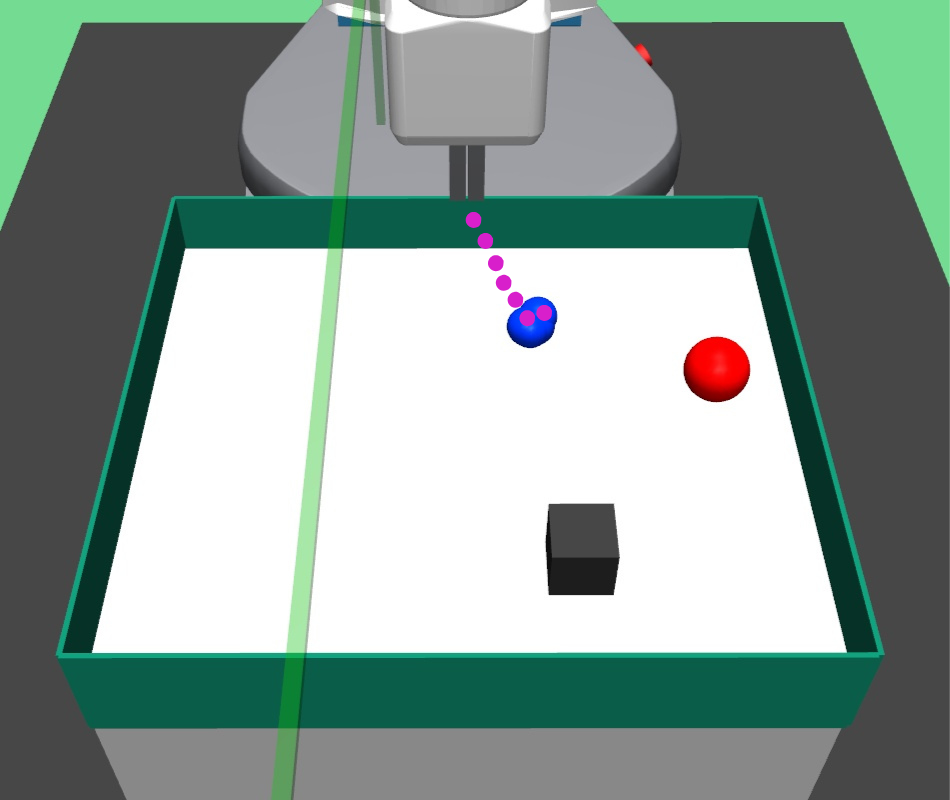}
\includegraphics[scale=0.12]{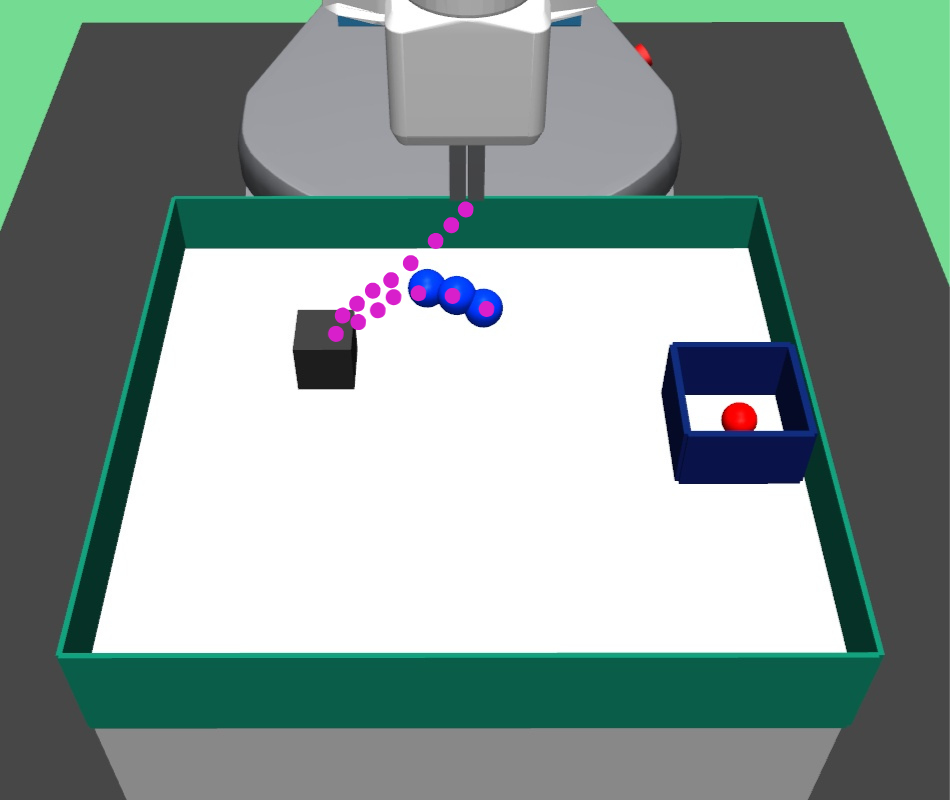}
\includegraphics[scale=0.12]{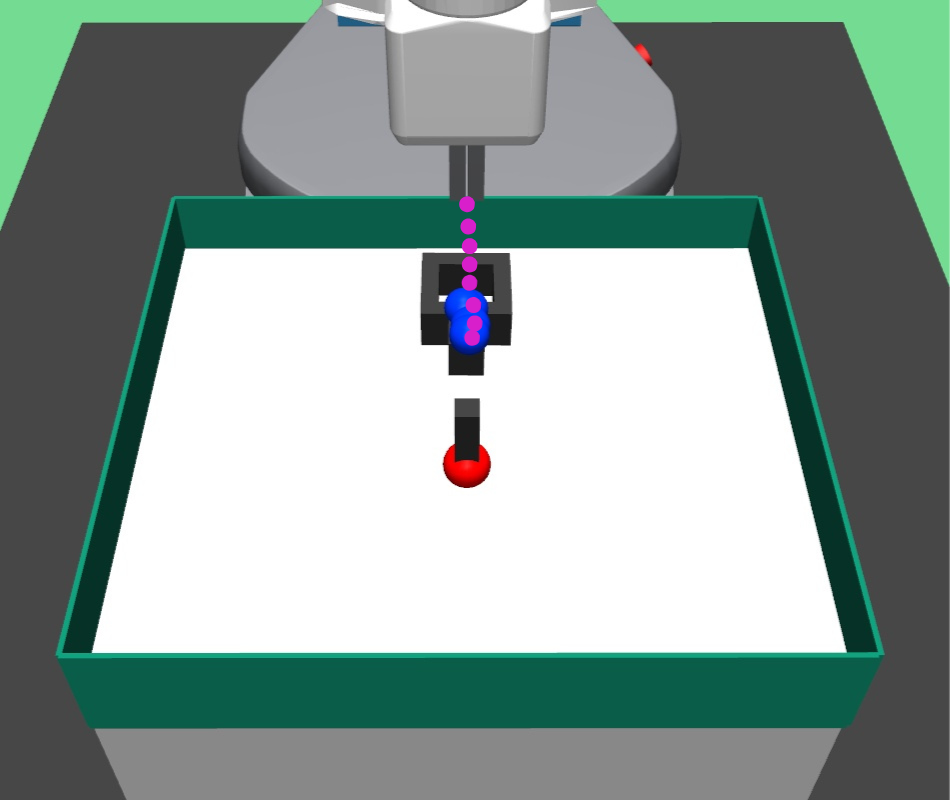}
\includegraphics[scale=0.12]{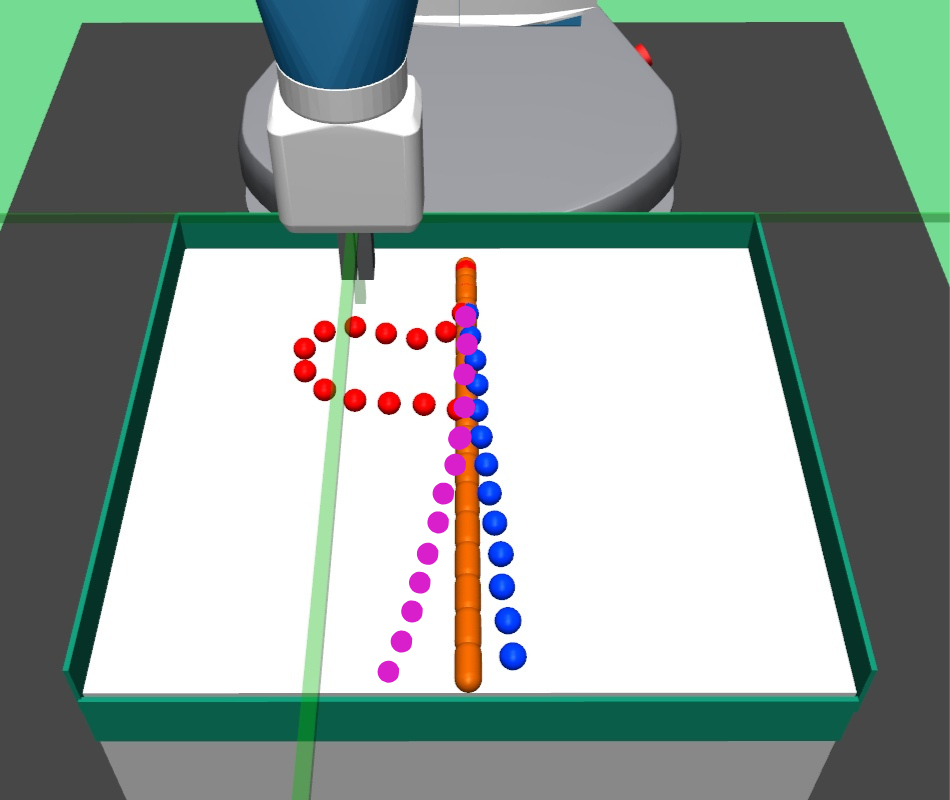}
\vspace{-0.2cm}
\\
\includegraphics[scale=0.12]{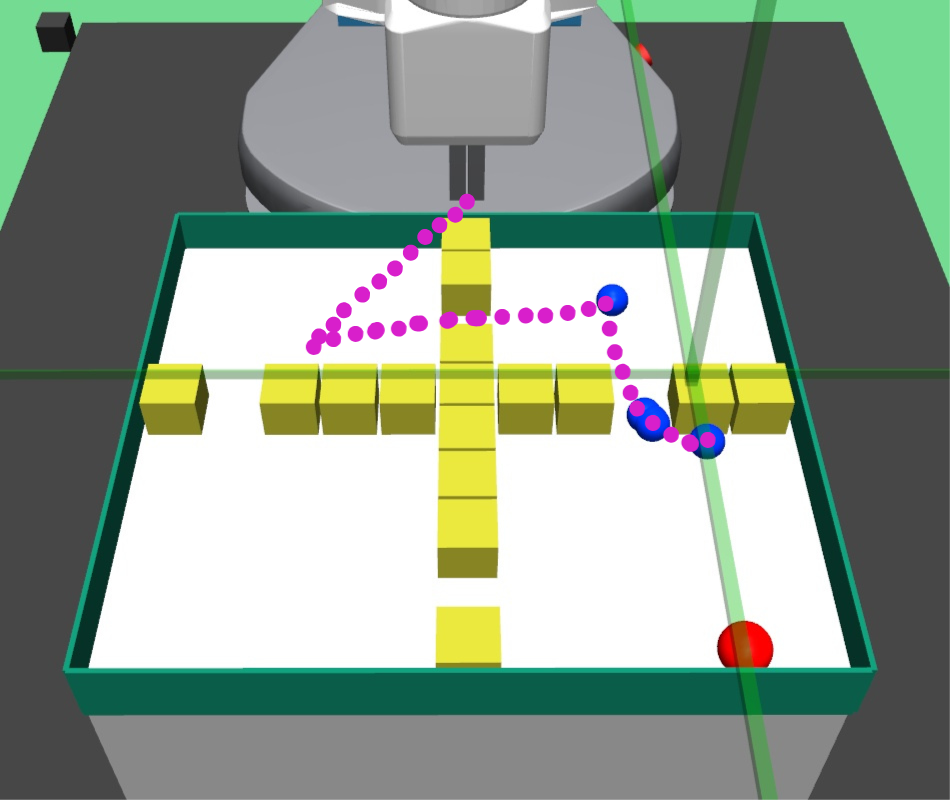}
\includegraphics[scale=0.12]{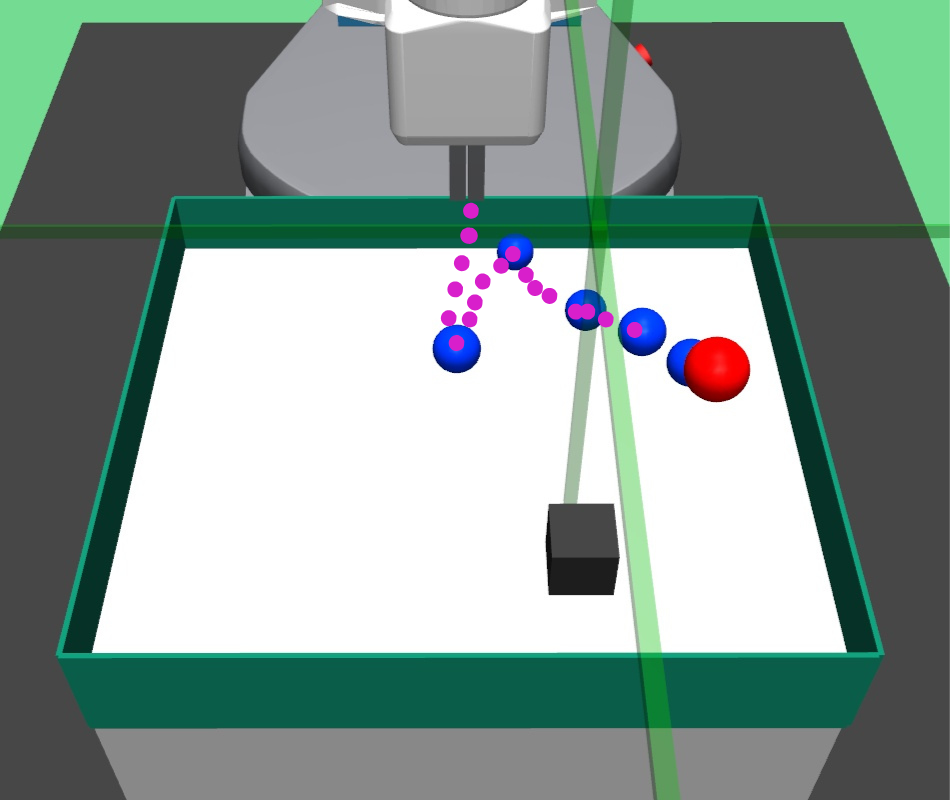}
\includegraphics[scale=0.12]{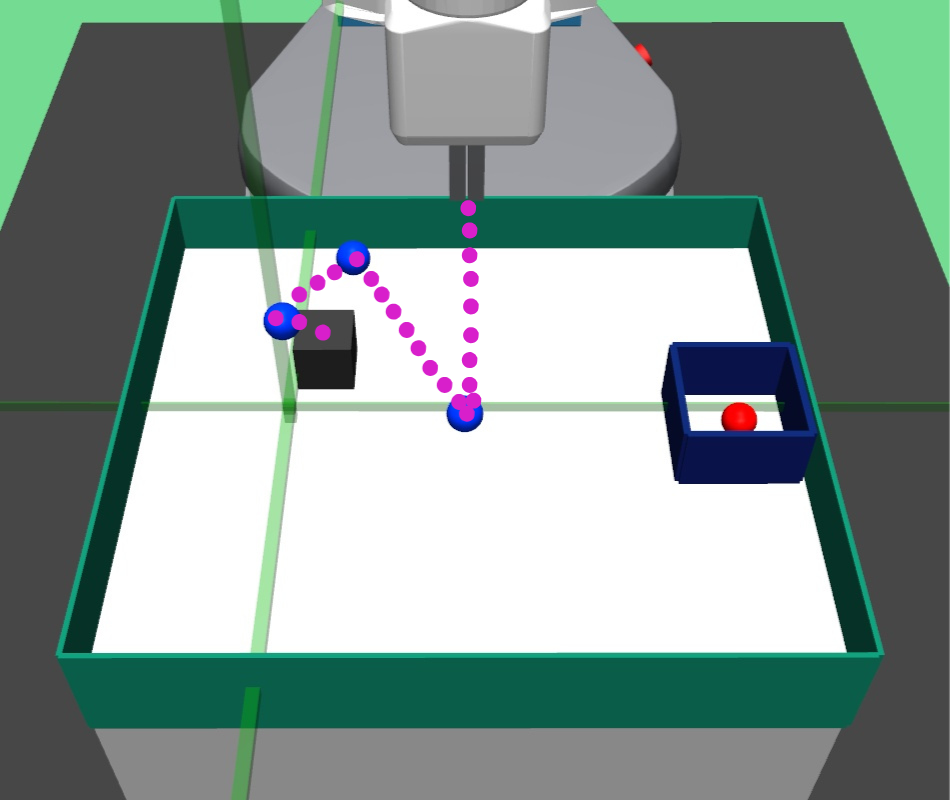}
\includegraphics[scale=0.12]{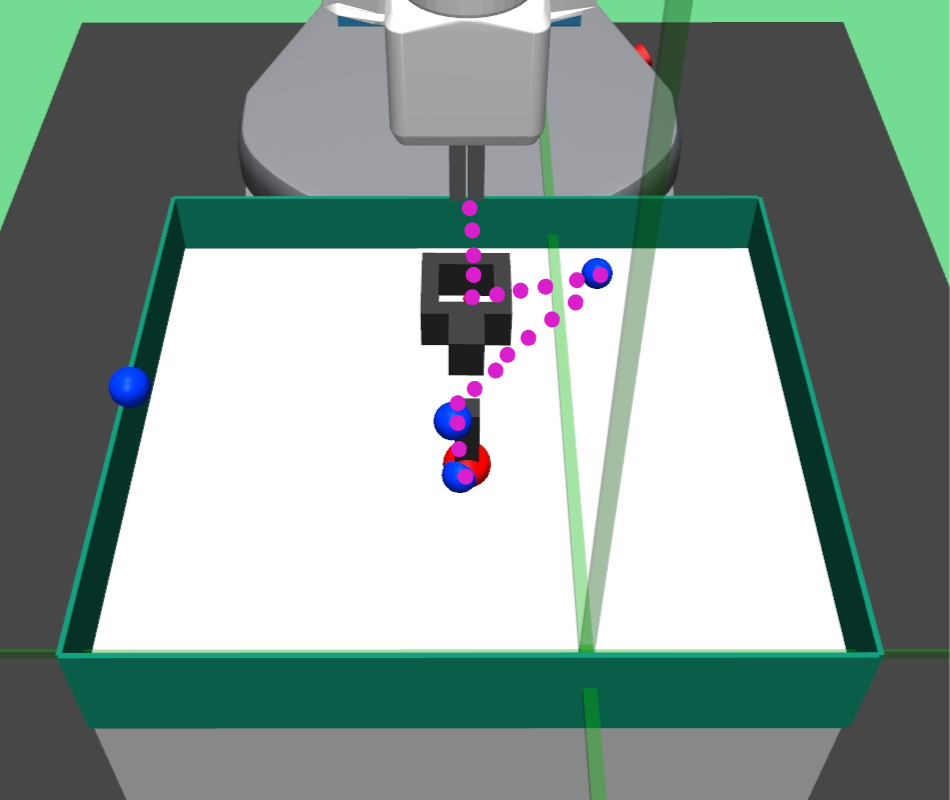}
\includegraphics[scale=0.12]{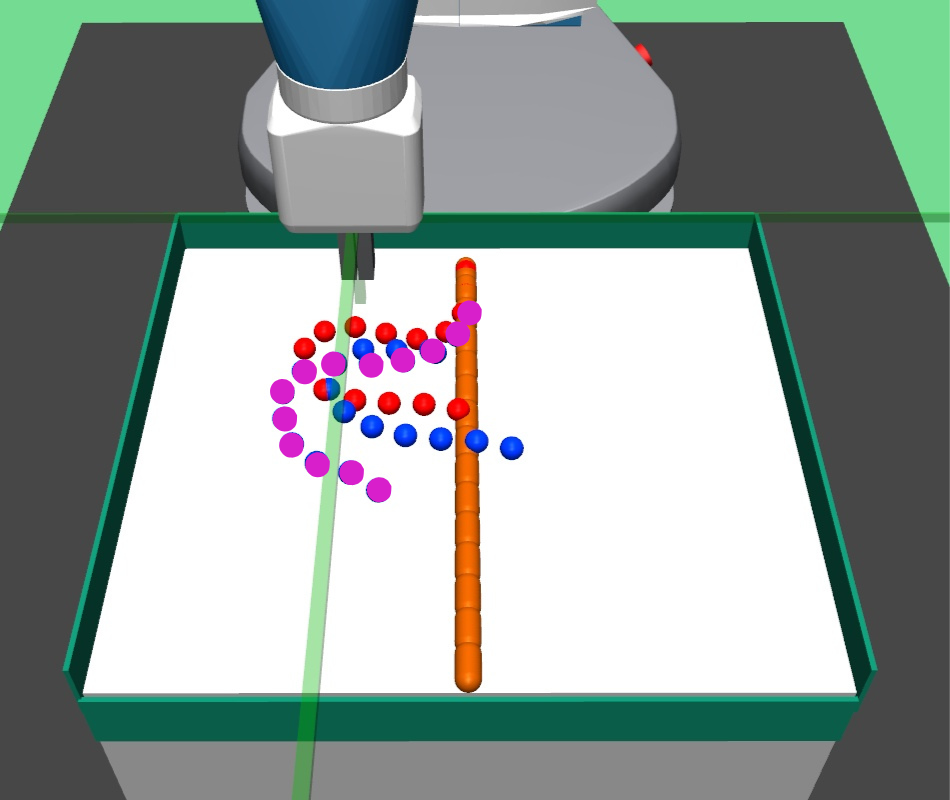}
\\
\vspace{-0.2cm}
\subfloat[][Maze]{\includegraphics[scale=0.12]{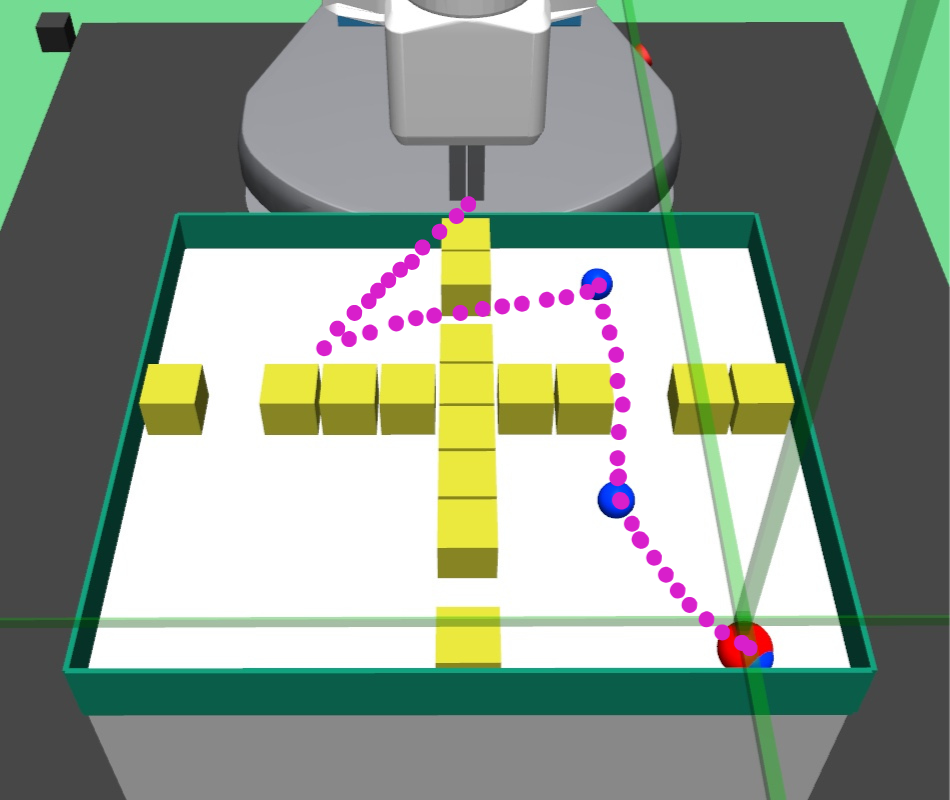}}
\hspace{0.01cm}
\subfloat[][Pick]{\includegraphics[scale=0.12]{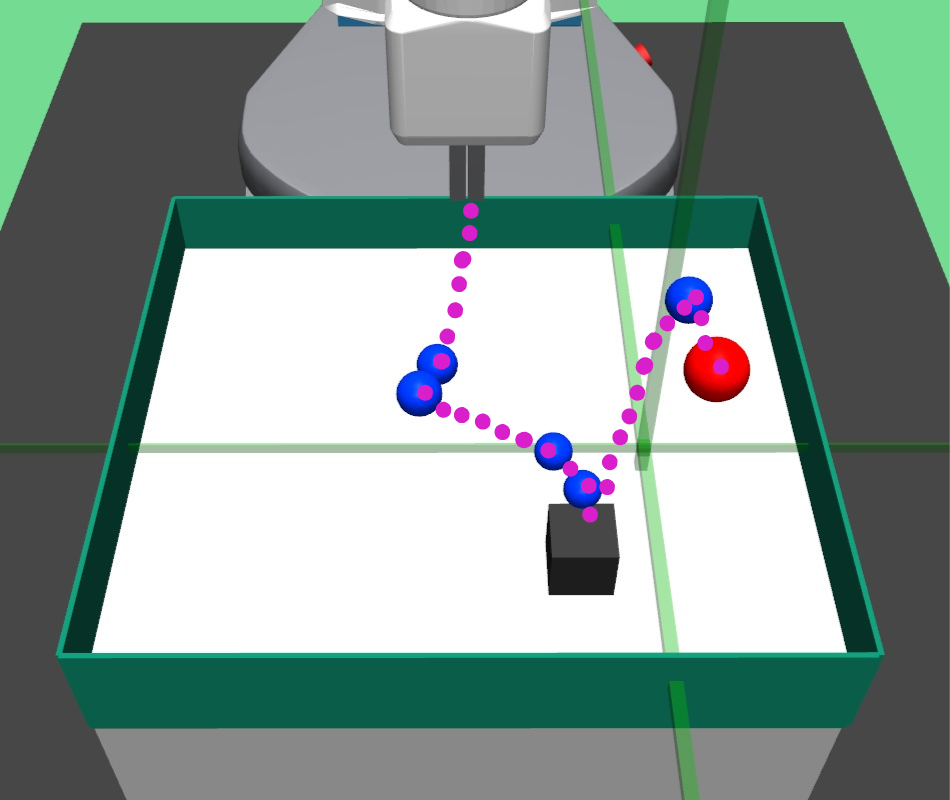}}
\hspace{0.01cm}
\subfloat[][Bin]{\includegraphics[scale=0.12]{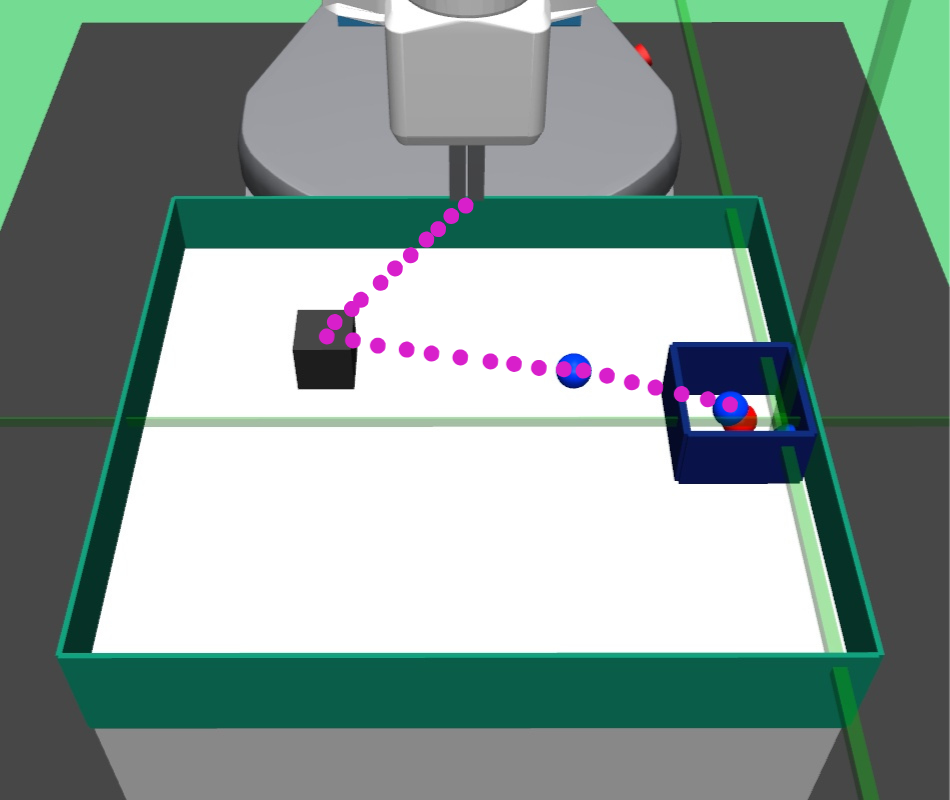}}
\hspace{0.01cm}
\subfloat[][Hollow]{\includegraphics[scale=0.12]{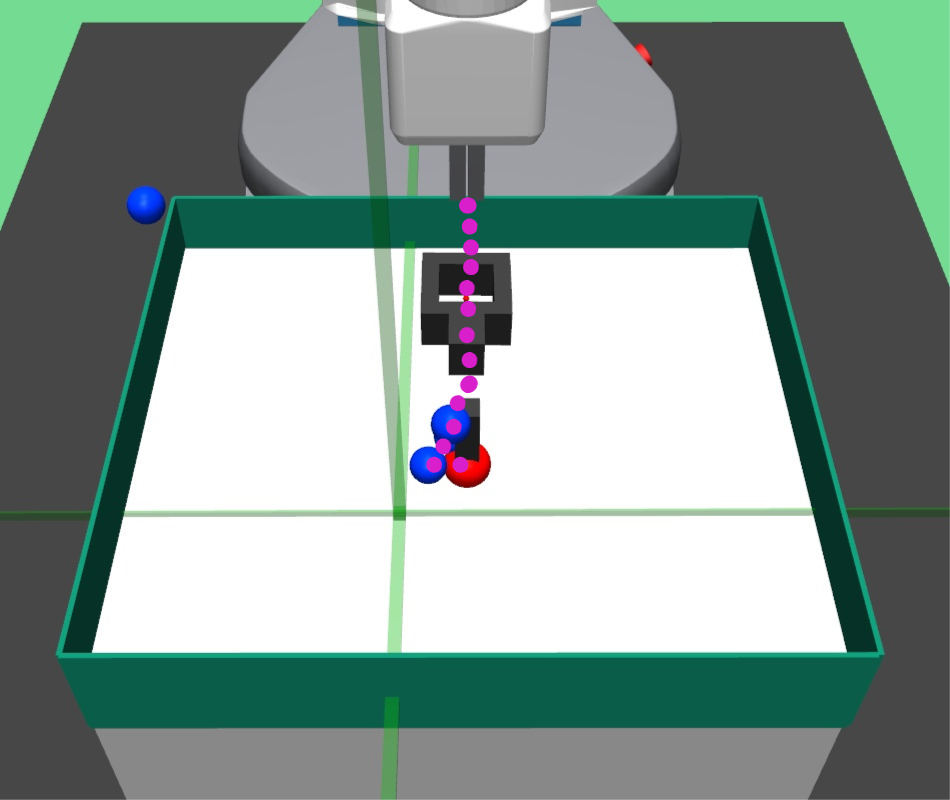}}
\hspace{0.01cm}
\subfloat[][Rope]{\includegraphics[scale=0.12]{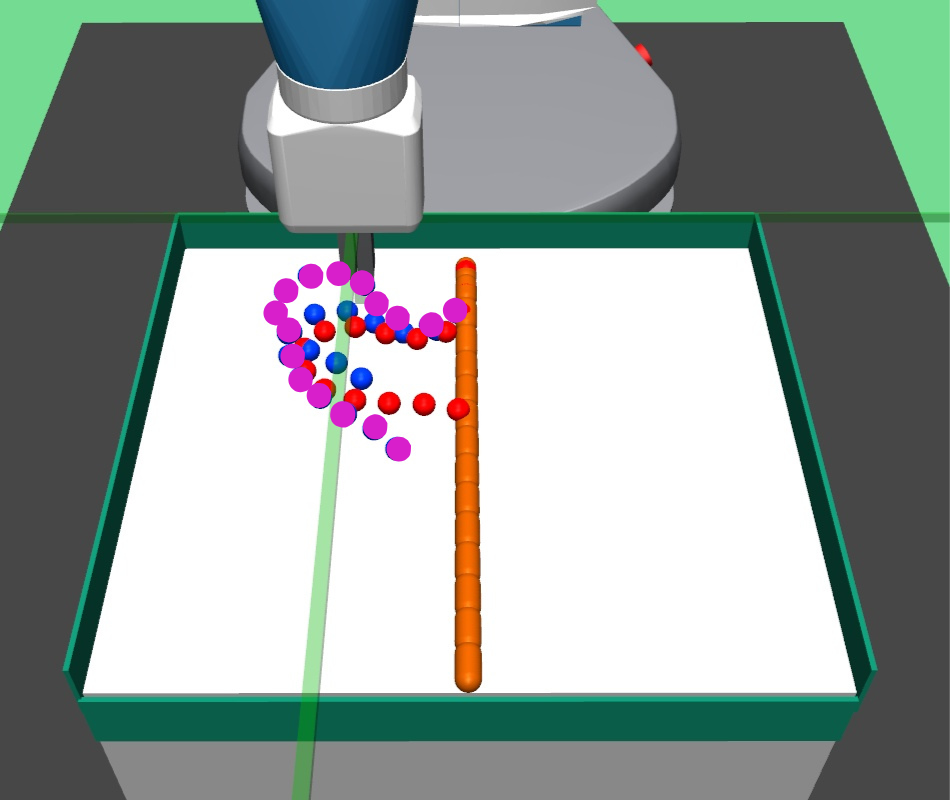}}

\caption{\textbf{Subgoal curriculum}: This figure depicts the progression of CRISP subgoal generation with training phases (Row 1: initial training, Row 2: mid training, Row 3: end training) in maze navigation, pick and place, bin, hollow, rope and kitchen environment, (walls:yellow, final goal:red, subgoals:blue, and the trajectories are shown in pink). As seen from figure, the subgoals get progressively harder, generating a curriculum for training higher level policy.}
\label{fig:env_curriculum}
\end{figure*}

%% file: sections/discussion.tex
\section{Discussion}
\label{sec:discussion1}
\textbf{Limitations.} Although primitive-informed parsing introduces some additional overhead, especially in long-horizon tasks, our empirical results indicate that this cost remains marginal. In future, we plan to investigate the use of undirected demonstrations to further enhance performance. Further, we recognize that CRISP uses environment resets to parse expert demonstrations (although CRISP does not require such resets during training), which is often impractical or costly in real-world environments without simulations. One promising solution is to integrate a learned reset controller, such as backward policies that return the system to previous states, allowing CRISP to function in the absence of manual resets~\cite{eysenbach2017leave}. Future work could also generate curricula using only states naturally visited during training or by dynamically identifying subgoals reachable from the current policy’s exploration, thus eliminating external resets.
\\
\\
\textbf{Conclusion.} We present CRISP, a versatile hierarchical curriculum based framework specifically designed to address non-stationarity in HRL. Leveraging our PIP approach to parse expert demonstrations, CRISP constructs a curriculum of achievable subgoals tailored for the low-level policy. We evaluate CRISP on difficult, sparsely rewarded robotic navigation and manipulation tasks in simulation, demonstrating substantial performance gains over existing baselines. In addition, we show that CRISP delivers strong results on real-world robotic tasks. Together, these findings underscore the promise of hierarchical curriculum learning as a pathway toward more capable and robust real-world robotic systems.

%% file: sections/appendix.tex
\onecolumn 
\tableofcontents

\section{Appendix}

\subsection{Impact Statement}
Our proposed approach and algorithm are not anticipated to result in immediate technological advancements. Rather, our main contributions are conceptual, emphasizing the foundational elements of Hierarchical Reinforcement Learning (HRL). This conceptual groundwork lays the foundation for future studies and could drive progress in HRL and associated fields.

\subsection{{Additional Ablation Experiments}}
\label{sec:appendix_ablations}
In this section we provide the ablation experiments in all six task environments. The ablation analysis includes comparison between CRISP-IRL, CRISP-BC, and CRISP-RPL ablation, performance comparison of $p$ hyper-parameter, $\psi$ hyper-parameter, RPL window size hyperparameter (Figure~\ref{fig:rpl_ablation}), and comparisons with varying number of expert demonstrations(Figure~\ref{fig:demos_ablation}) used during relabeling and training.
\input{figures_tex/rpl_irl_ablation}
\input{figures_tex/p_ablation}

\input{figures_tex/psi_ablation}

\input{figures_tex/rpl_ablation}

\input{figures_tex/demos_ablation}

\subsection{{Real Robot Experiments}}
\label{sec:real_robot}
In this section we provide the figure for the environments used for deploying the policies in real world robotic tasks. The environments used are: $(a)$ pick and place, $(b)$ bin, and $(c)$ rope manipulation environments.
\input{figures_tex/dobot_real}

\subsection{Additional hyper-parameters}
\label{appendix:hyperparameters}
Here, we enlist the additional hyper-parameters used in PIPER:

\begin{table}[ht]
\centering
\begin{tabular}{|l|l|p{8cm}|}
\hline
\textbf{Parameter} & \textbf{Value} & \textbf{Description} \\
\hline
activation       & tanh         & Activation function for hierarchical policies \\
layers           & 3            & Number of layers in the critic/actor networks \\
hidden           & 512          & Number of neurons in each hidden layer \\
Q\_lr            & 0.001        & Critic learning rate \\
pi\_lr           & 0.001        & Actor learning rate \\
buffer\_size     & int($1\times10^{7}$) & Size of experience replay buffer \\
tau              & 0.8          & Polyak averaging coefficient \\
clip\_obs        & 200          & Observation clipping value \\
n\_cycles        & 1            & Number of cycles per epoch \\
n\_batches       & 10           & Training batches per cycle \\
batch\_size      & 1024         & Batch size hyper-parameter \\
random\_eps      & 0.2          & Percentage of time a random action is taken \\
alpha            & 0.05         & Weight parameter for SAC \\
noise\_eps       & 0.05         & Std of Gaussian noise added to nearly random actions \\
norm\_eps        & 0.01         & Epsilon used for observation normalization \\
norm\_clip       & 5            & Clipping value for normalized observations \\
adam\_beta1      & 0.9          & Beta 1 parameter for Adam optimizer \\
adam\_beta2      & 0.999        & Beta 2 parameter for Adam optimizer \\
\hline
\end{tabular}
\caption{Hyperparameters used in the experiments.}
\label{tab:hyperparameters}
\end{table}

\subsection{Implementation details}
\label{appendix:implementation_details}

Since the main focus of this work it to develop an efficient algorithm scalable to sparse reward scnarios, we implement the simulation environments designed in Mujoco as sparse reward environments, where the agent only gets a reward if it achieves the final goal. The robotic maze navigation, pick and place, bin, hollow and rope environments employ a $7$-DOF robotic arm gripper whereas the kitchen environment employs a $9$-DoF franka robot gripper. In the maze navigation task, the gripper (whose height is kept fixed at table height) has to navigate across randomly generated four room mazes (the wall and gate positions are randomly generated) to achieve final goal. In the pick and place task, the gripper has to pick a randomly placed square block and bring it to a randomly generated goal position. In bin environment, the gripper has to pick up the block and place it in a specific bin. In the hollow task, the gripper has to pick a square hollow block and place it across a fixed vertical pole on the table such that the block goes through the pole. In rope manipulation task, a deformable soft rope is kept on the table and the gripper performs pokes on the rope to nudge it towards the goal rope configuration. The rope manipulation task involves learning challenging dynamics and goes beyond prior work on navigation-like tasks where the goal space is limited. In kitchen task, the gripper has to first open the microwave door, and then switch on the specific gas knob where the kettle is placed. 

\par \textbf{Hyper-parameter and training details} 
% In our experiments, the regularization weight hyper-parameter $\Psi$ is set to $1e-3$, $5e-3$, $5e-3$, $5e-3$, $5e-3$, and $1e-3$ and population hyper-parameter $p$ is set to be $1.1e4$, $2500$, $2500$, $2500$, $3.9e5$, and $1.4e4$ for environments $(i)-(vi)$ respectively. 
% The maximum task horizon $T$ is kept at $225$, $50$, $60$, $100$ $25$, and $280$ timesteps, the lower primitive is allowed to execute for $15$, $7$, $6$, $10$, $5$, and $17$ timesteps, and the experiments are run for $c$ $4.73e5$, $1.1e5$, $1.32E5$, $3.5E5$, $1.58e6$, and $5.32e5$ timesteps in environments $(i)-(vi)$ tasks respectively. 
We collect $28$ expert demonstrations in kitchen and $100$ demonstrations in all other tasks. For collecting expert demonstrations, we use RRT based trajectories in maze task, Mujoco VR based direction in pick and place, bin and hollow tasks, poking based expert controller in rope task, and D4RL expert data in kitchen task. We provide more environments details and detailed expert demonstrations collection procedures for all tasks in Section~\ref{sec:appendix_env_details} and~\ref{sec:appendix_expert_demos} respectively.
\par \textbf{RAPS baseline setup} The hand designed action primitives from RAPS are designed as follows: $(i)$ in maze navigation, the lower level primitive travels in a straight line directly towards the subgoal predicted by higher level policy, $(ii)$ in pick and place, bin and hollow tasks, we hand-designed three primitives: \textit{gripper-reach} (where the gripper has to reach the goal position, \textit{gripper-open} (where the robotic arm has to open the gripper) and \textit{gripper-close} (where the robotic arm has to close the gripper). In kitchen environment, we use the action primitives implemented in RAPS. Since it is hard to design the action primitives in rope environment, we do not evaluate RAPS in the rope environment.

\subsection{{Environment details}}
\label{sec:environment_details}
In this section, we provide the environment and implementation details for all the tasks:

\subsubsection{{Maze navigation task}}
In this environment, a $7$-DOF robotic arm gripper navigates across random four room mazes to reach the goal position. The gripper arm is kept closed and fixed at table height, and the positions of walls and gates are randomly generated. The table is discretized into a rectangular $W*H$ grid, and the vertical and horizontal wall positions $W_{P}$ and $H_{P}$ are randomly picked from $(1,W-2)$ and $(1,H-2)$ respectively. In the four room environment thus constructed, the four gate positions are randomly picked from $(1,W_{P}-1)$, $(W_{P}+1,W-2)$, $(1,H_{P}-1)$ and $(H_{P}+1,H-2)$. 

\par In the maze environment, the state is represented as the vector $[dx,M]$, where $dx$ is current gripper position and $M$ is the sparse maze array. The higher level policy input is thus a concatenated vector $[dx,M,g]$, where $g$ is the target goal position, whereas the lower level policy input is concatenated vector $[dx,M,s_g]$, where $s_g$ is the sub-goal provided by the higher level policy. $M$ is a discrete $2D$ one-hot vector array, where $1$ represents presence of a wall block, and otherwise. The lower primitive action $a$ is a $4$ dimensional vector with every dimension $a_i \in [0,1]$. The first $3$ dimensions provide offsets to be scaled and added to gripper position for moving it to the intended position. The last dimension provides gripper control($0$ implies a closed gripper and $1$ implies an open gripper). We select $100$ randomly generated mazes each for training, testing and validation.

\subsubsection{{Pick and place, bin and hollow environments}}
In this section, we explain the environment details for the pick and place, bin and hollow tasks. The state is represented as the vector $[dx,o,q,e]$, where $dx$ is the current gripper position, $o$ is the position of the block object placed on the table, $q$ is the relative position of the block with respect to the gripper, and $e$ consists of linear and angular velocities of the gripper and the block object. The higher level policy input is thus a concatenated vector $[dx,o,q,e,g]$, where $g$ is the target goal position. The lower level policy input is concatenated vector $[dx,o,q,e,s_g]$, where $s_g$ is the sub-goal provided by the higher level policy. In our experiments, we keep the sizes of $dx$, $o$, $q$, $e$ to be $3$, $3$, $3$ and $11$ respectively. The lower primitive action $a$ is a $4$ dimensional vector with every dimension $a_i \in [0,1]$. The first $3$ dimensions provide gripper position offsets, and the last dimension provides gripper control. While training, the position of block object and goal are randomly generated (block is always initialized on the table, and goal is always above the table at a fixed height). We select $100$ random each for training, testing and validation.

\subsubsection{{Rope Manipulation Environment}}
In this environment, the deformable rope is formed from $15$ constituent cylinders joined together. The state space for the rope manipulation environment is a vector formed by concatenation of the intermediate joint positions. The upper level predicts subgoal $s_g$ for the lower primitive. The action space of the poke is $(dx, dy, \eta)$, where $(x, y)$ is the initial position of the poke, and $\eta$ is the angle describing the direction of the poke. We fix the poke length to be $0.08$. We select $100$ randomly generated initial and final rope configurations each for training, testing and validation.

\subsubsection{{Franka kitchen Environment}}
For this environment please refer to the D4RL environment. In this environment, the franka robot has to perform a complex multi-stage task in order to achieve the final goal.

\subsection{{Generating expert demonstrations}}
\label{sec:appendix_expert_demos}
We explain the procedure for geenrating expert demonstrations as follows:
\subsubsection{{Maze navigation Environment}}
\label{appendix_maze_expert}
We use the path planning RRT algorithm to generate optimal paths $P=(p_t, p_{t+1}, p_{t+2},...p_{n})$ from the initial state to the goal state. Using these expert paths, we generate state-action expert demonstration dataset for the lower level policy, which is later used to generate subgoal transition dataset. Since the procedure is automated using RRT algorithm, we can generate expert demonstrations without the expert.

\subsubsection{{Pick and place Environment}}
\label{appendix_pick_and_place_expert}
For generating expert demonstrations, we initially used a human agent in virtual reality based Mujoco simulation to generate demonstrations. We later found that hard coding a control policy also works reasonably well in this environment. Hence, we used a hard-coded policy to generate the expert demonstrations. In this task, the robot firstly picks up the block using robotic gripper, and then takes it to the target goal position.

\subsubsection{{Bin Environment}}
\label{appendix_bin_expert}
In this environment, we used a hard-coded policy to generate the expert demonstrations. In this task, the robot firstly picks up the block using robotic gripper, and then places it in the target bin. Using these expert trajectories, we generate expert demonstration dataset for the lower level policy.

\subsubsection{{Hollow Environment}}
\label{appendix_hollow_expert}
In this environment, we used a hard-coded policy to generate the expert demonstrations. In this task, the robotic gripper has to pick up the square hollow block and place it such that a vertical structure on the table goes through the hollow block.

\subsubsection{{Rope Manipulation Environment}}
We hand coded an expert policy to automatically generate expert demonstrations $e=(s^e_0, s^e_1, \ldots, s^e_{T-1})$, where $s^e_i$ are demonstration states. The states $s^e_i$ here are rope configuration vectors. The expert policy is explained below.
\par Let the starting and goal rope configurations be $sc$ and $gc$. We find the cylinder position pair $(sc_m, gc_m)$ where $m \in [1,n]$, such that $sc_m$ and $gc_m$ are farthest from each other among all other cylinder pairs. Then, we perform a poke $(x,y,\theta)$ to drag $sc_m$ towards $gc_m$. The $(x,y)$ position of the poke is kept close to $sc_m$, and poke direction $\theta$ is the direction from $sc_m$ towards $gc_m$. After the poke execution, the next pair of farthest cylinder pair is again selected and another poke is executed. This is repeatedly done for $k$ pokes, until either the rope configuration $sc$ comes within $\delta$ distance of goal $gc$, or we reach maximum episode horizon $T$. Although, this policy is not the perfect policy for goal based rope manipulation, but it still is a good expert policy for collecting demonstrations $D$. Moreover, as our method requires states and not primitive actions (pokes), we can use these demonstrations $D$ to collect good higher level subgoal dataset $D_g$ using primitive parsing.

\subsubsection{{Kitchen Environment}}
In this environment, we used the expert demonstrations provided in D4RL dataset. We use directed demonstrations from this dataset to solve the multi-stage task in this environment.

\subsection{{Qualitative visualizations}}
\label{sec:appendix_qualitative_viz}
In this section, we provide some visualizations in various environments:

% \subsubsection{Maze Navigation environment}
\input{figures_tex/maze_success_visualization_1}

% \input{figures_tex/maze_success_visualization_2}
% \input{figures_tex/maze_failure_visualization_1}
% \input{figures_tex/maze_failure_visualization_2}

% \subsubsection{Pick and place environment}
\input{figures_tex/pick_success_visualization_1}

% \input{figures_tex/pick_success_visualization_2}

% \subsubsection{Bin environment}
\input{figures_tex/bin_success_visualization_1}

% \input{figures_tex/bin_success_visualization_2}

% \subsubsection{Hollow environment}
\input{figures_tex/hollow_success_visualization_1}

% \input{figures_tex/hollow_failure_visualization_1}

% \subsubsection{Rope Manipulation environment}
\input{figures_tex/rope_success_visualization_1}

% \input{figures_tex/rope_success_visualization_2}
% \input{figures_tex/rope_failure_visualization_1}
% \input{figures_tex/rope_failure_visualization_2}

% \subsubsection{Kitchen environment}
\input{figures_tex/kitchen_success_visualization_1}

%% file: figures_tex/rpl_irl_ablation.tex
\begin{figure}[H]
\vspace{5pt}
\centering
    % \captionsetup{font=footnotesize}
    \subfloat[][Maze navigation]{\includegraphics[scale=0.27]{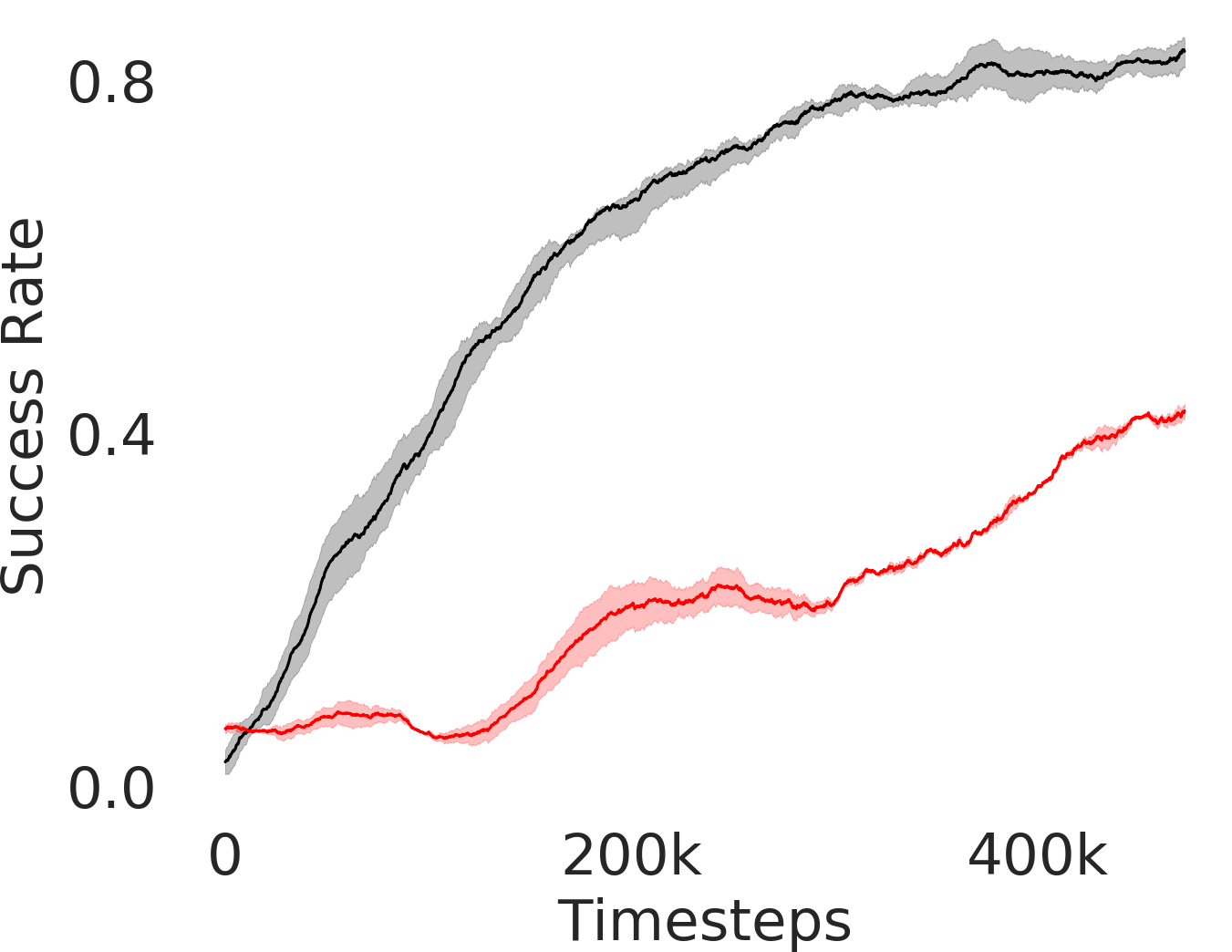}}
    \subfloat[][Pick and place]{\includegraphics[scale=0.27]{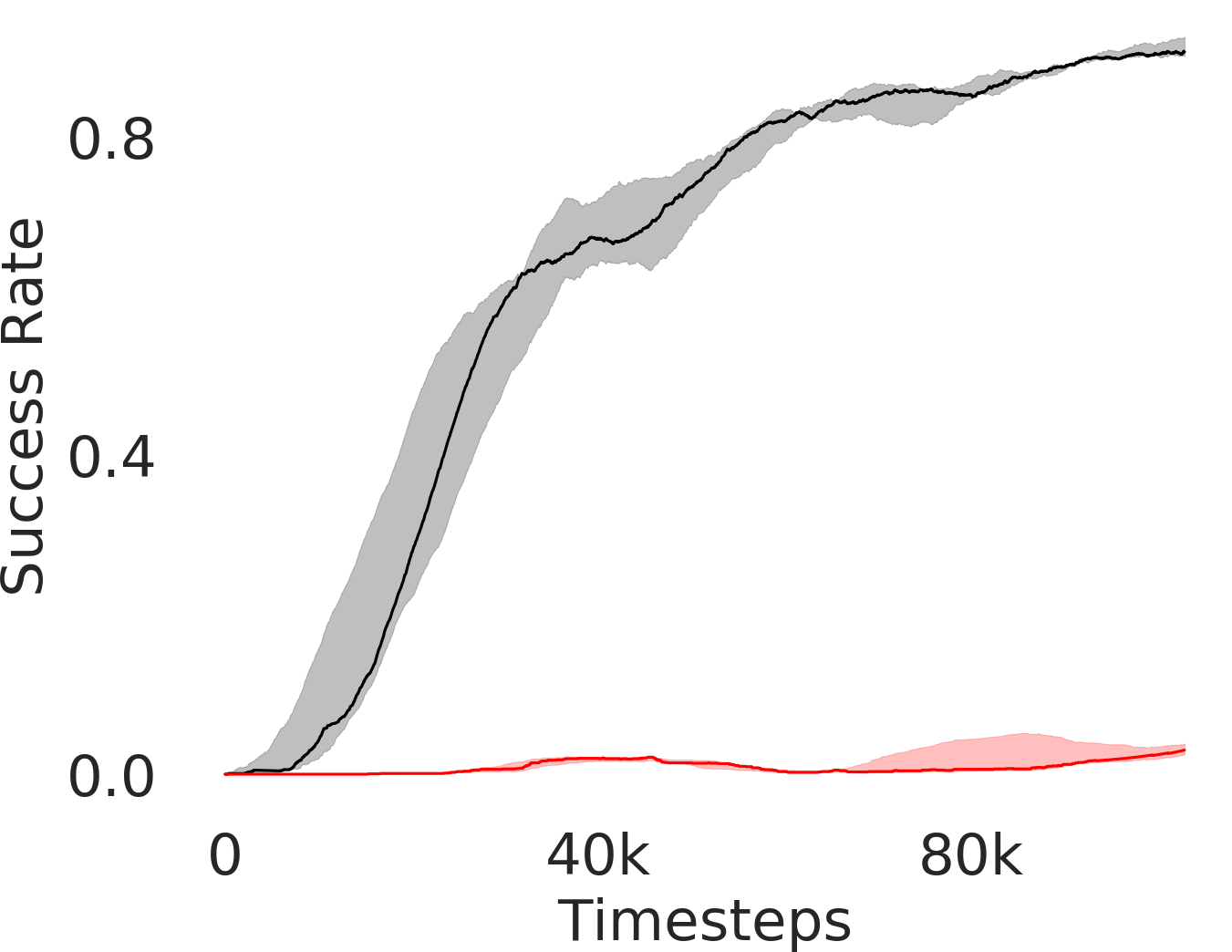}}
    \subfloat[][Bin]{\includegraphics[scale=0.27]{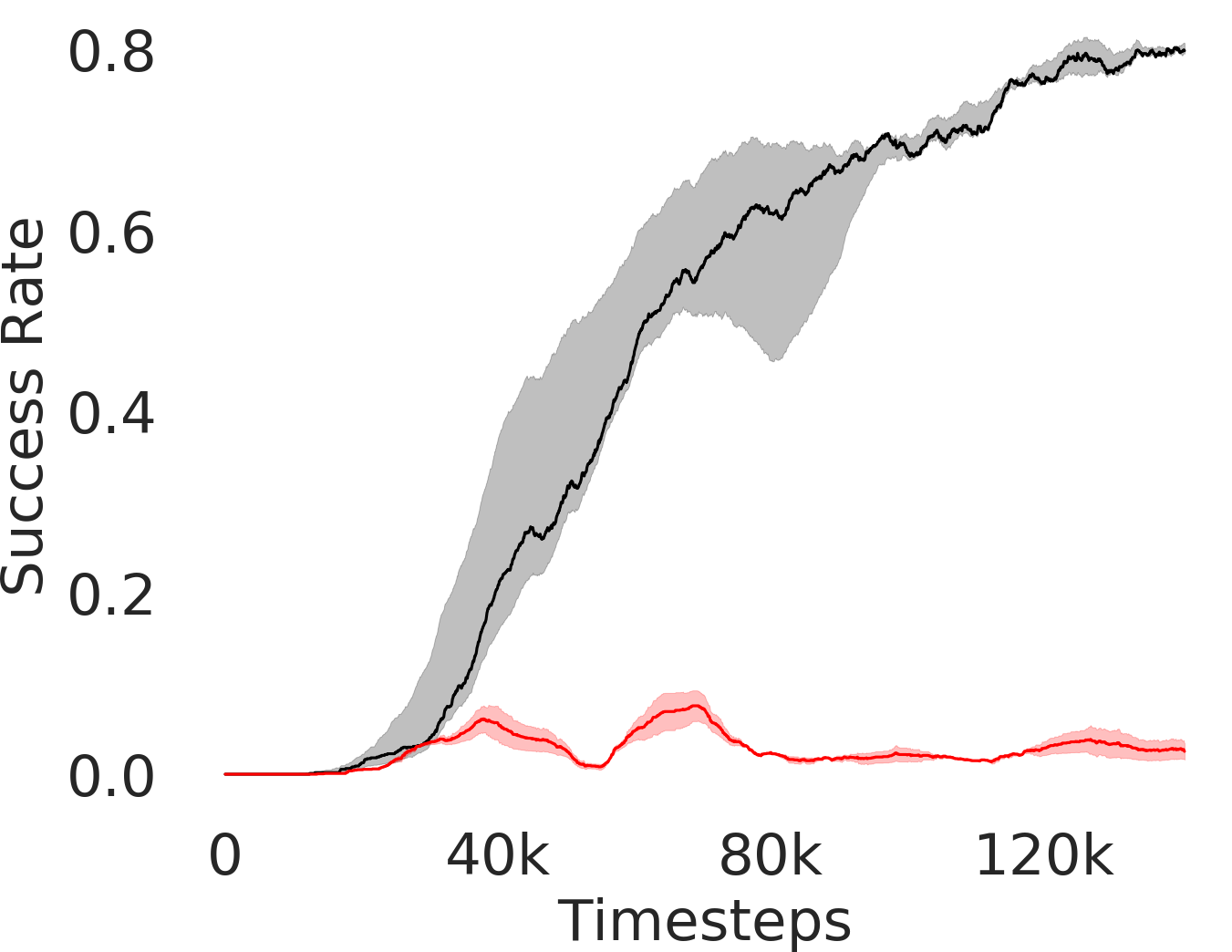}}
    \\
    \subfloat[][Hollow]{\includegraphics[scale=0.27]{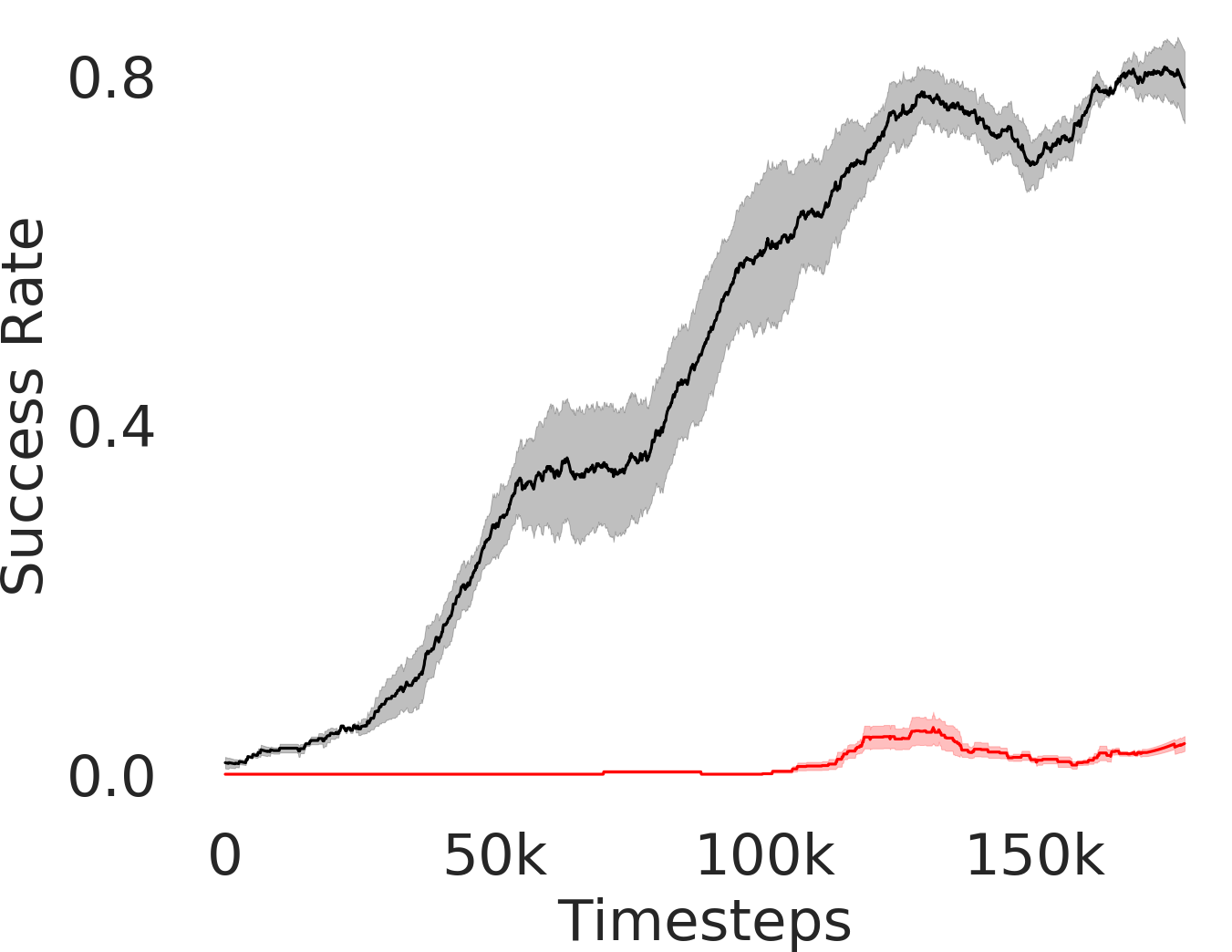}}
    \subfloat[][Rope]{\includegraphics[scale=0.27]{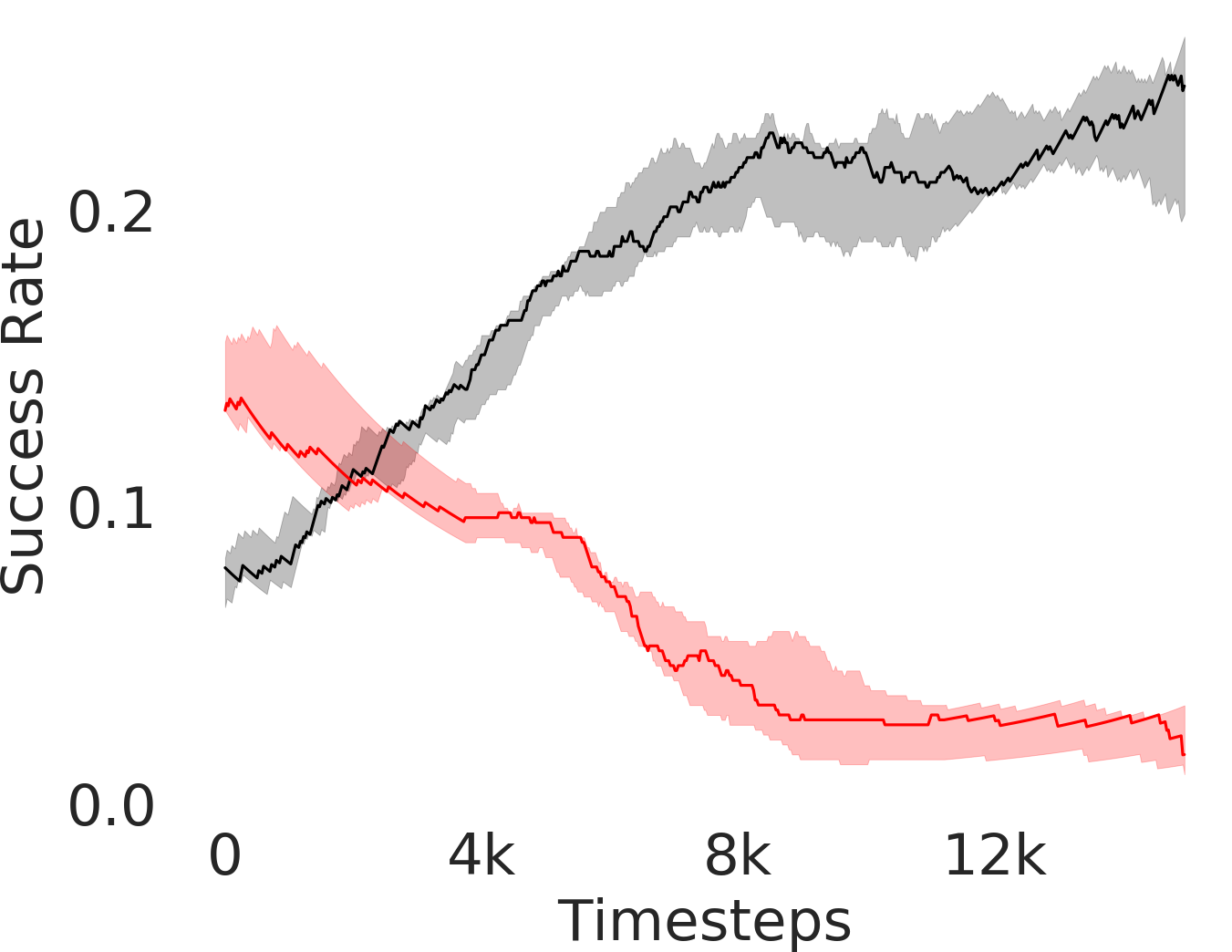}}
    \subfloat[][Franka kitchen]{\includegraphics[scale=0.27]{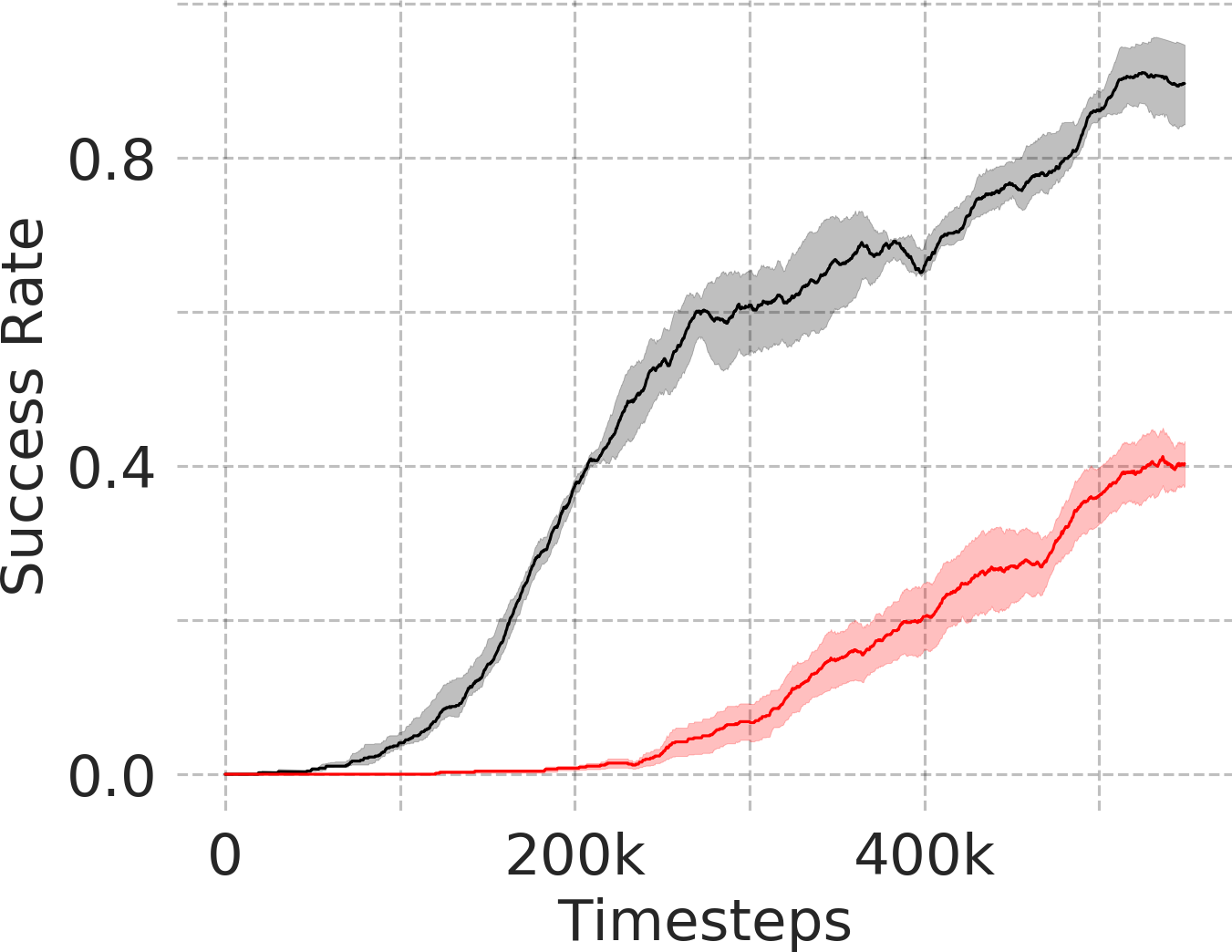}}
    \caption{\textbf{Comparison with CRISP-RPL ablation} This figure depicts comparison between CRISP-IRL and CRISP-RPL ablation. CRISP-IRL consistently outperforms CRISP-RPL, showing that proposed primitive informed parsing is crucial for improved performance.}
    \label{fig:rpl_rpl_ablation}
\label{fig:env_curriculum}
\end{figure}

%% file: figures_tex/p_ablation.tex
\begin{figure}[H]
\vspace{1pt}
\centering
% \captionsetup{font=footnotesize,labelfont=scriptsize,textfont=scriptsize}
% \captionsetup{font=footnotesize}
\subfloat[][Maze navigation]{\includegraphics[scale=0.25]{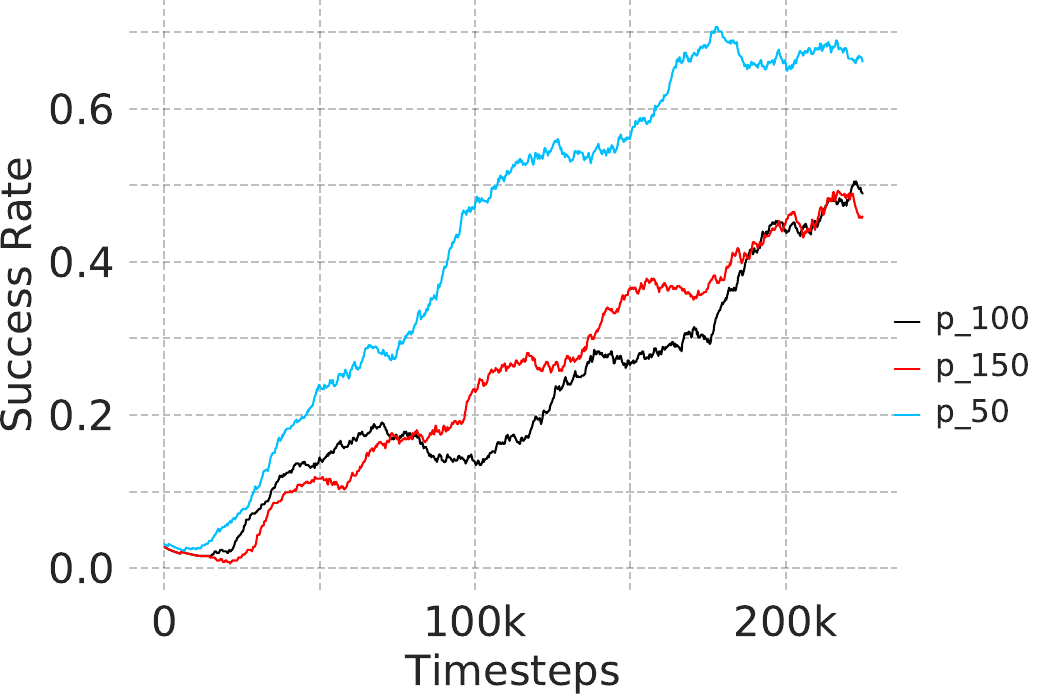}}
\subfloat[][Pick and place]{\includegraphics[scale=0.25]{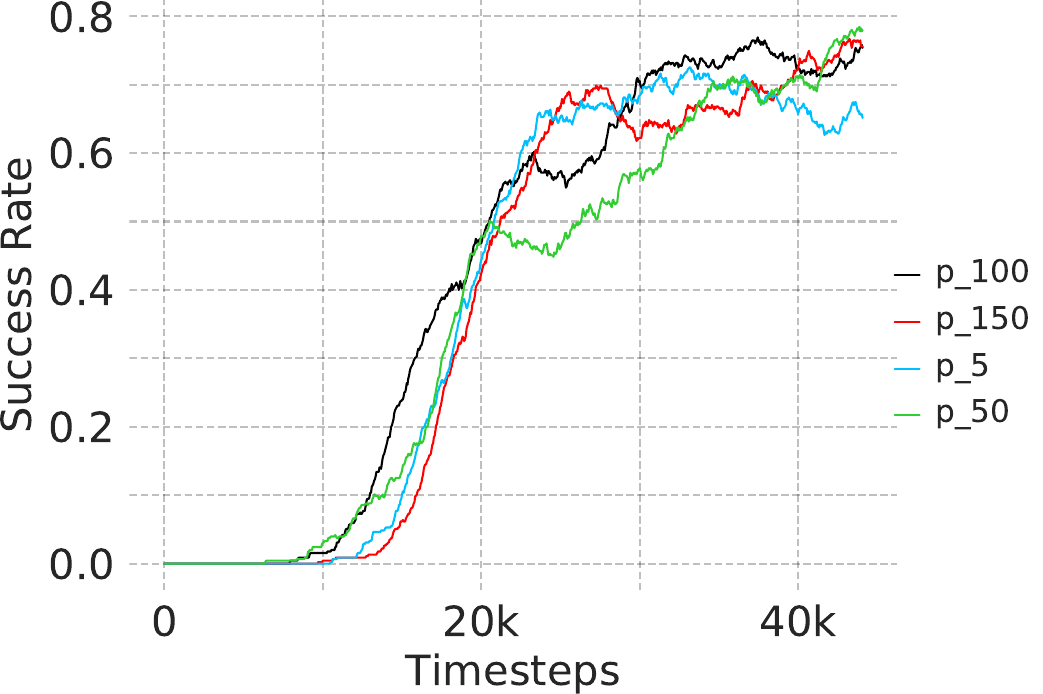}}
\subfloat[][Bin]{\includegraphics[scale=0.25]{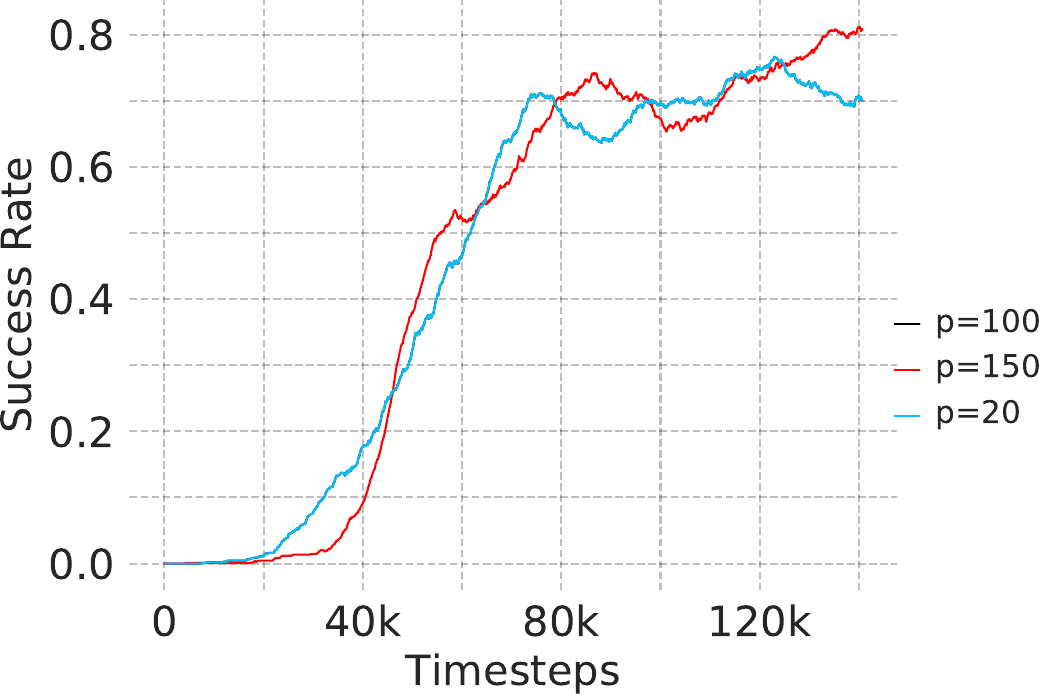}}
\\
\subfloat[][Hollow]{\includegraphics[scale=0.25]{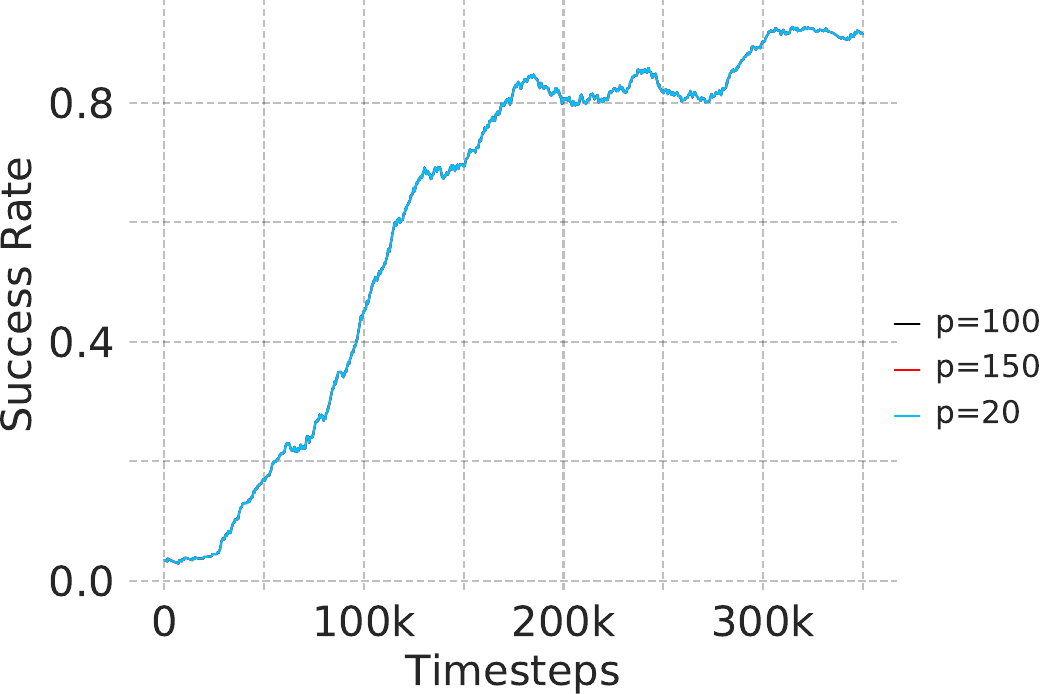}}
\subfloat[][Rope]{\includegraphics[scale=0.25]{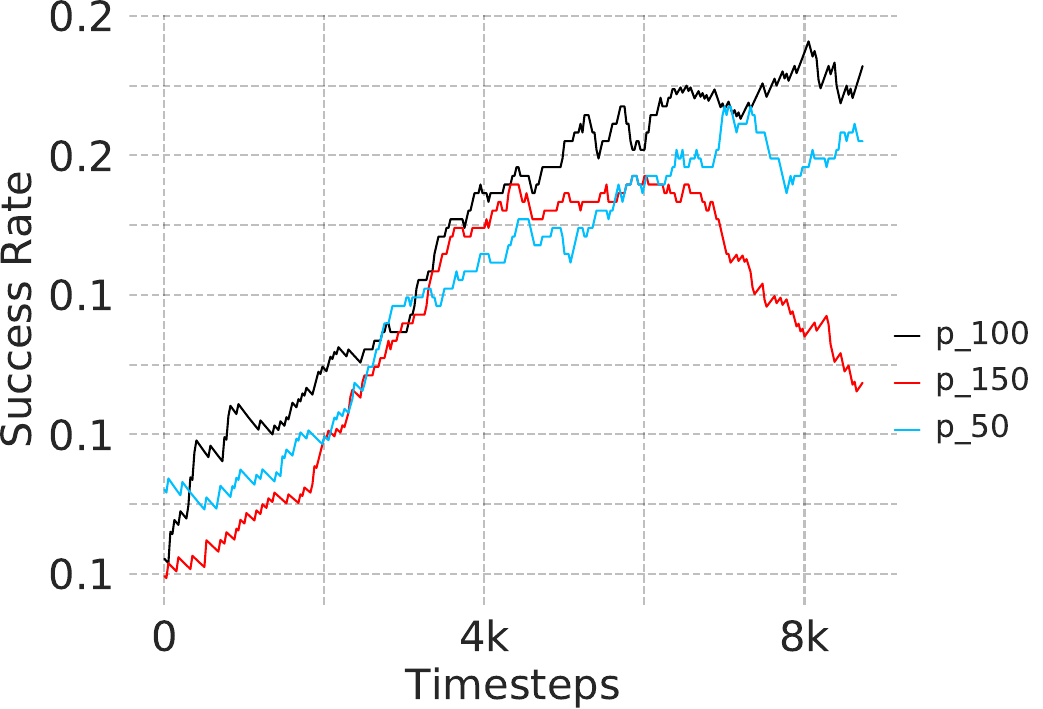}}
\subfloat[][Franka kitchen]{\includegraphics[scale=0.25]{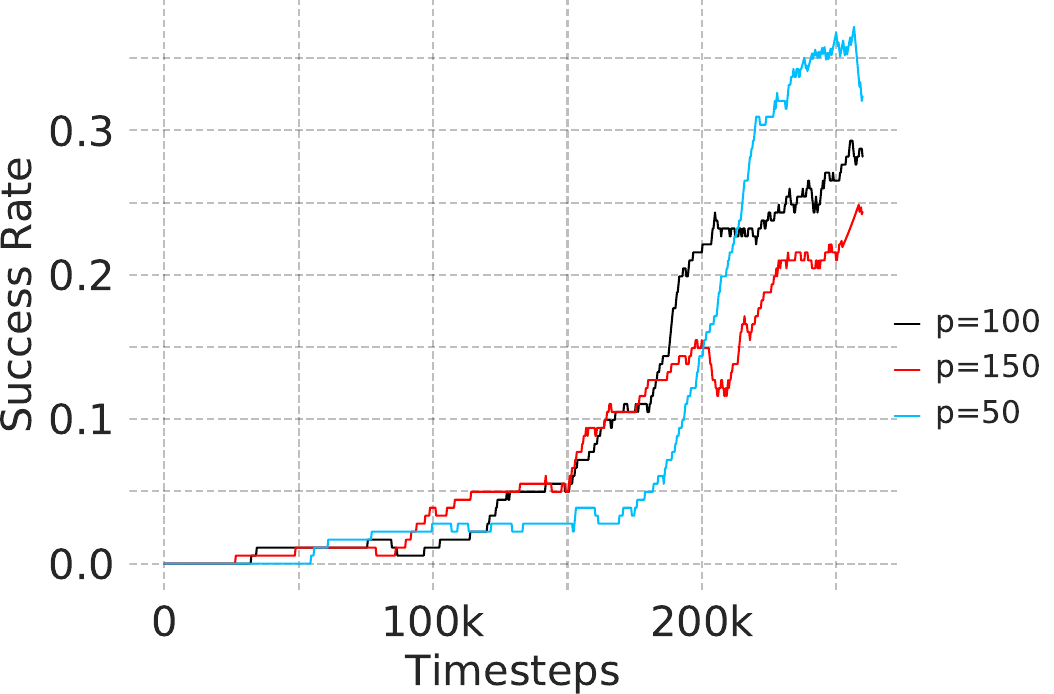}}

\caption{\textbf{$p$ ablation} This figure depicts comparisons for various $p$ values in multiple environments}
\label{fig:p_ablation}
\end{figure}

%% file: figures_tex/psi_ablation.tex
\begin{figure}[H]
\vspace{1pt}
\centering
% \captionsetup{font=footnotesize}
\subfloat[][Maze navigation]{\includegraphics[scale=0.25]{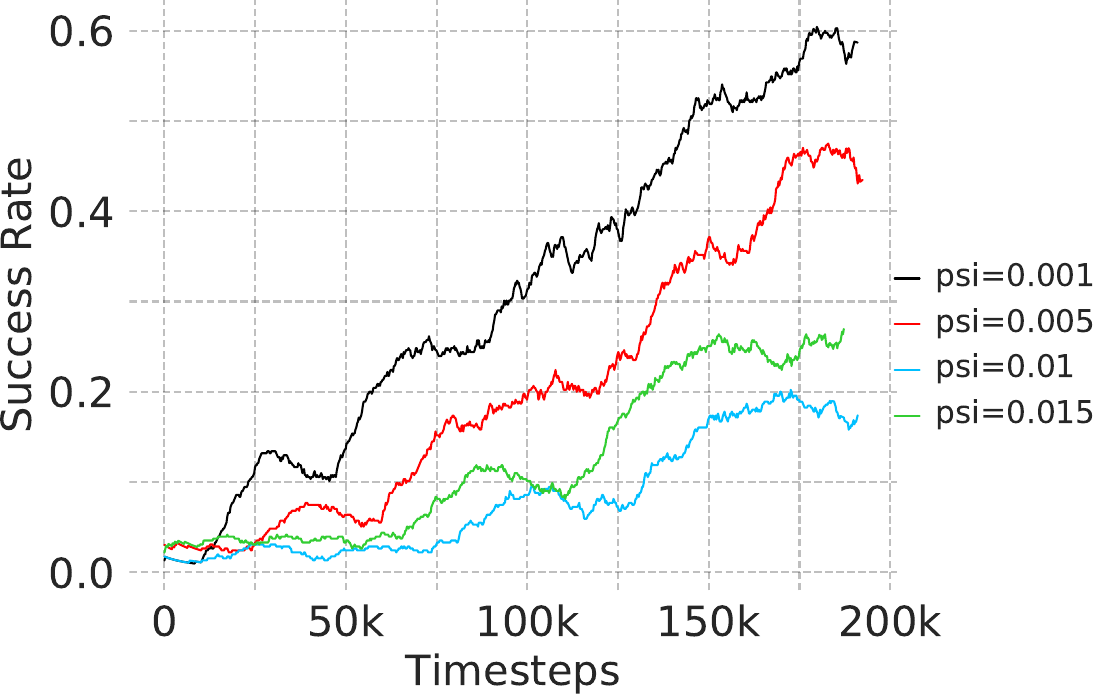}}
\subfloat[][Pick and place]{\includegraphics[scale=0.25]{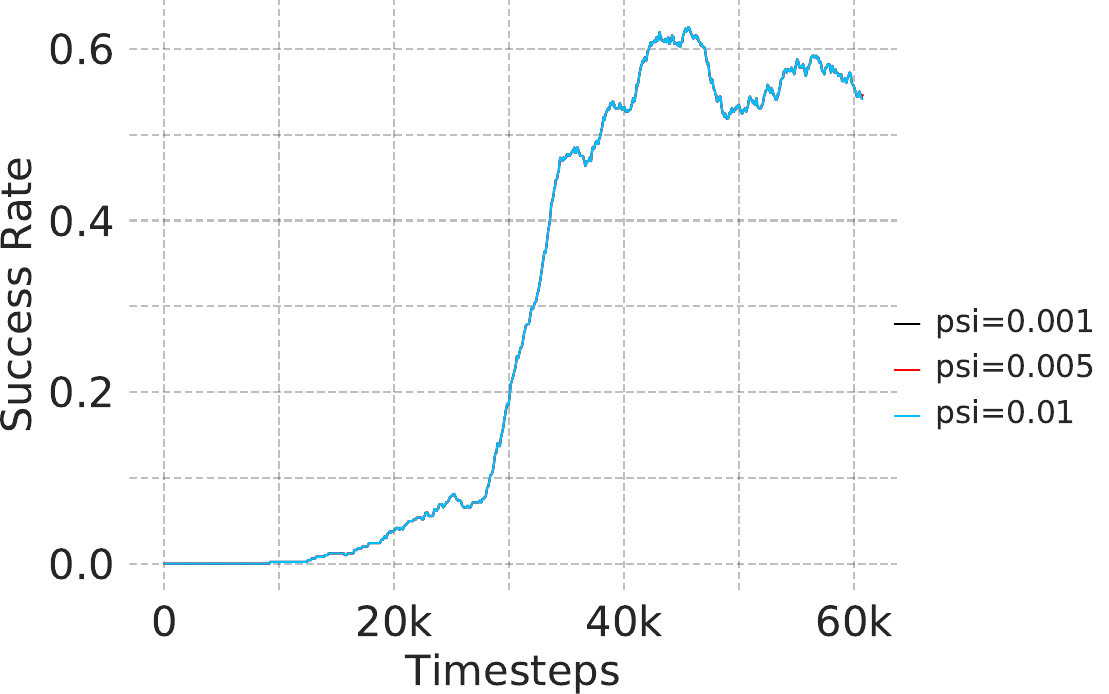}}
\subfloat[][Bin]{\includegraphics[scale=0.25]{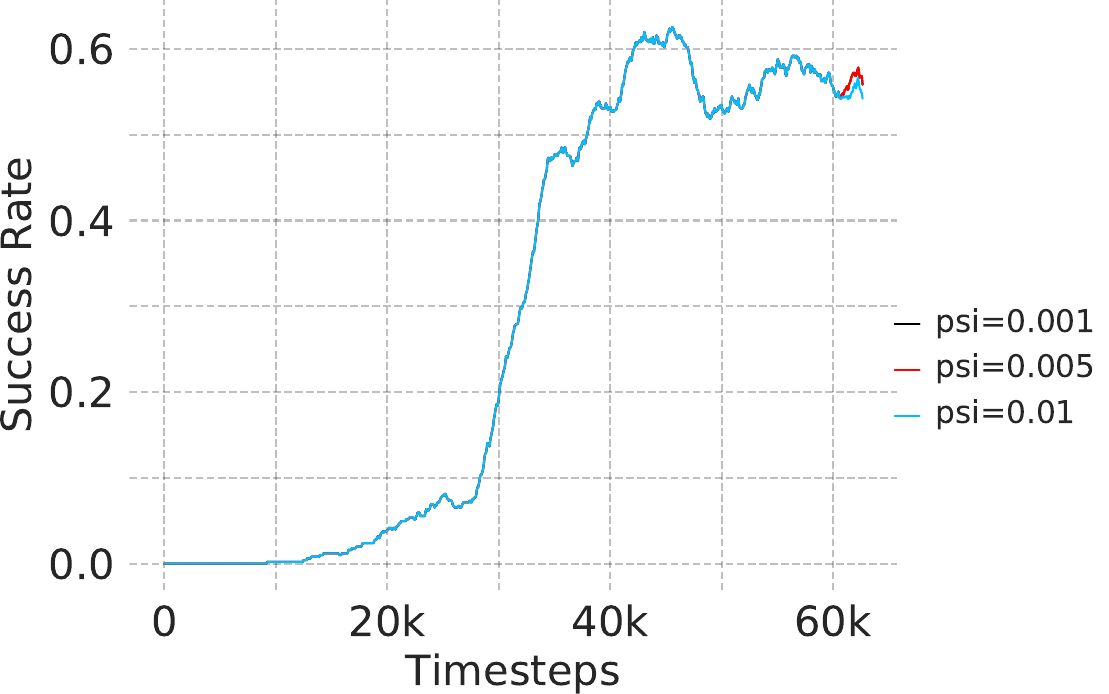}}
\\
\subfloat[][Hollow]{\includegraphics[scale=0.25]{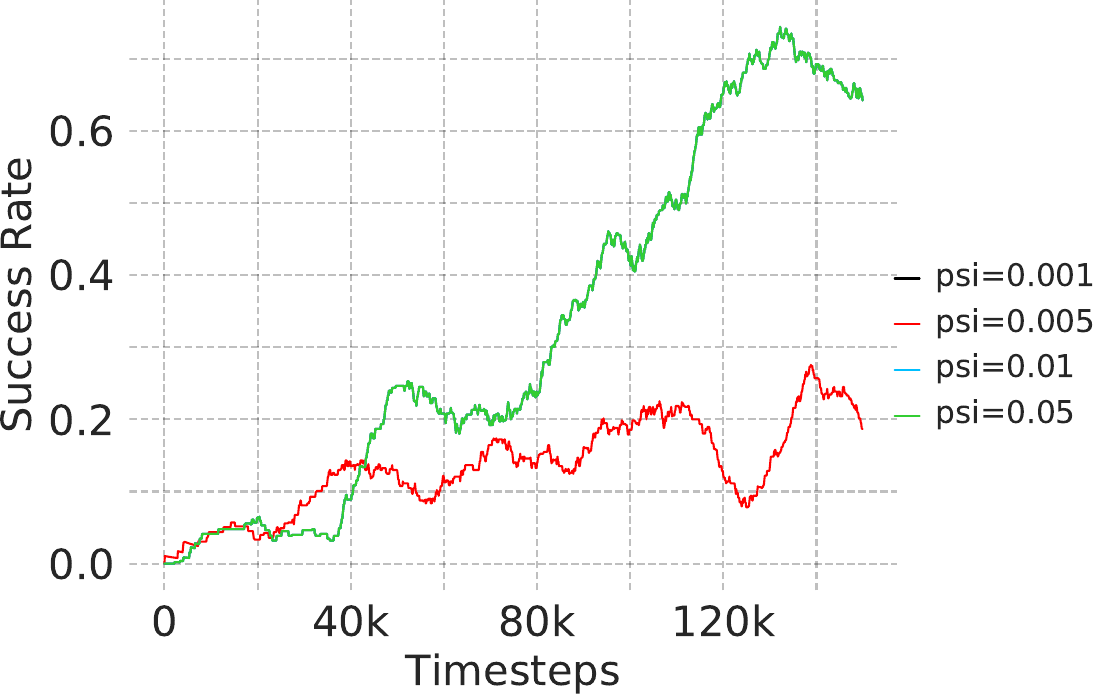}}
\subfloat[][Rope]{\includegraphics[scale=0.25]{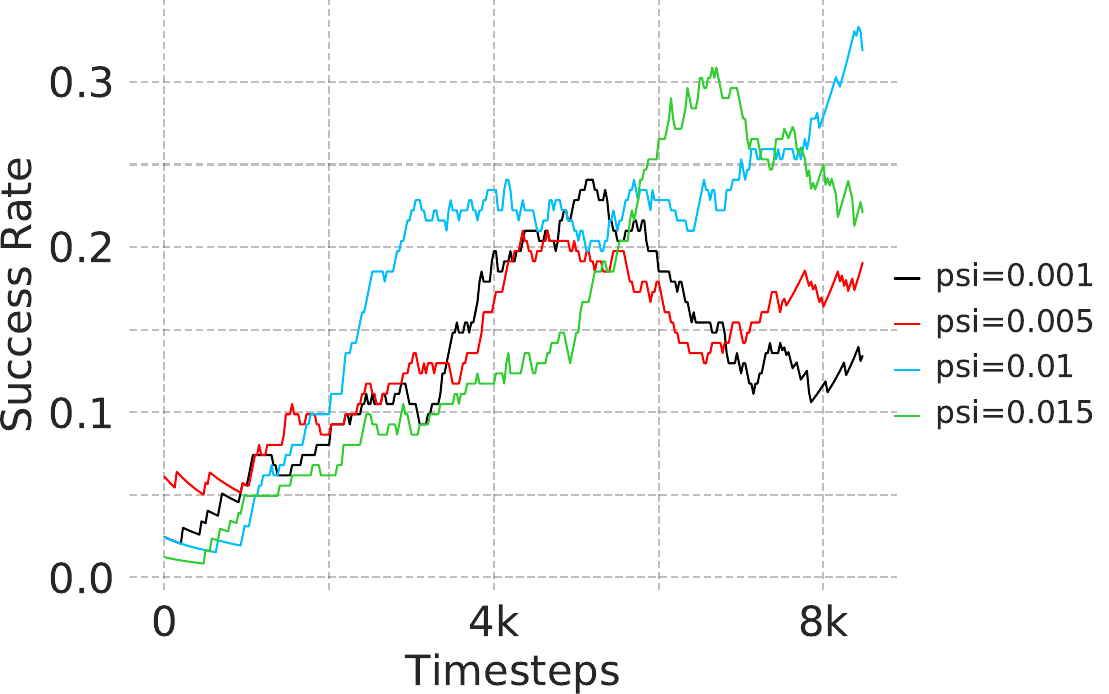}}
\subfloat[][Franka kitchen]{\includegraphics[scale=0.25]{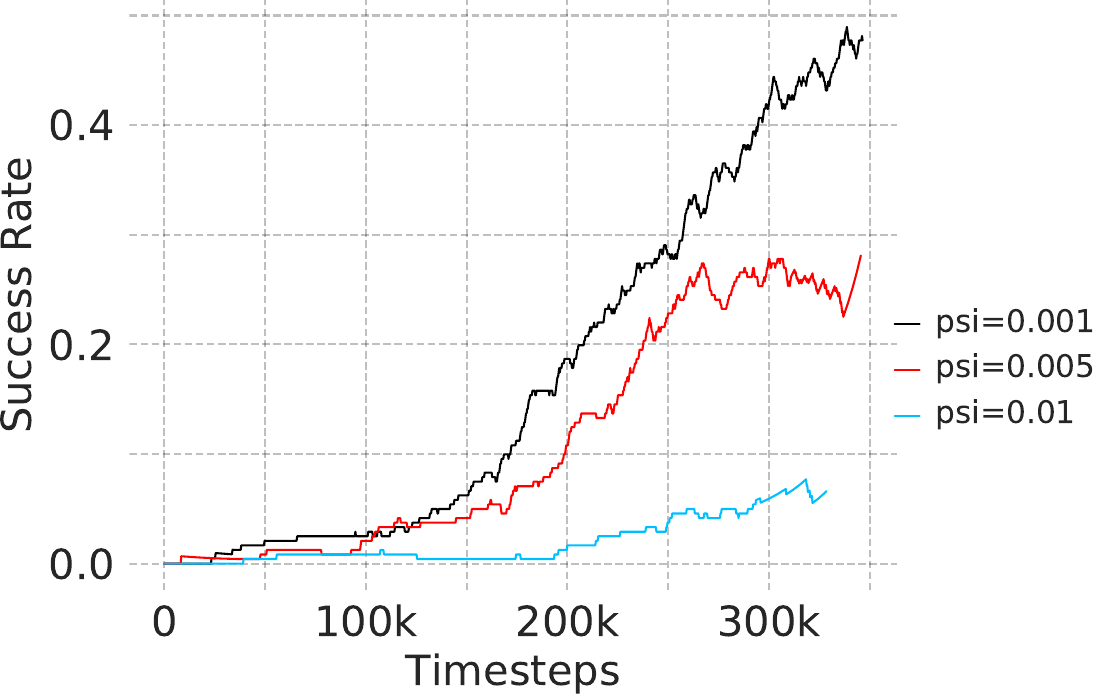}}

\caption{\textbf{$\psi$ ablation}  This figure depicts comparisons for various learning rate $\psi$ values in multiple environments}
\label{fig:psi_ablation}
\end{figure}

%% file: figures_tex/rpl_ablation.tex
\begin{figure}[H]
\vspace{5pt}
\centering
% \captionsetup{font=footnotesize,labelfont=scriptsize,textfont=scriptsize}
% \captionsetup{font=footnotesize}
\subfloat[][Maze navigation]{\includegraphics[scale=0.25]{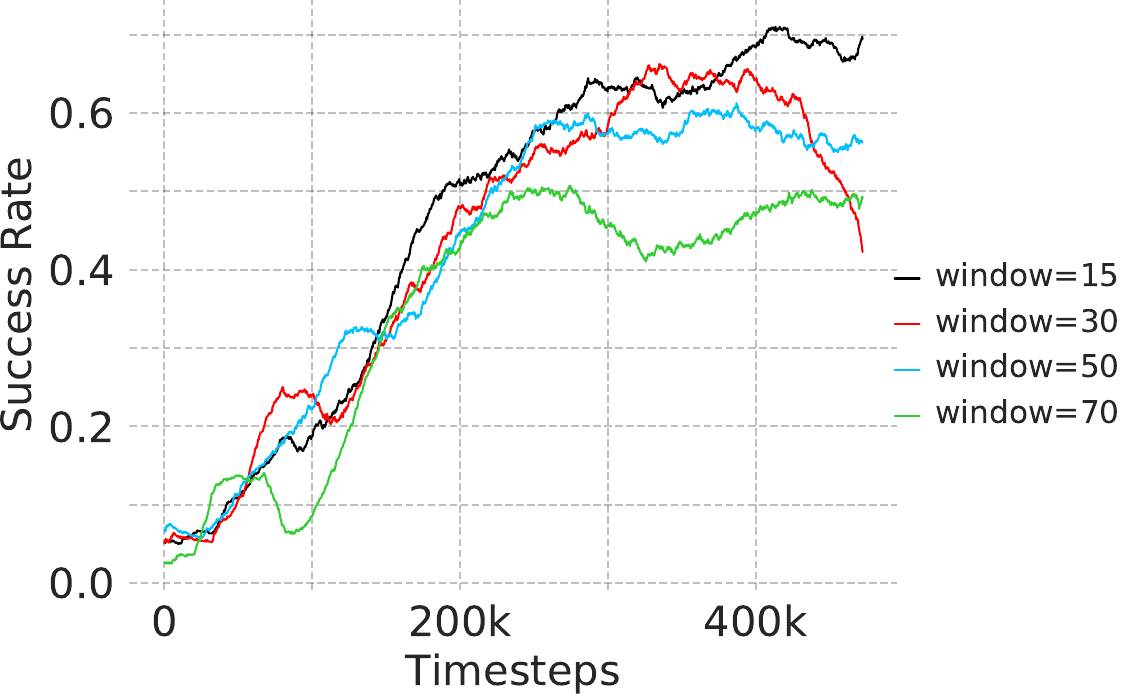}}
\subfloat[][Pick and place]{\includegraphics[scale=0.25]{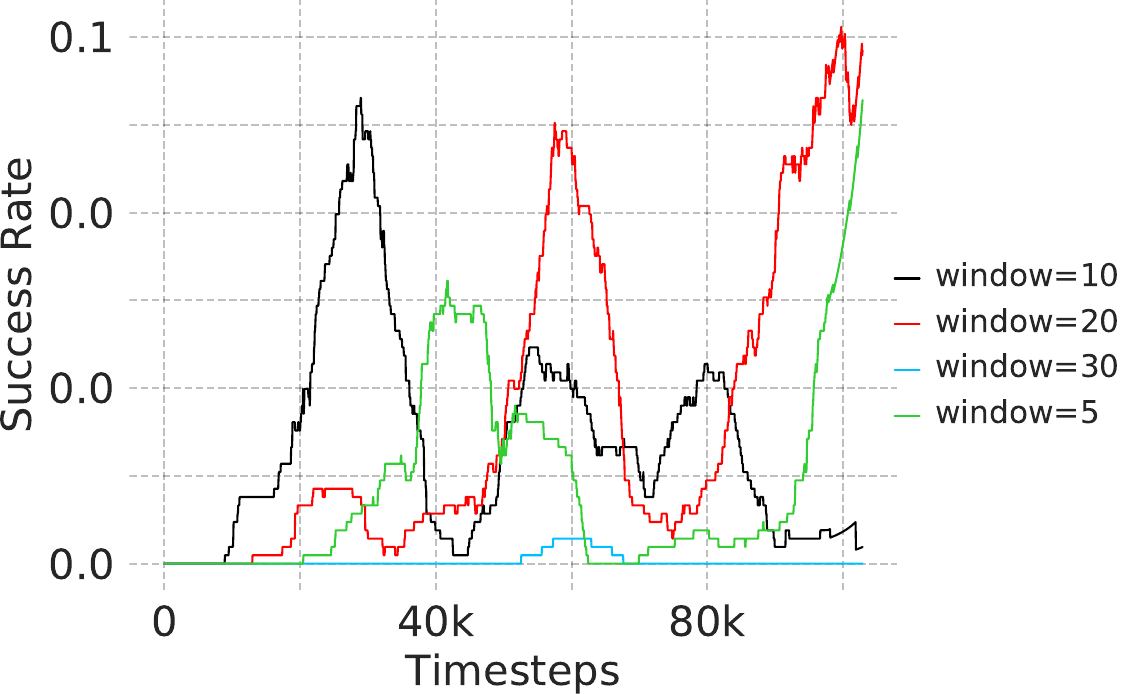}}
\subfloat[][Bin]{\includegraphics[scale=0.25]{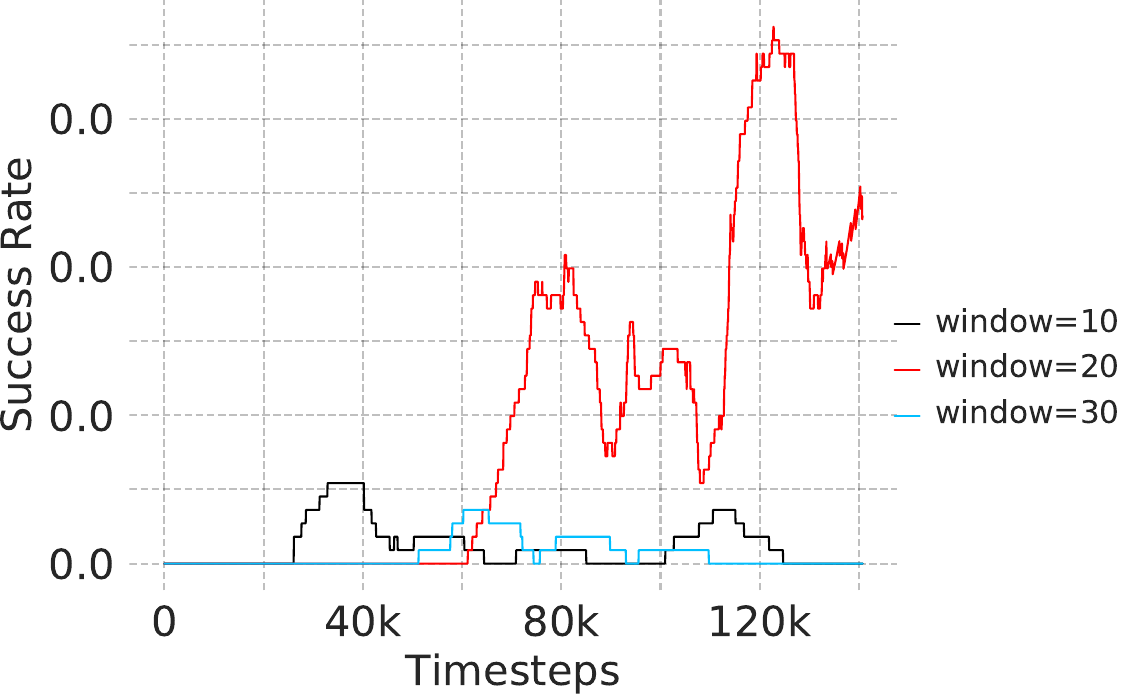}}
\\
\subfloat[][Hollow]{\includegraphics[scale=0.25]{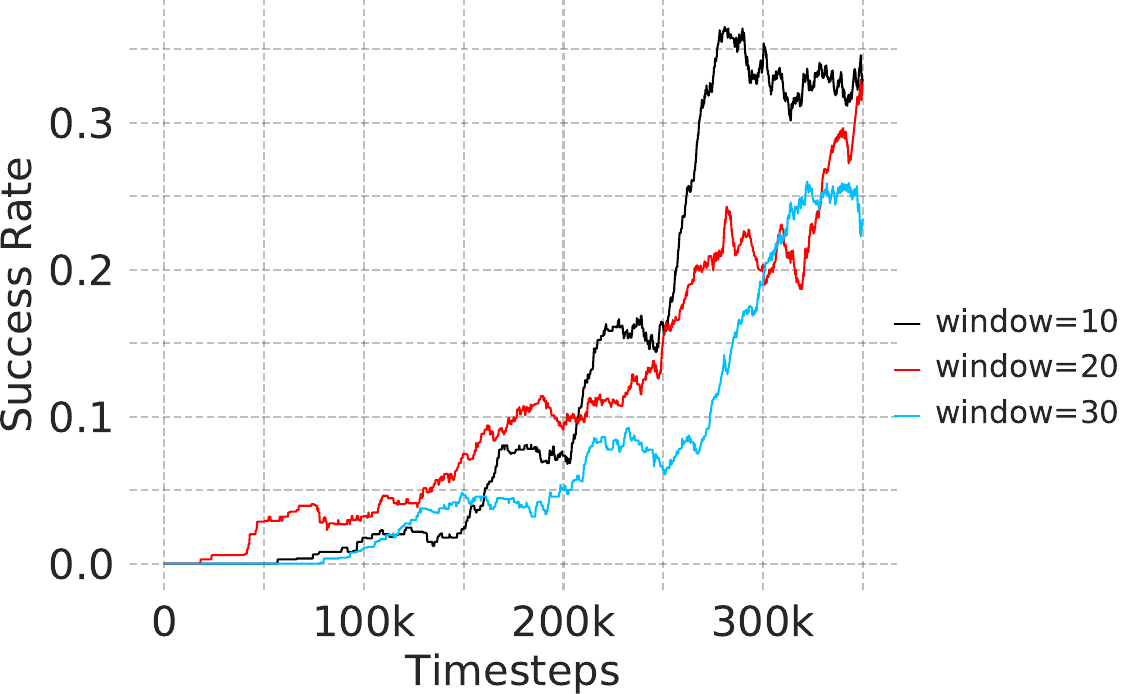}}
\subfloat[][Rope]{\includegraphics[scale=0.25]{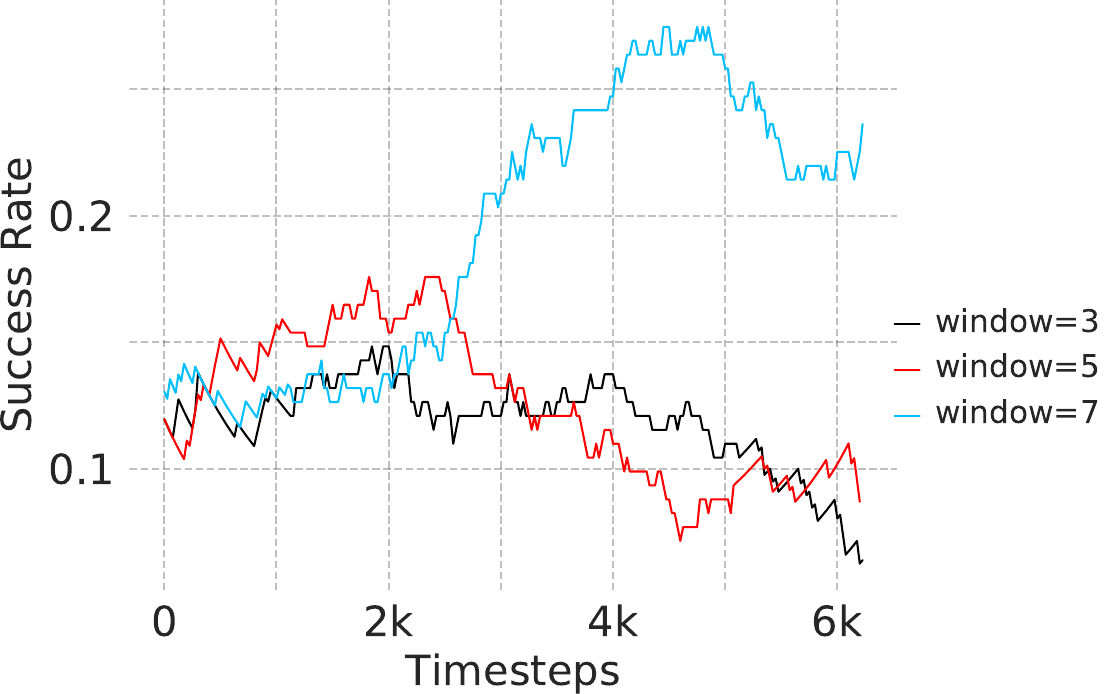}}
\subfloat[][Franka kitchen]{\includegraphics[scale=0.25]{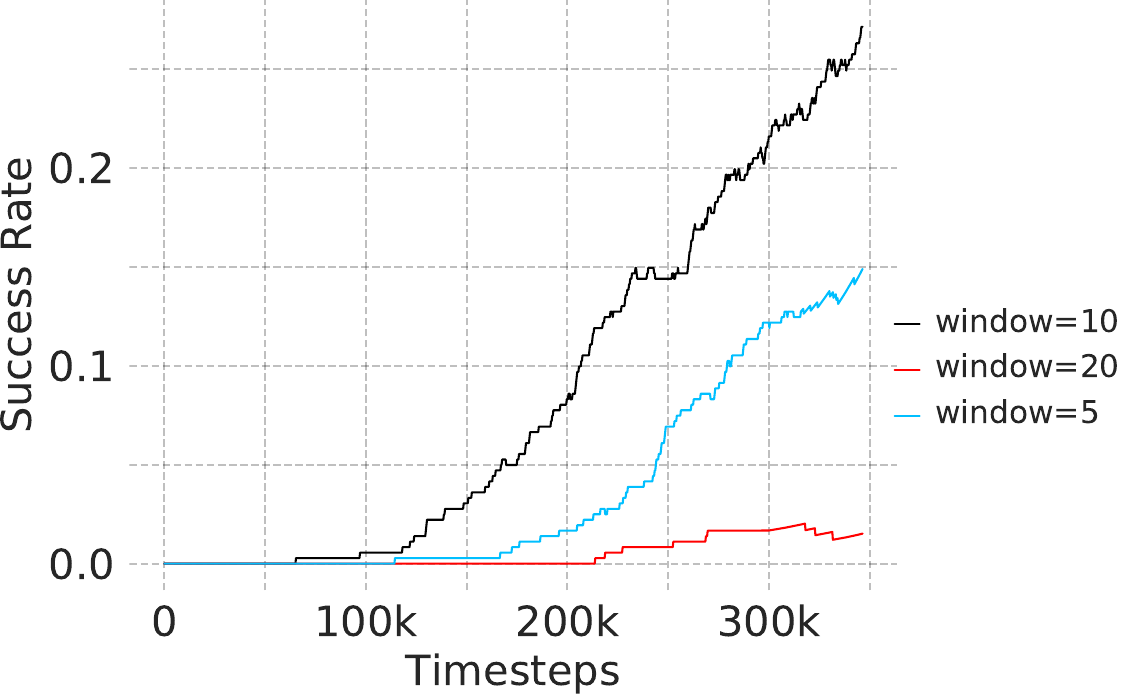}}
\caption{\textbf{Window size ablation} This figure depicts depict ablation experiments for window size hyper-parameter in RPL experiments.}
\label{fig:rpl_ablation}
\end{figure}

%% file: figures_tex/demos_ablation.tex
\begin{figure}[H]
\vspace{5pt}
\centering
% \captionsetup{font=footnotesize,labelfont=scriptsize,textfont=scriptsize}
% \captionsetup{font=footnotesize}
\subfloat[][Maze navigation]{\includegraphics[scale=0.25]{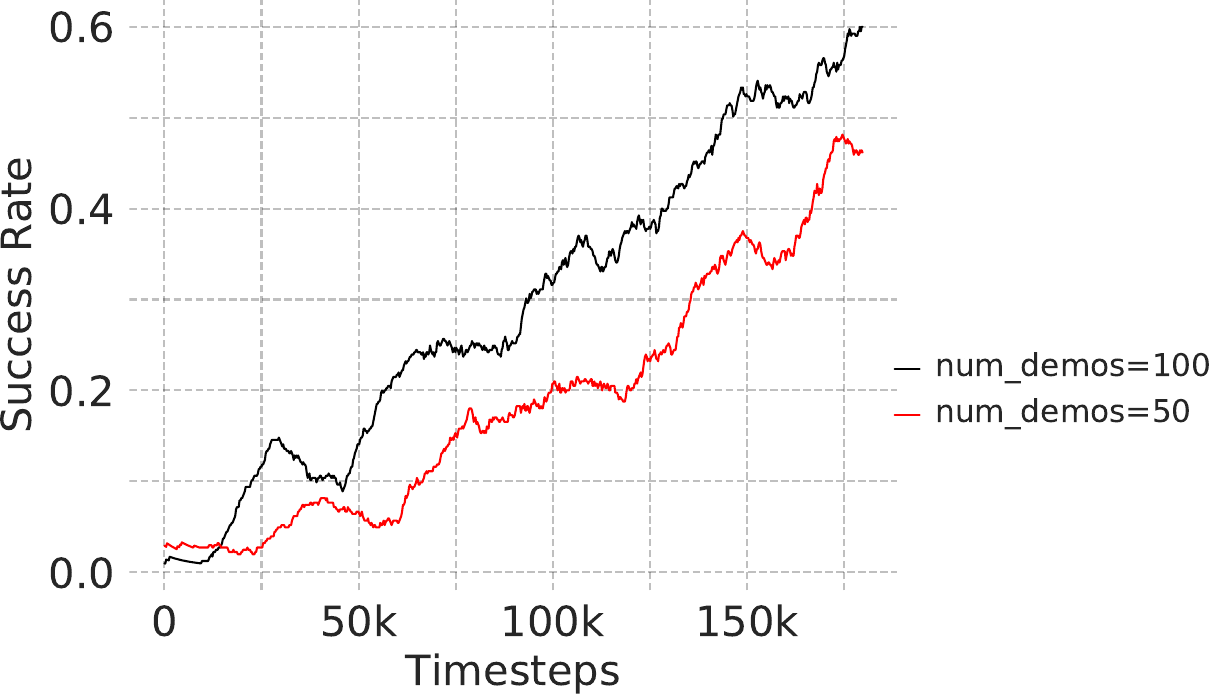}}
\subfloat[][Pick and place]{\includegraphics[scale=0.25]{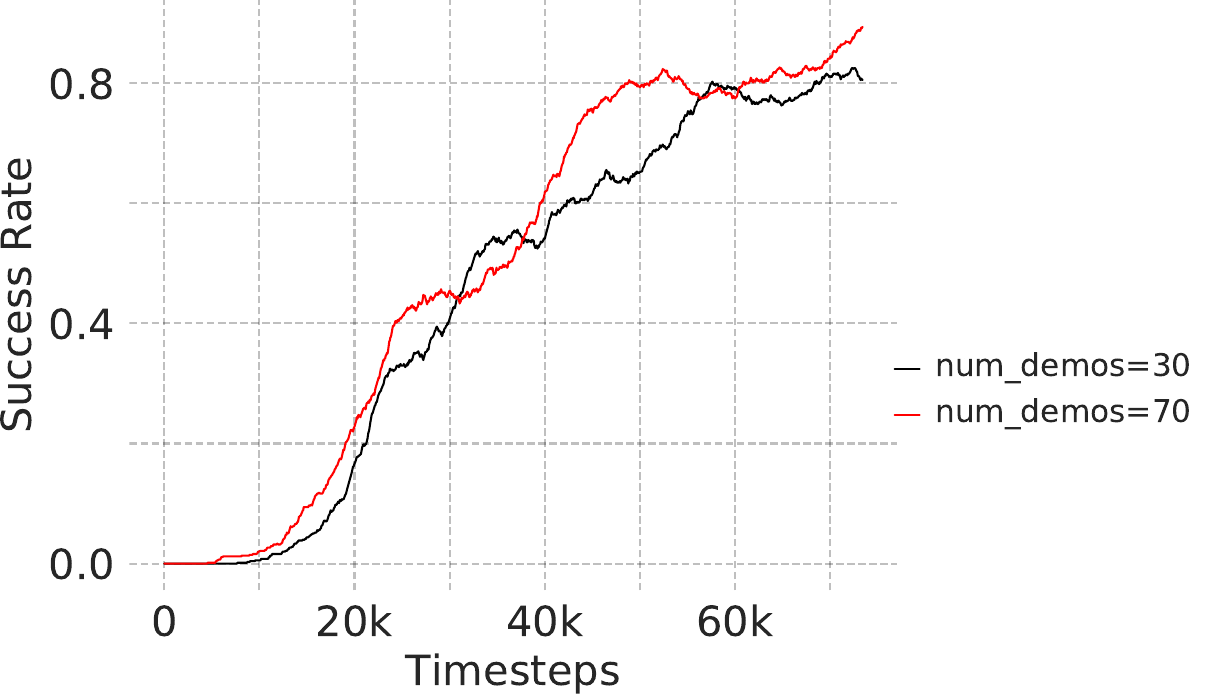}}
\subfloat[][Bin]{\includegraphics[scale=0.25]{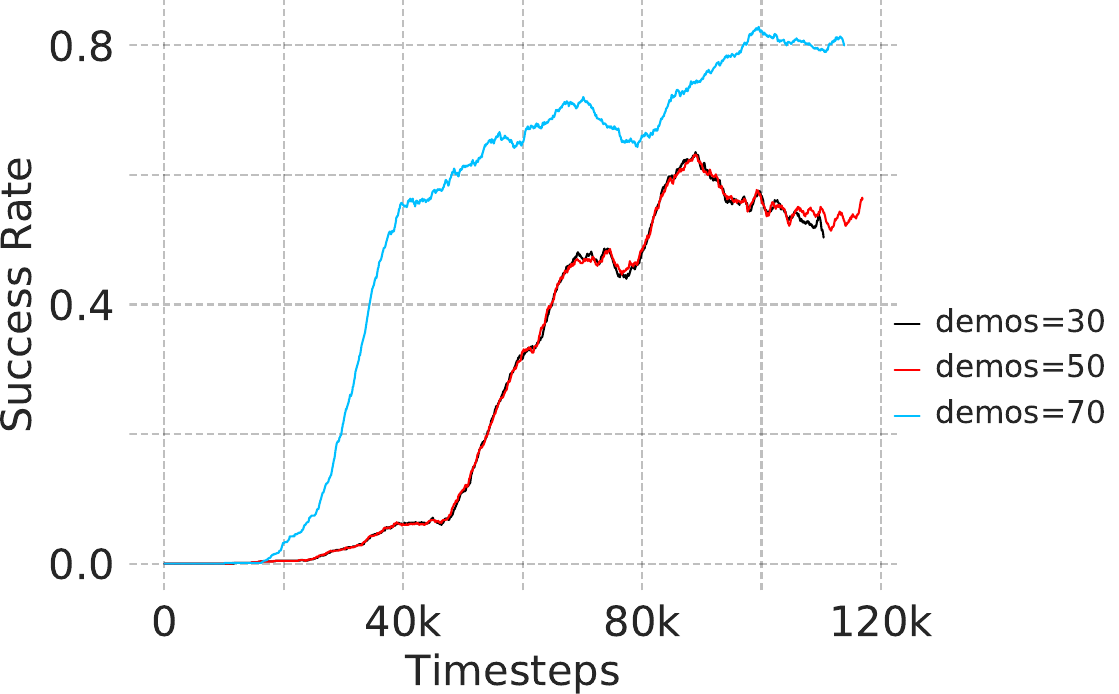}}
\\
\subfloat[][Hollow]{\includegraphics[scale=0.25]{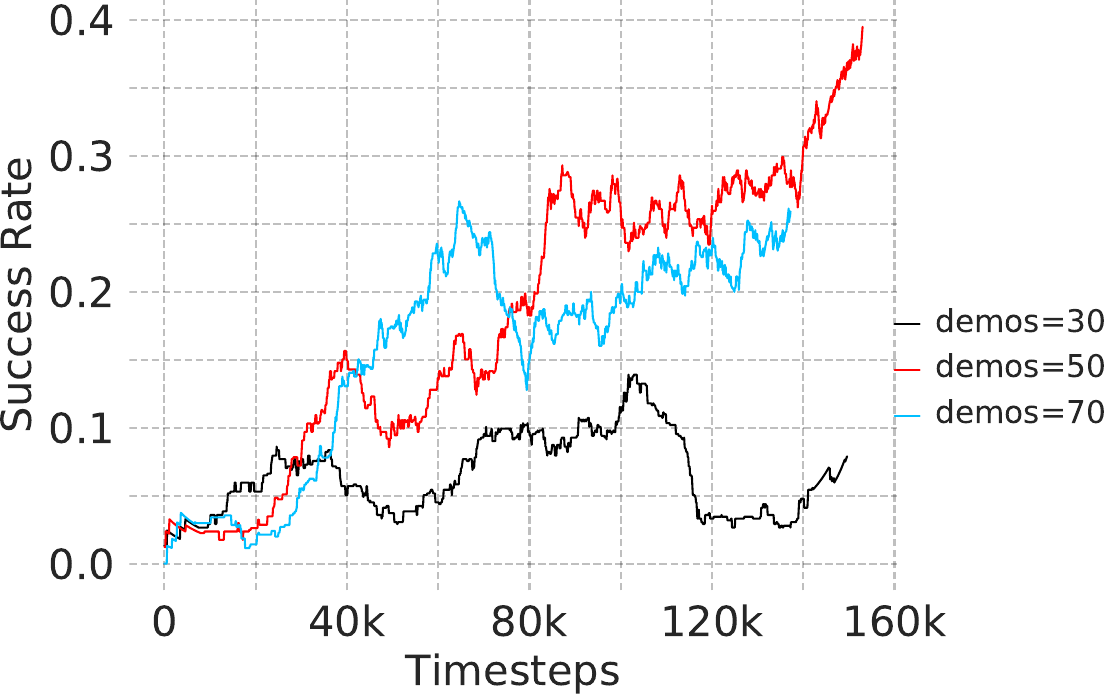}}
\subfloat[][Rope]{\includegraphics[scale=0.25]{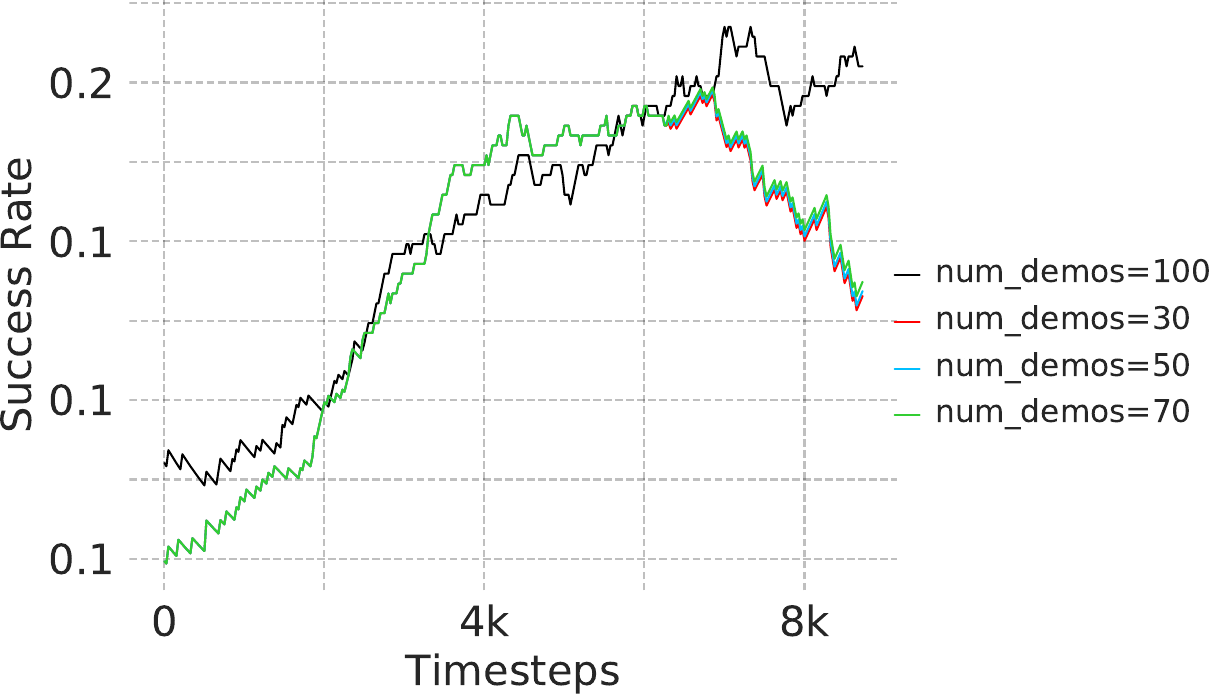}}
\subfloat[][Franka kitchen]{\includegraphics[scale=0.25]{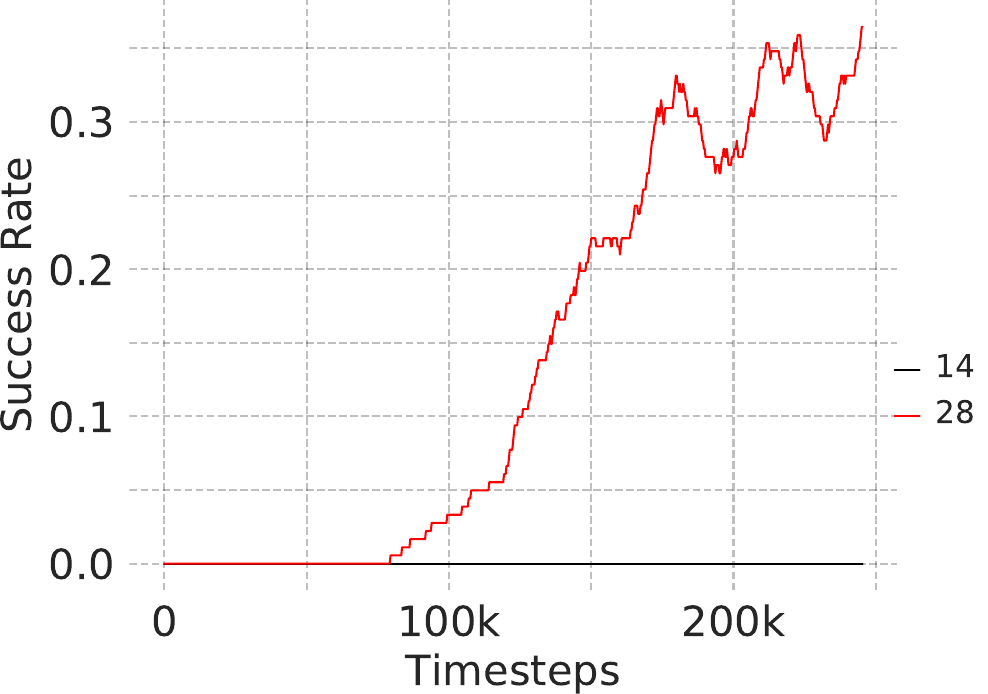}}
\caption{\textbf{Number of expert demonstrations ablation} This figure depicts ablation experiments of varying number of expert demonstrations versus number of training epochs.}
\label{fig:demos_ablation}
\end{figure}

%% file: figures_tex/dobot_real.tex
\begin{figure}[H]
\vspace{1pt}
\centering
% \captionsetup{font=footnotesize,labelfont=scriptsize,textfont=scriptsize}
% \captionsetup{font=footnotesize,labelfont=scriptsize,textfont=scriptsize}
\includegraphics[scale=0.13]{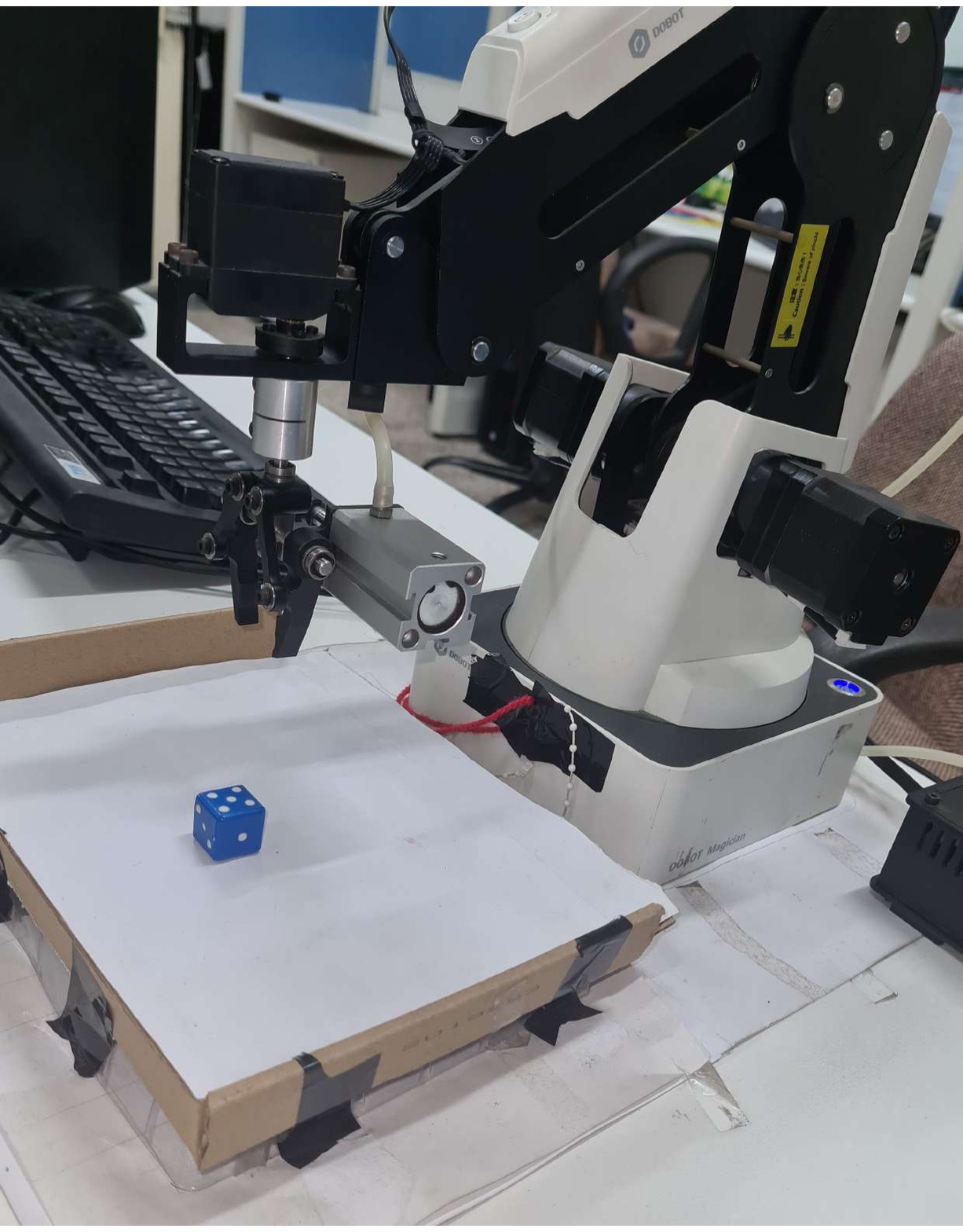}
\hspace{0.005cm}
\includegraphics[scale=0.13]{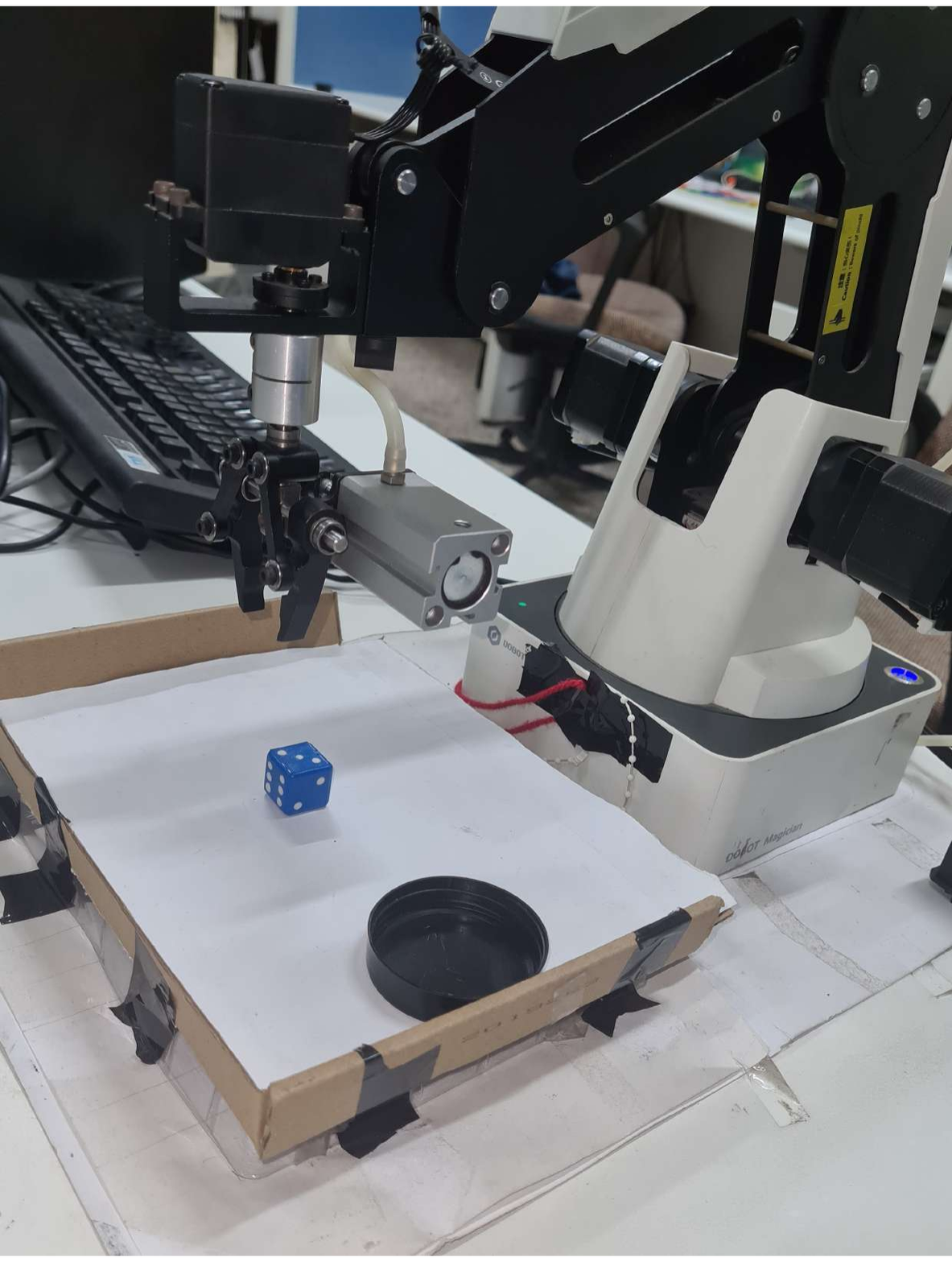}
\hspace{0.005cm}
\includegraphics[scale=0.13]{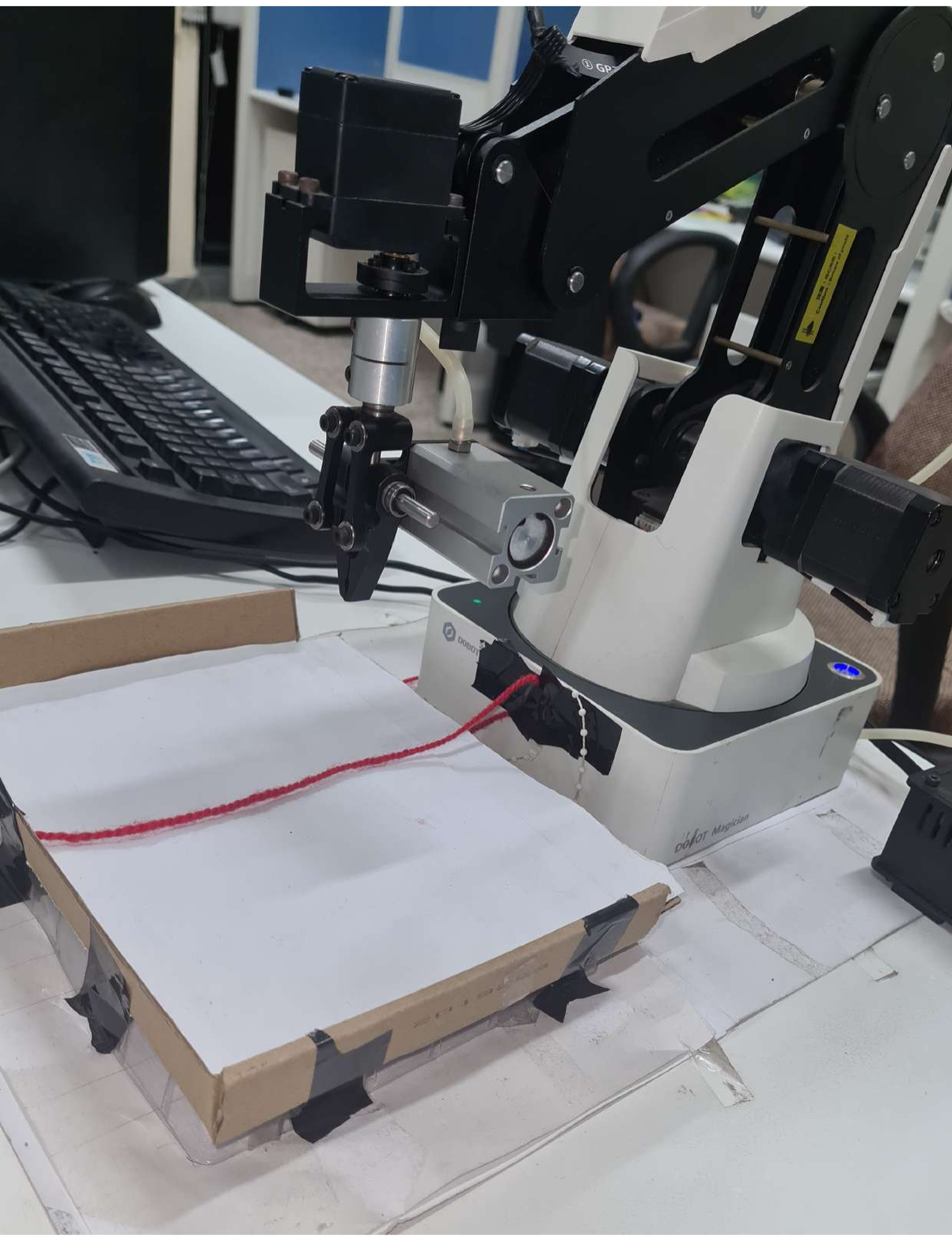}
\vspace{-0.5cm}
\\
\subfloat[][Pick]{\includegraphics[scale=0.13]{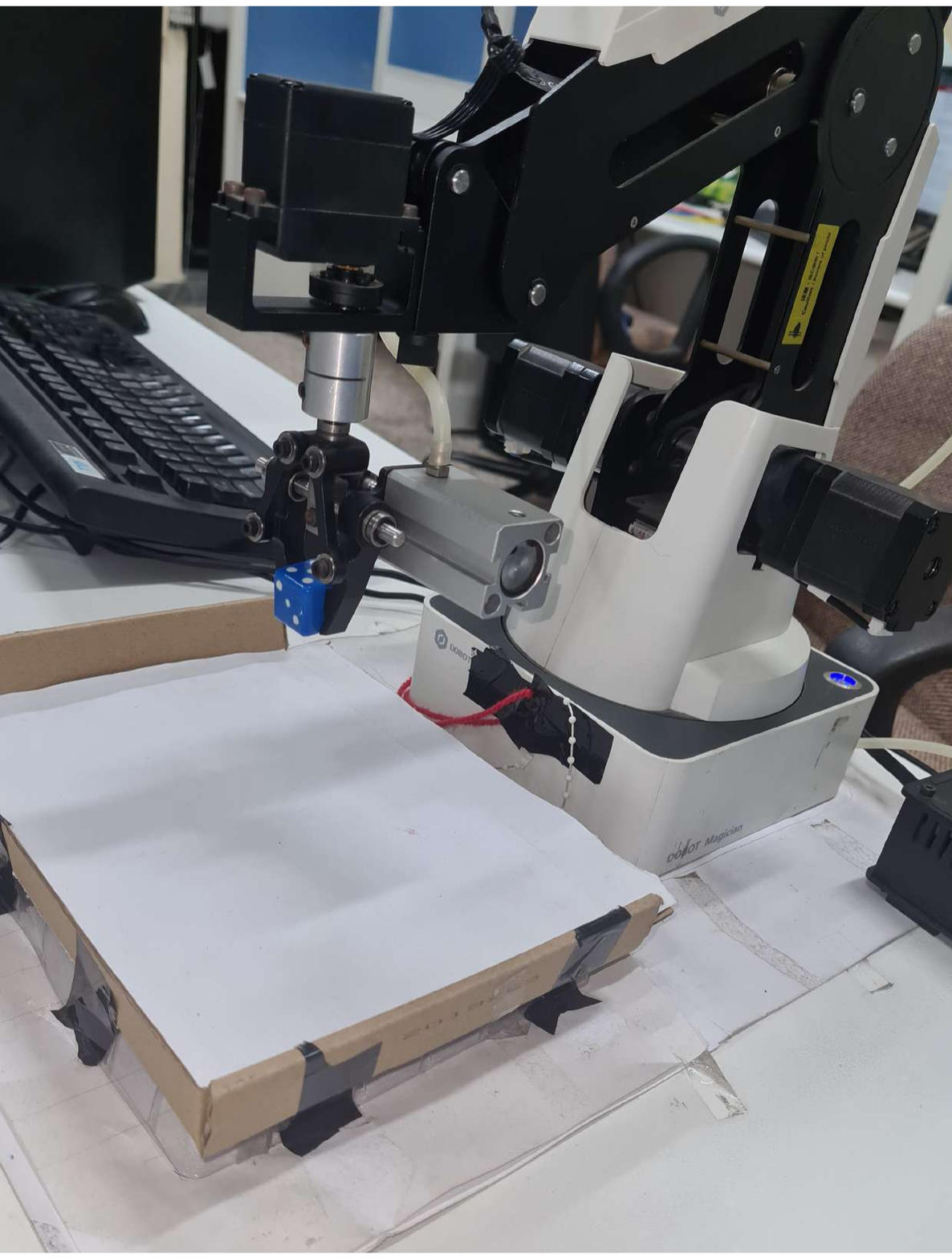}}
\hspace{0.1cm}
\subfloat[][Bin]{\includegraphics[scale=0.13]{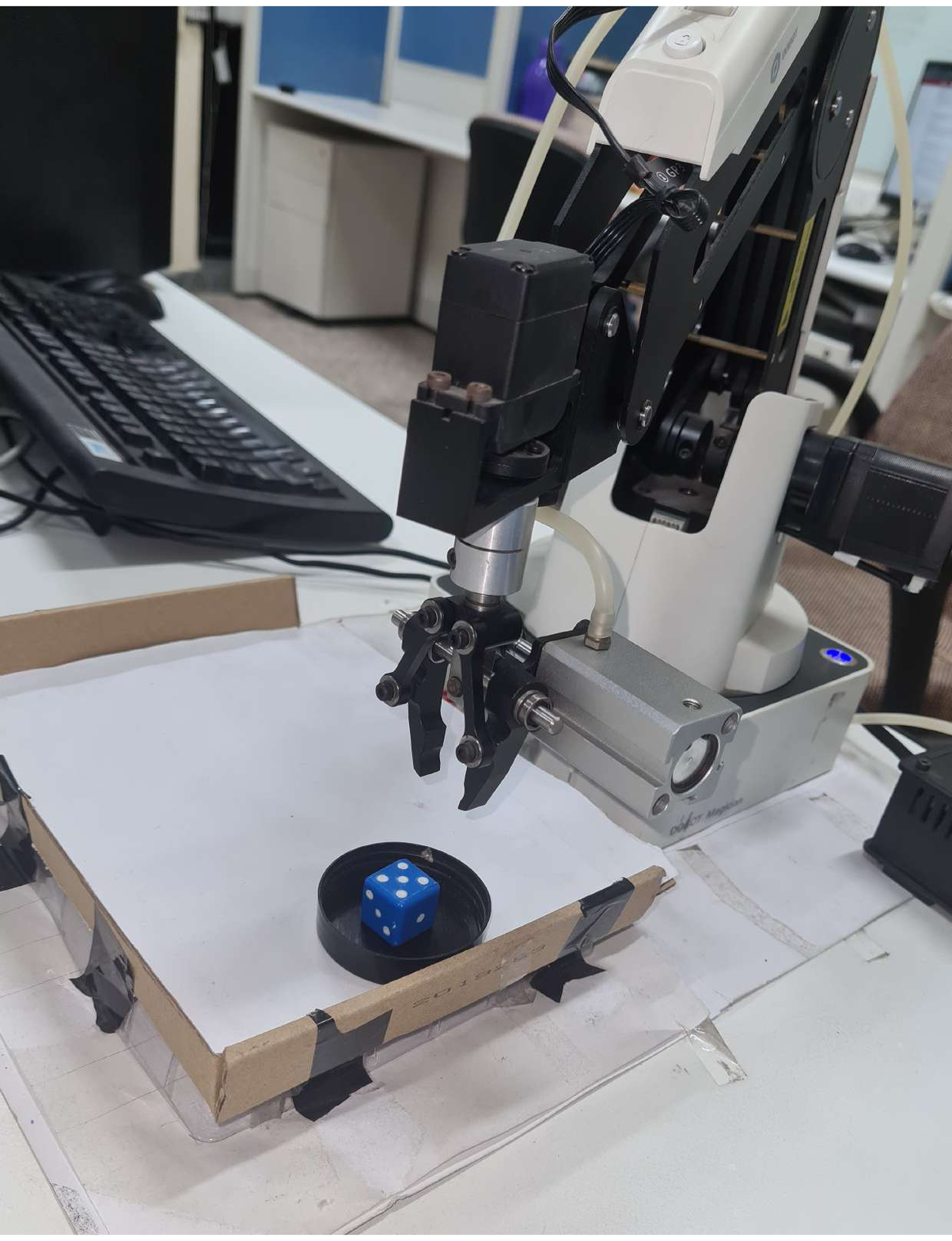}}
\hspace{0.1cm}
\subfloat[][Rope]{\includegraphics[scale=0.13]{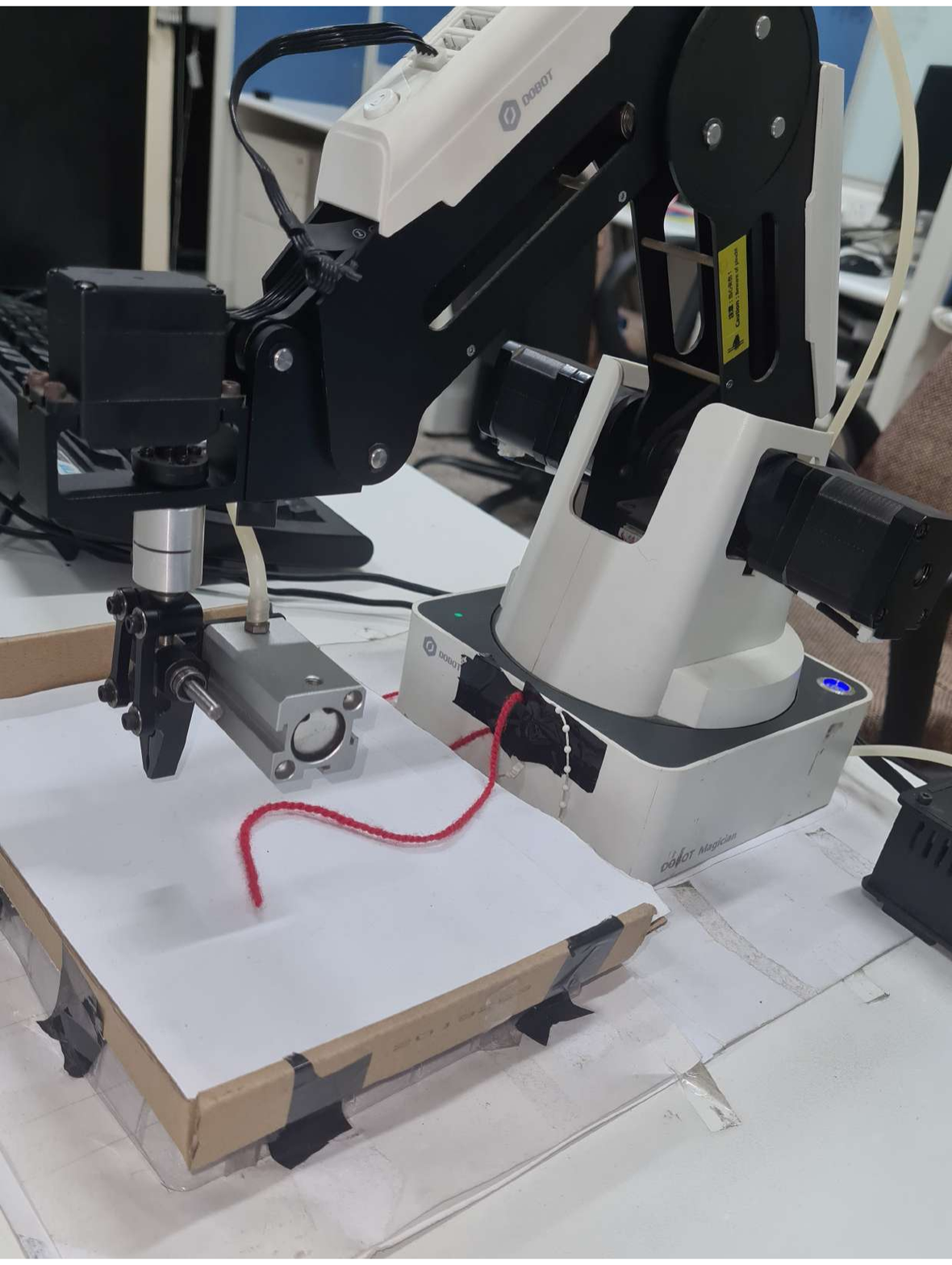}}
\caption{\textbf{Real world tasks} This figure depicts the three real world environments in our experiments: $(a)$ pick and place, $(b)$ bin, and $(c)$ rope manipulation tasks. In pick and place, the robotic gripper has to pick the block and bring it to goal position. In bin task, the gripper has to pick the block and place it in the bin. In kitchen task, the gripper has to poke to deformable rope, and align it to the goal rope configuration. Row 1 depicts initial state and Row 2 depicts final goal configurations.}
\label{fig:dobot_real}
\end{figure}

%% file: figures_tex/maze_success_visualization_1.tex
\begin{figure}[H]
\vspace{5pt}
\centering
\captionsetup{font=footnotesize,labelfont=scriptsize,textfont=scriptsize}
\includegraphics[height=1.8cm,width=2.1cm]{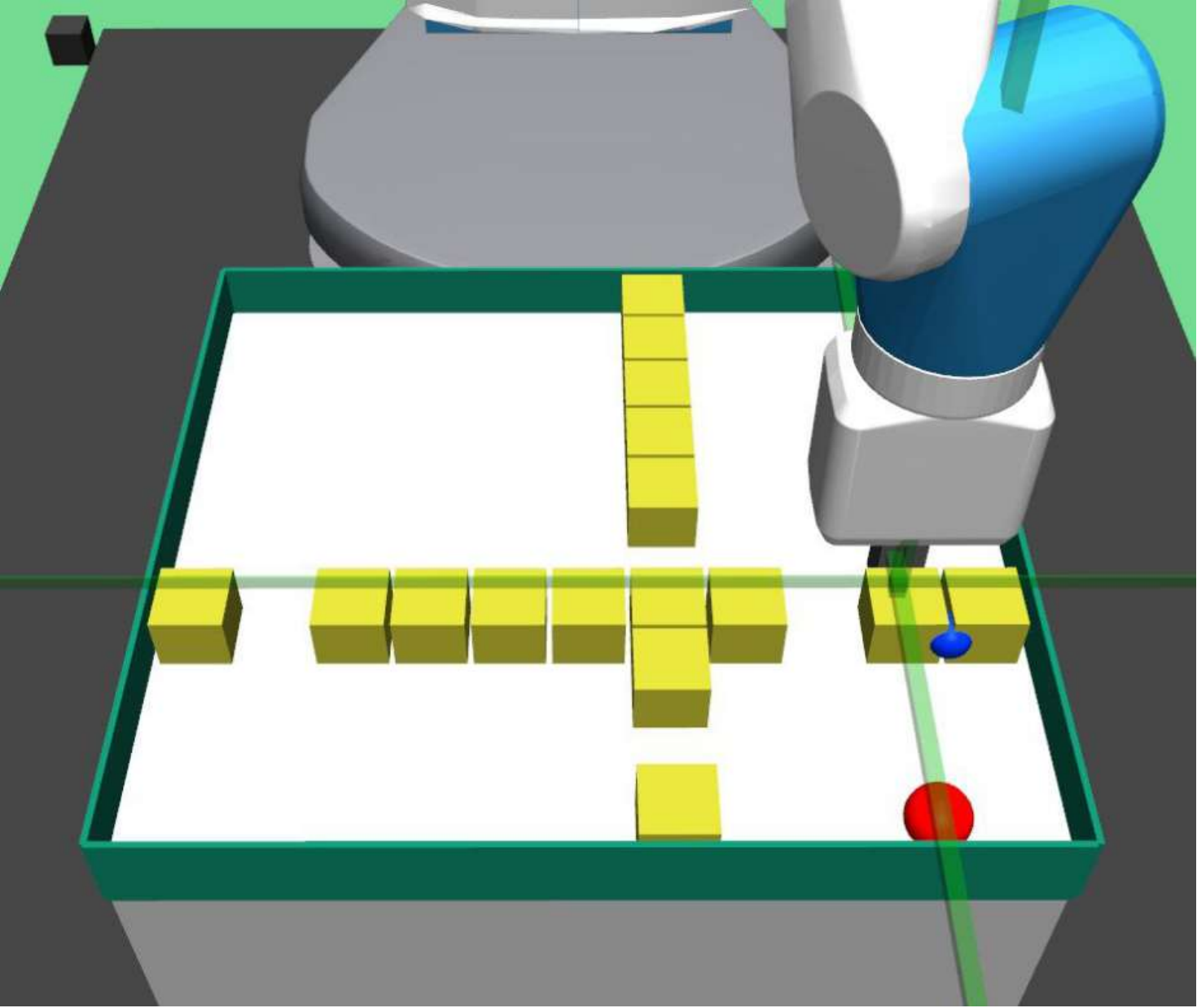}
\includegraphics[height=1.8cm,width=2.1cm]{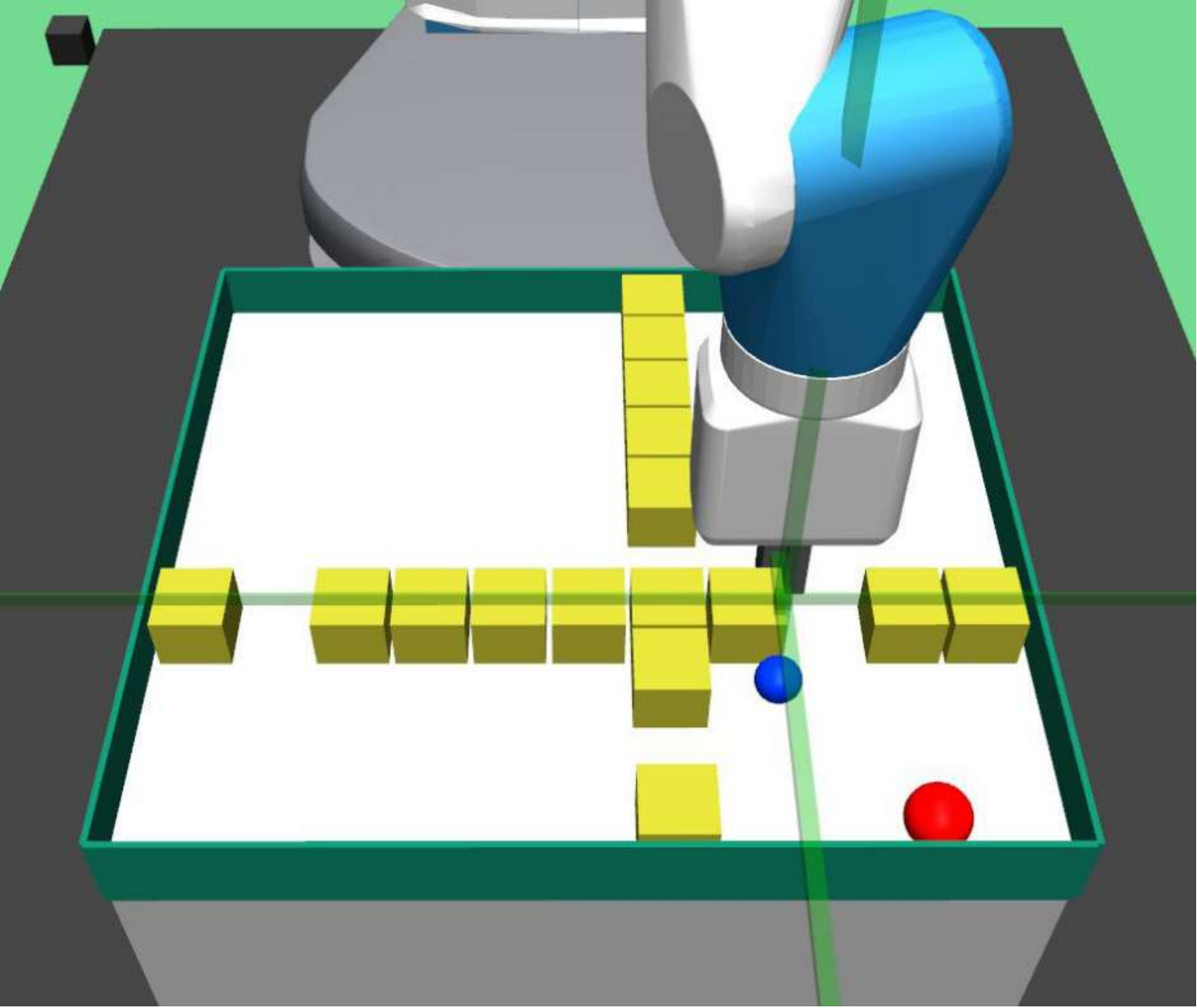}
\includegraphics[height=1.8cm,width=2.1cm]{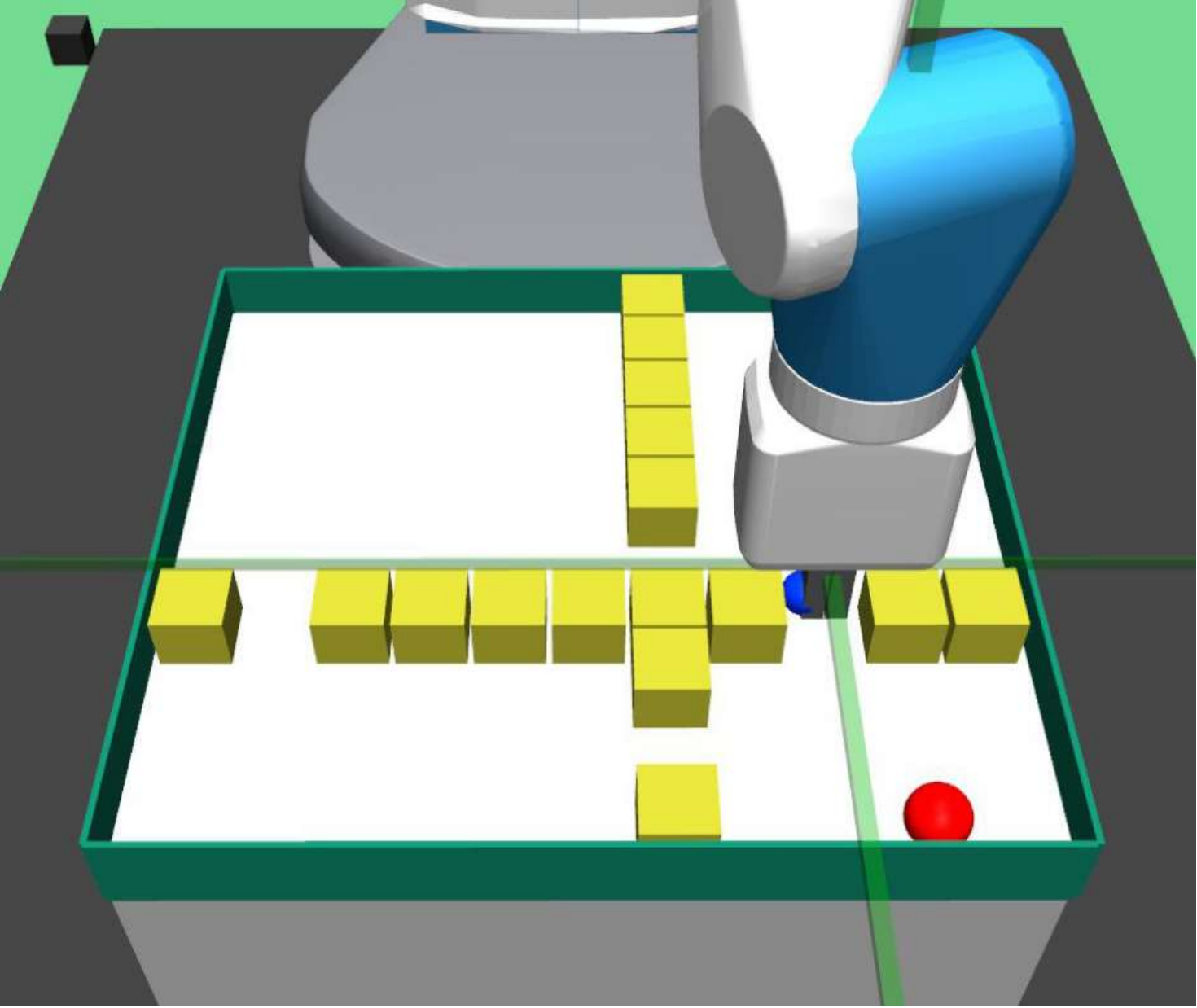}
\includegraphics[height=1.8cm,width=2.1cm]{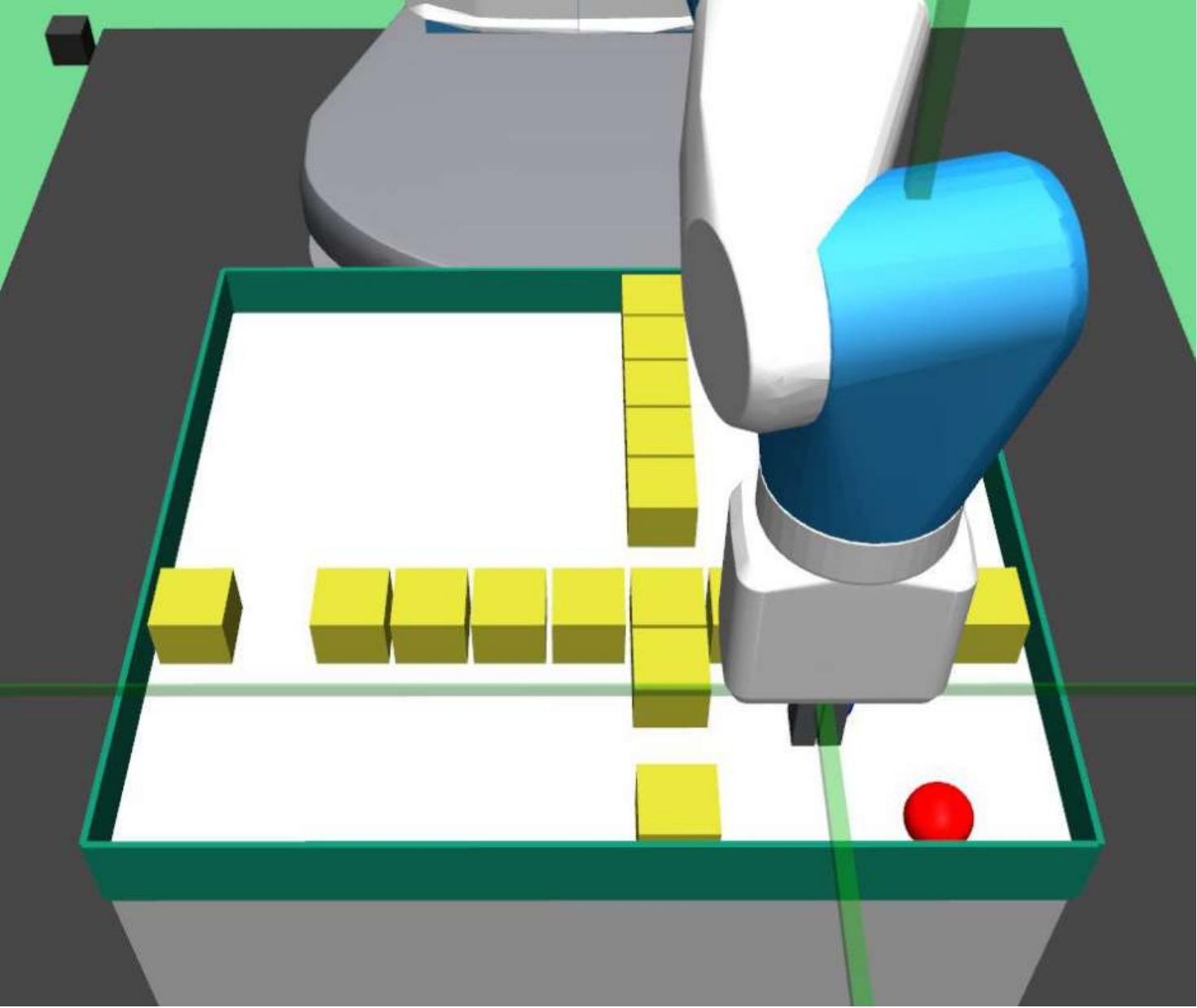}
\includegraphics[height=1.8cm,width=2.1cm]{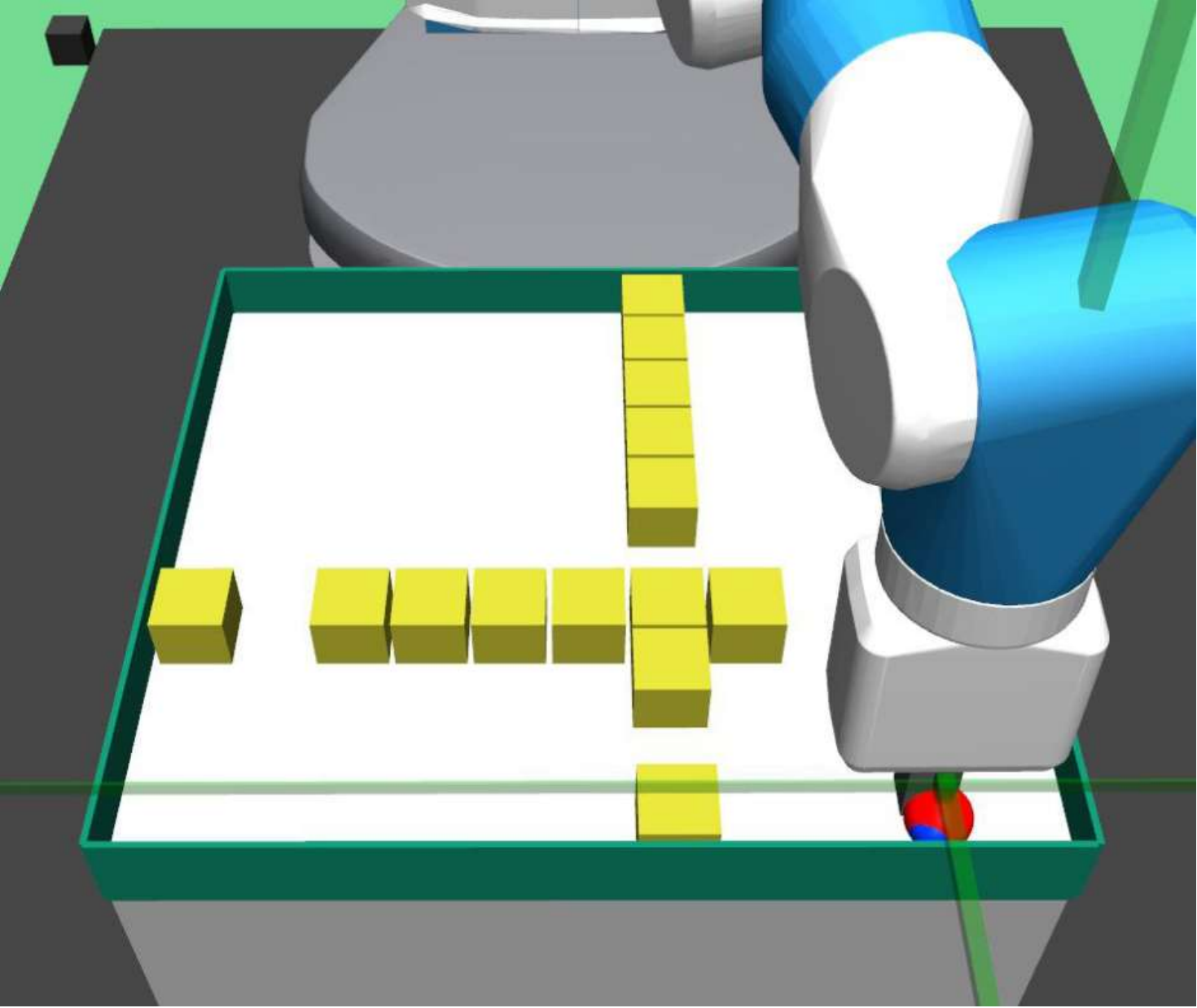}
\includegraphics[height=1.8cm,width=2.1cm]{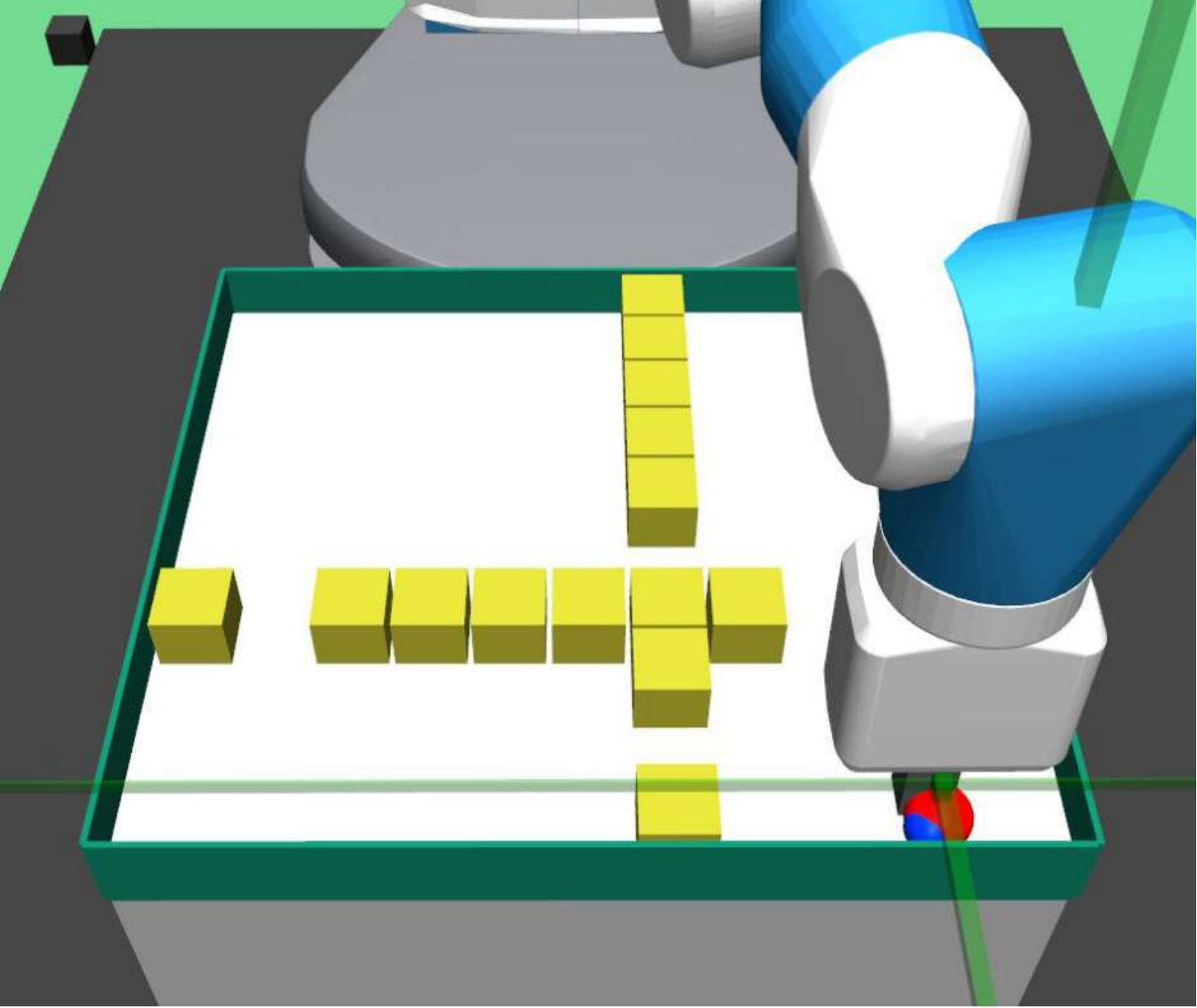}
\caption{The visualization is a successful attempt at performing maze navigation task}
\label{fig:maze_viz_success_1_ablation}
\end{figure}

%% file: figures_tex/pick_success_visualization_1.tex
\begin{figure}[H]
\vspace{5pt}
\centering
\captionsetup{font=footnotesize,labelfont=scriptsize,textfont=scriptsize}
\includegraphics[height=1.8cm,width=2.1cm]{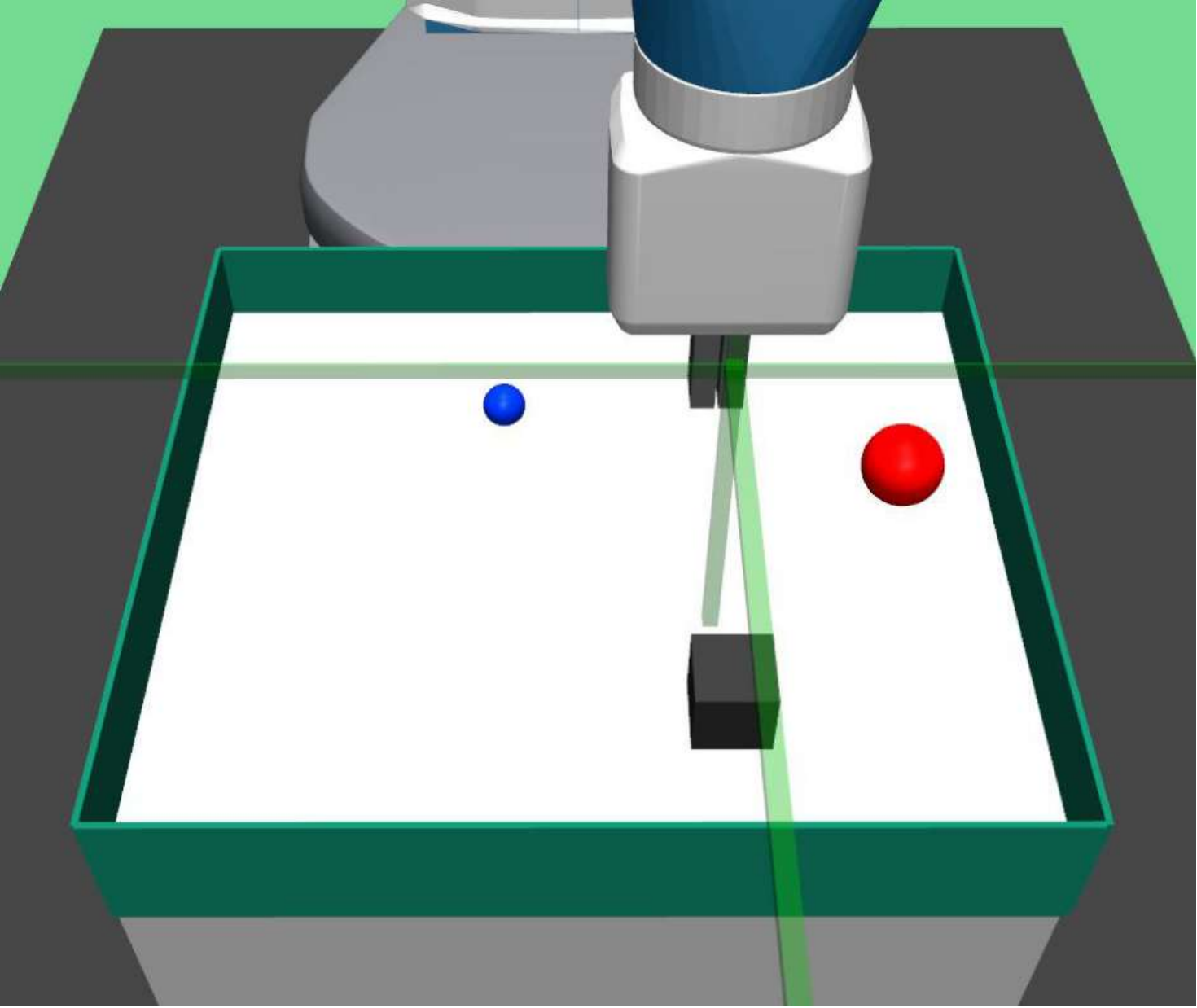}
\includegraphics[height=1.8cm,width=2.1cm]{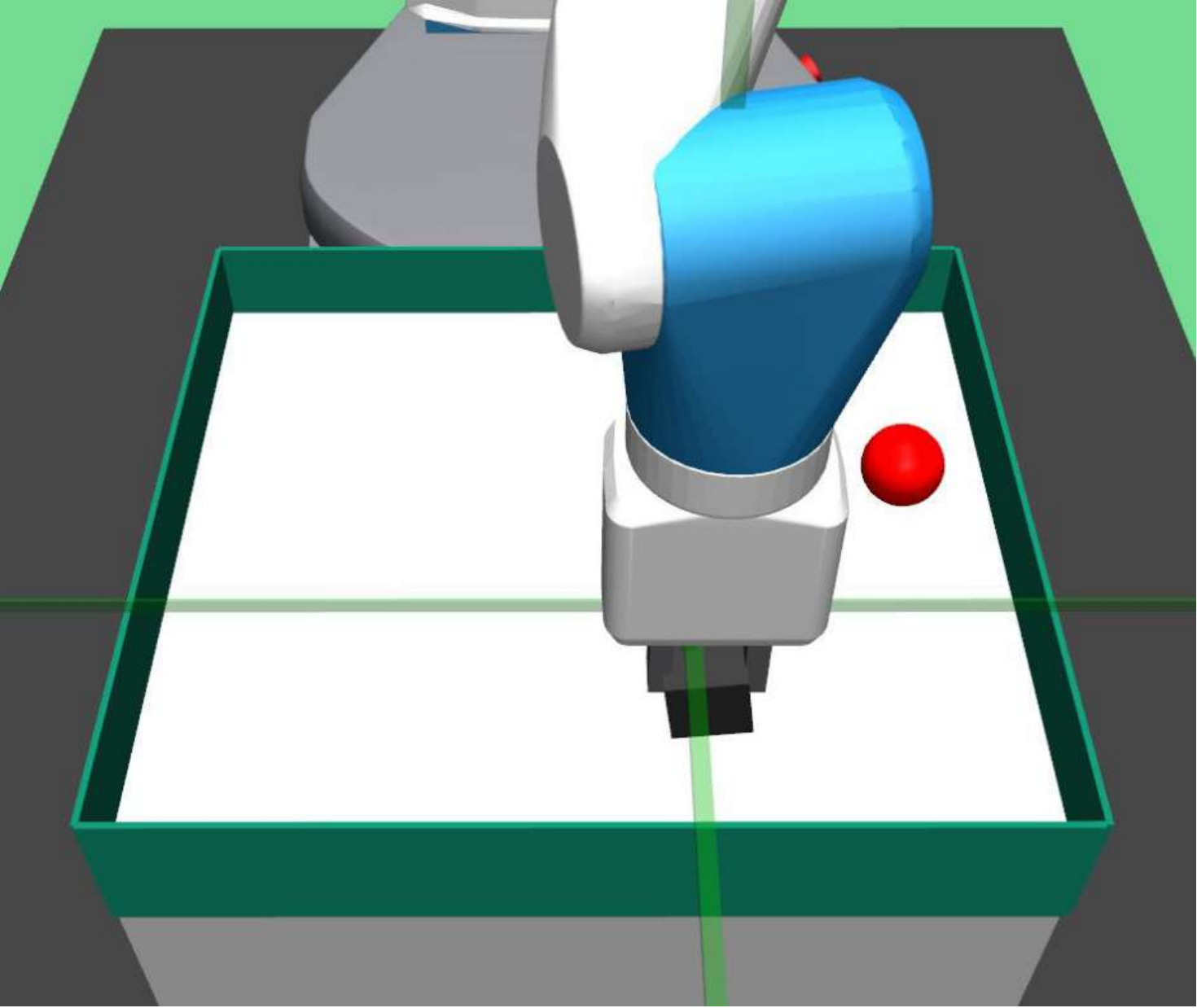}
\includegraphics[height=1.8cm,width=2.1cm]{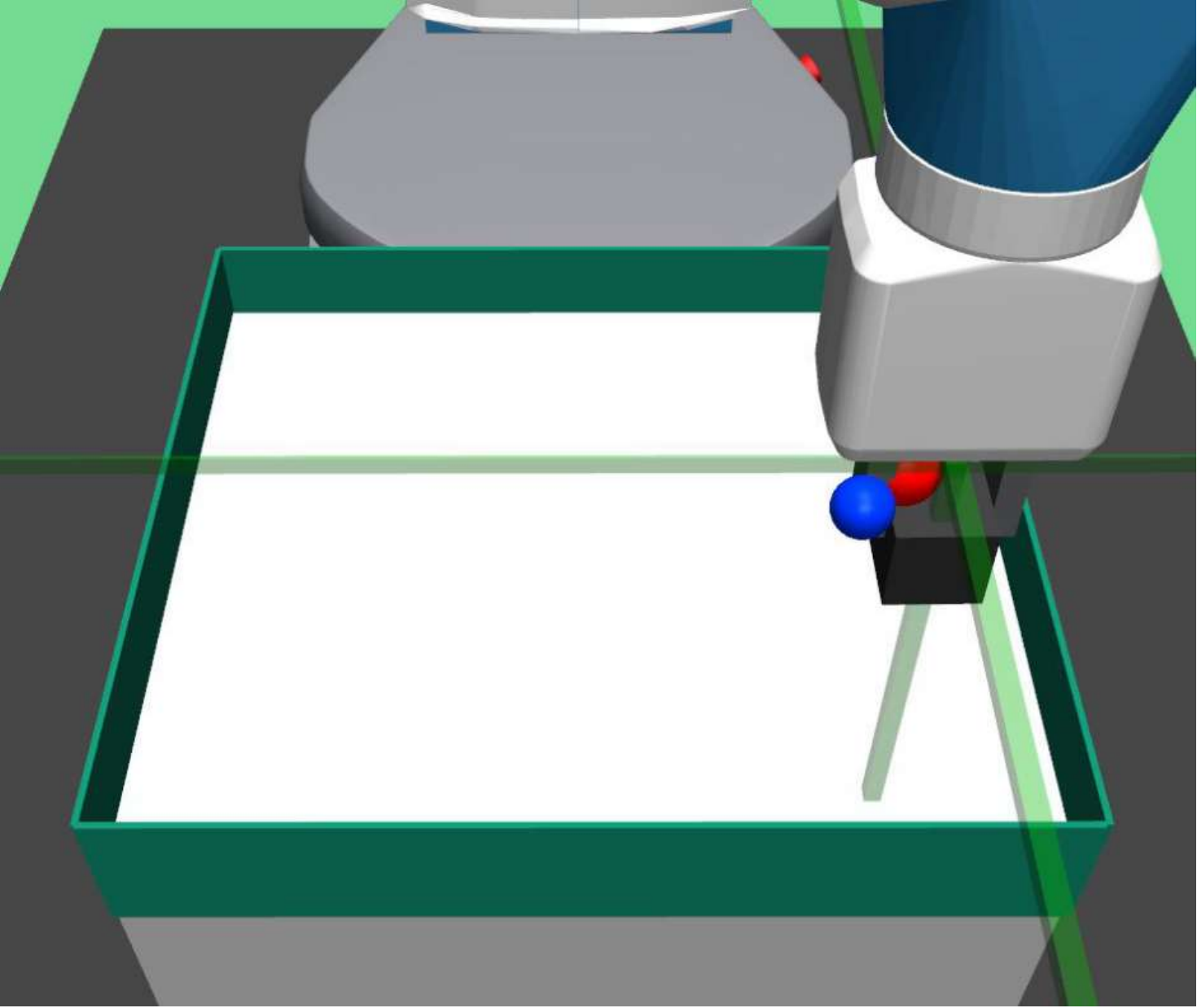}
\includegraphics[height=1.8cm,width=2.1cm]{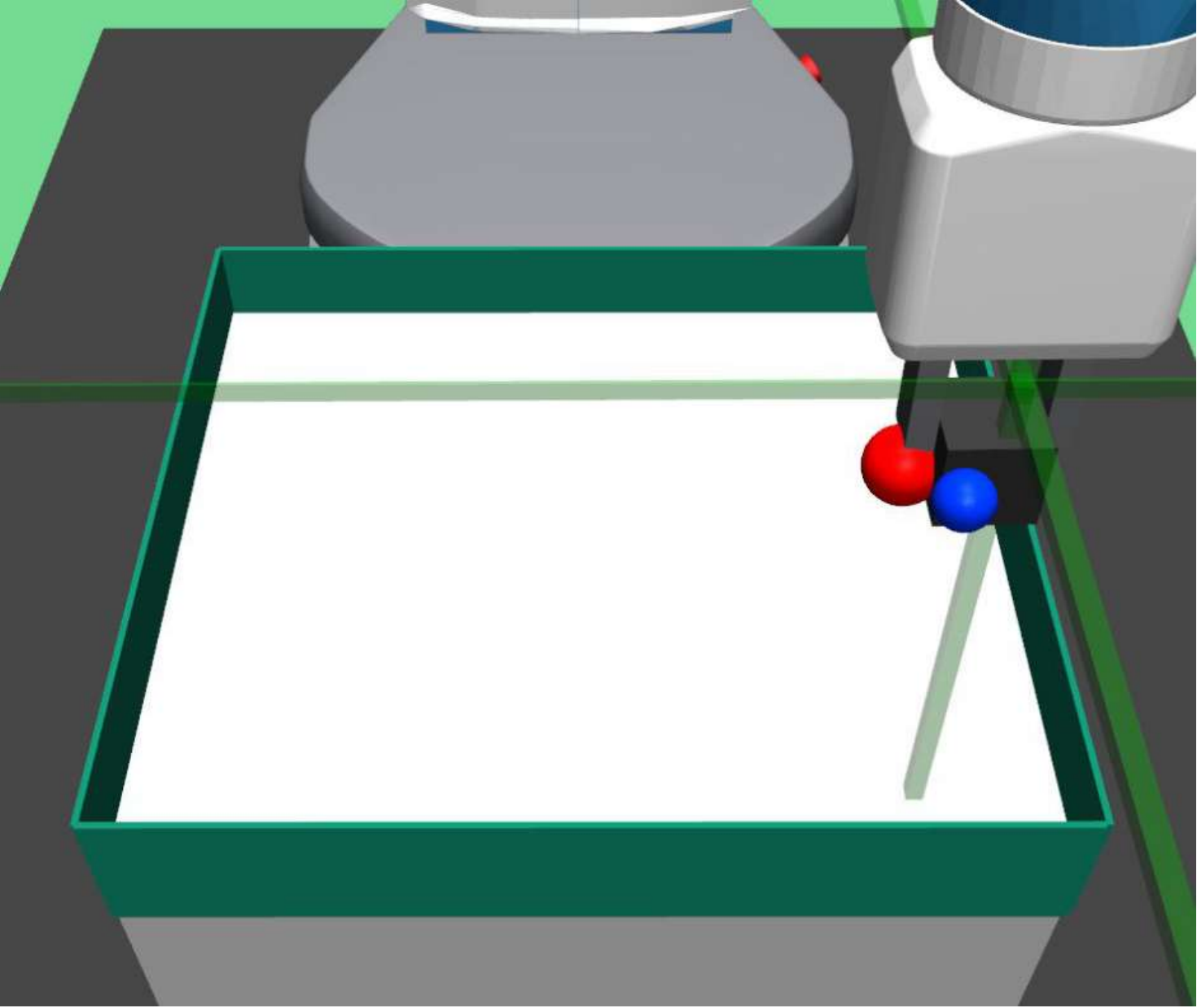}
\includegraphics[height=1.8cm,width=2.1cm]{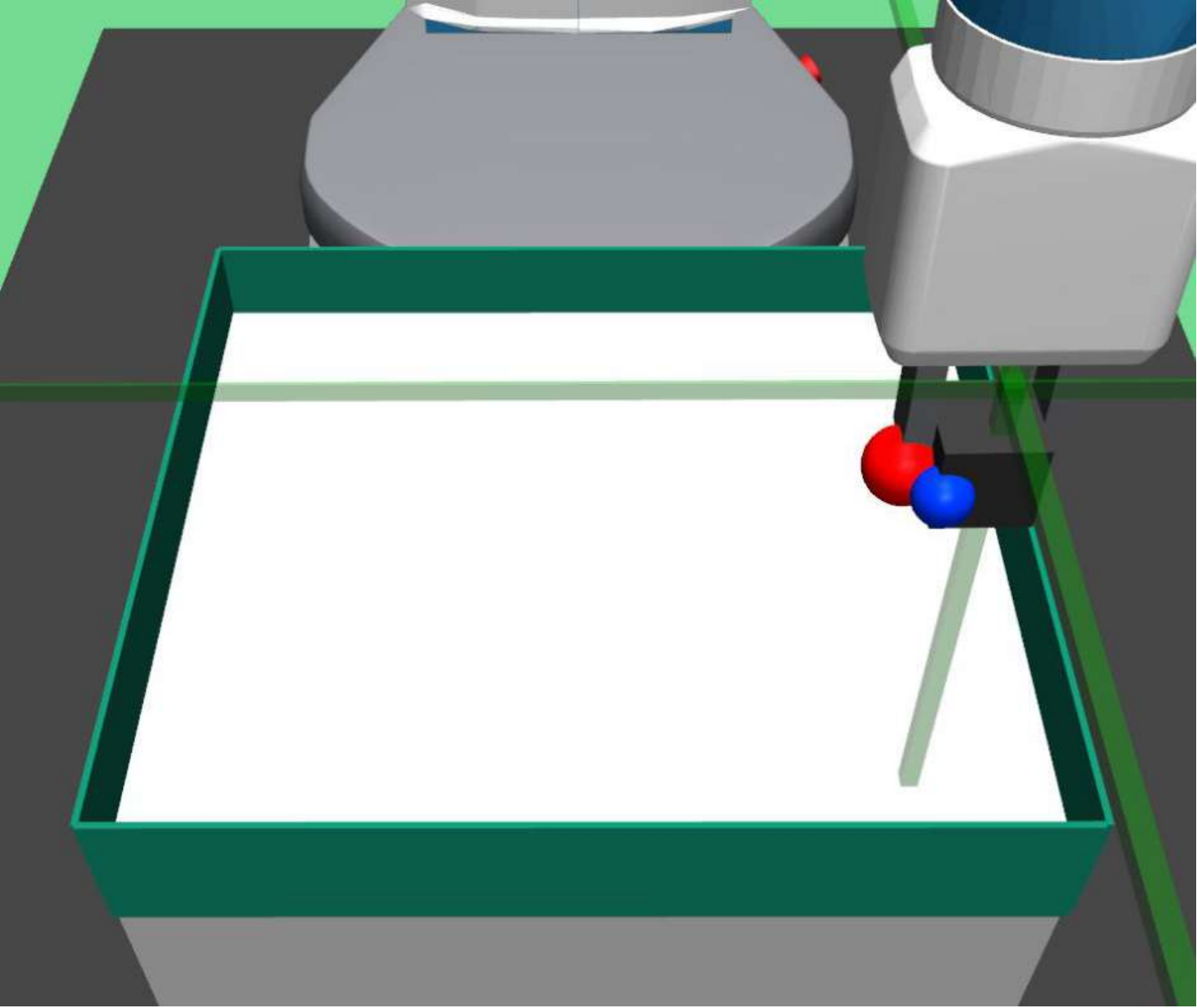}
\includegraphics[height=1.8cm,width=2.1cm]{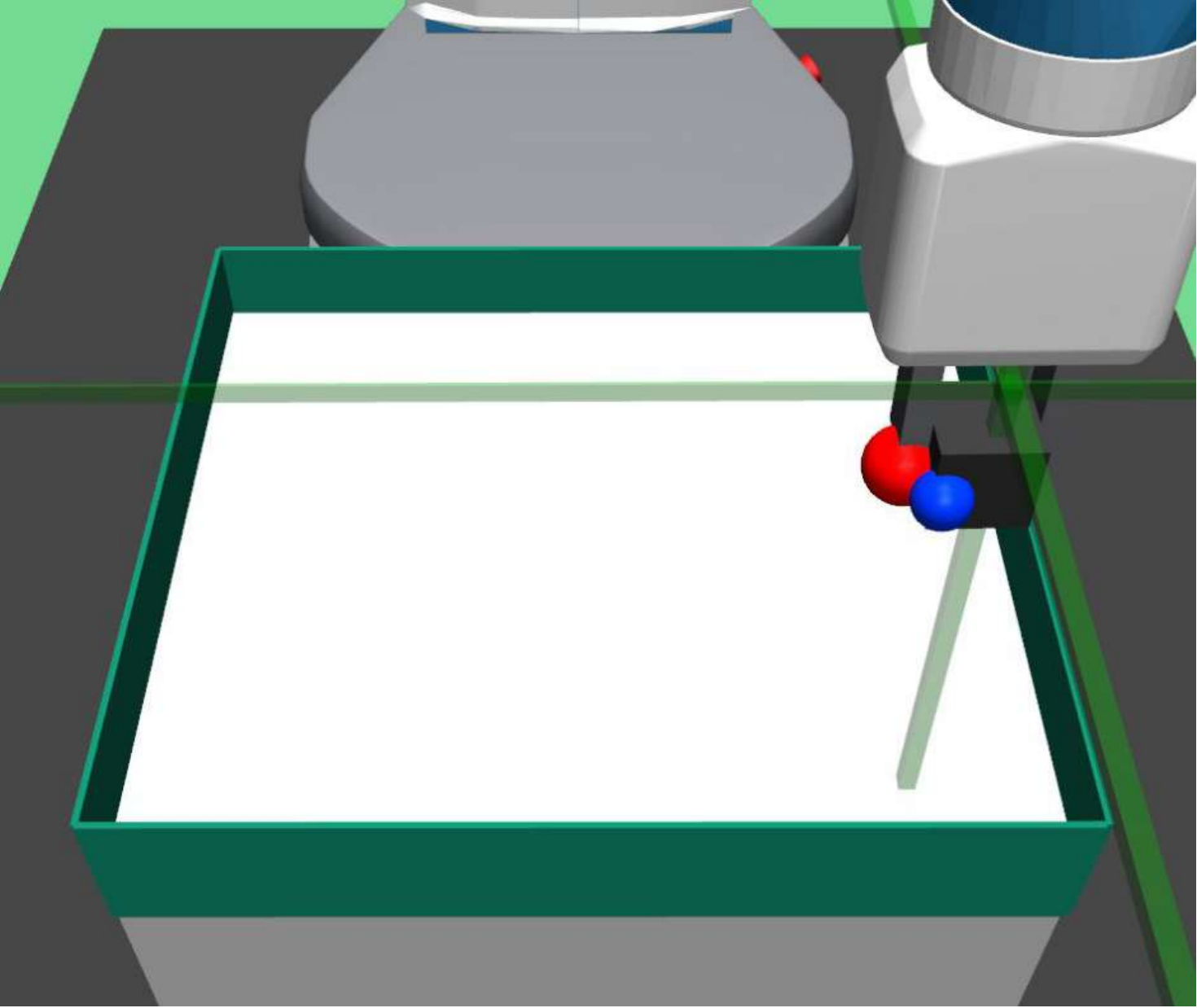}
\caption{The visualization is a successful attempt at performing pick navigation task}
\label{fig:pick_viz_success_1_ablation}
\end{figure}

%% file: figures_tex/bin_success_visualization_1.tex
\begin{figure}[H]
\vspace{5pt}
\centering
\captionsetup{font=footnotesize,labelfont=scriptsize,textfont=scriptsize}
\includegraphics[height=1.8cm,width=2.1cm]{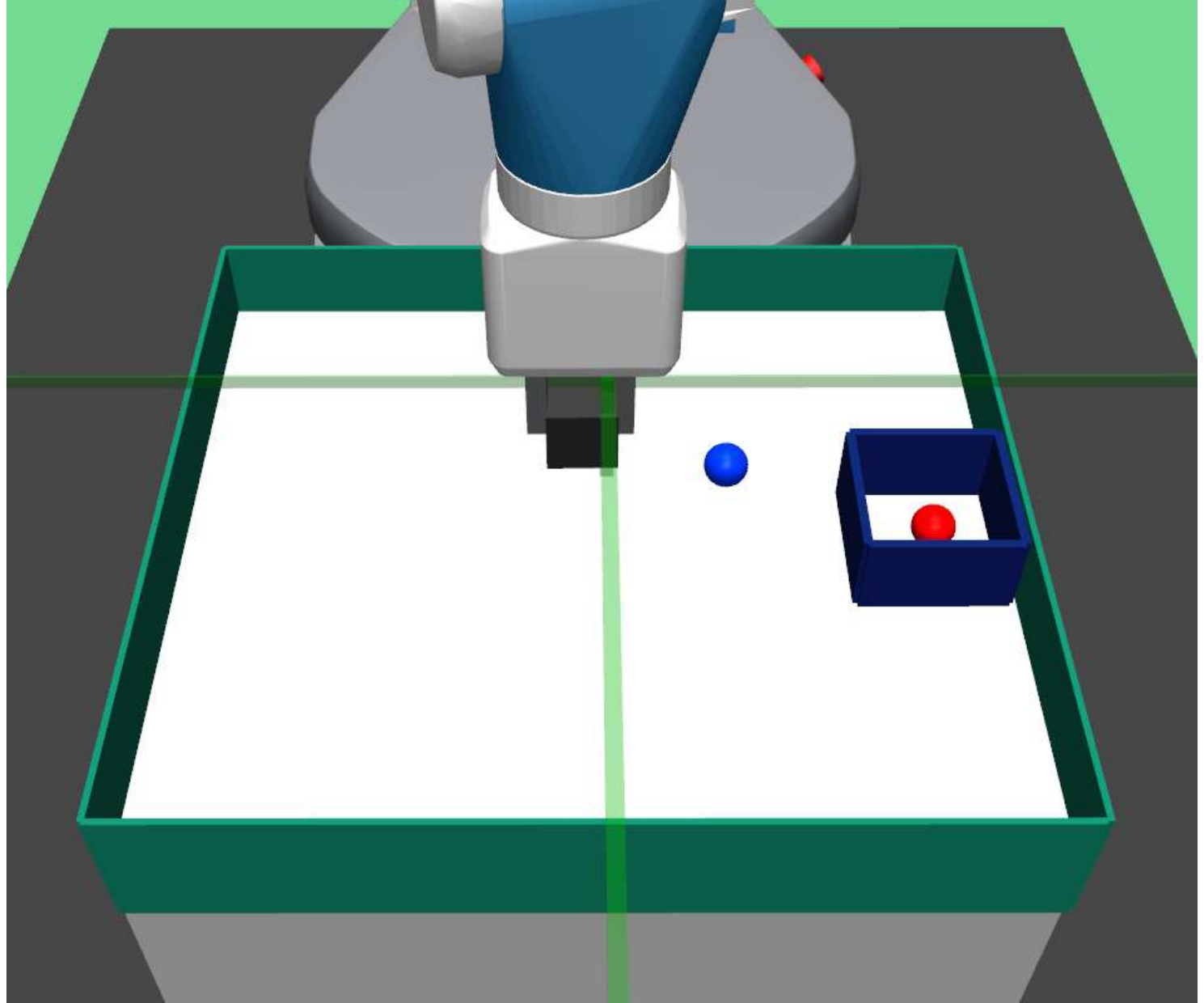}
\includegraphics[height=1.8cm,width=2.1cm]{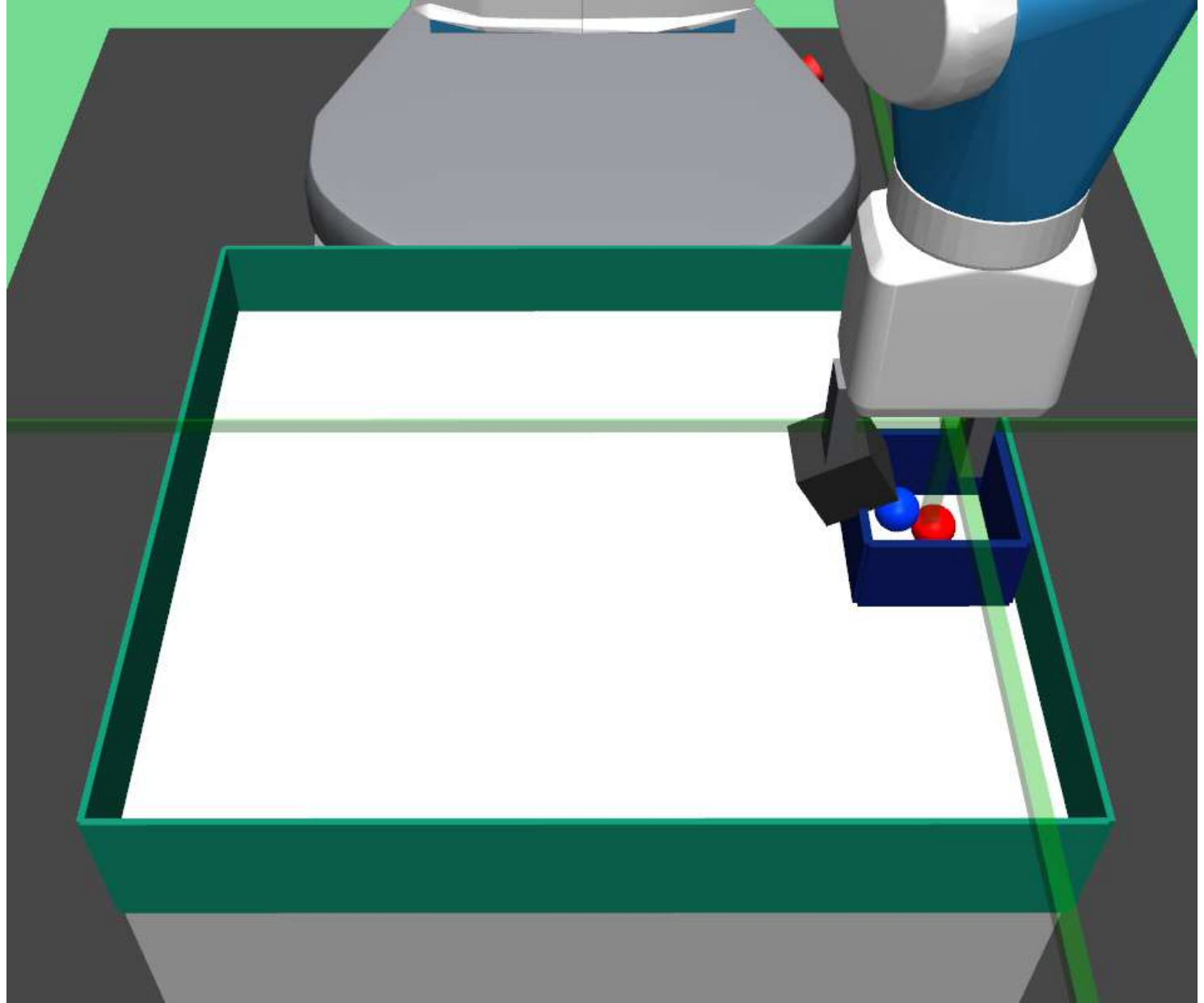}
\includegraphics[height=1.8cm,width=2.1cm]{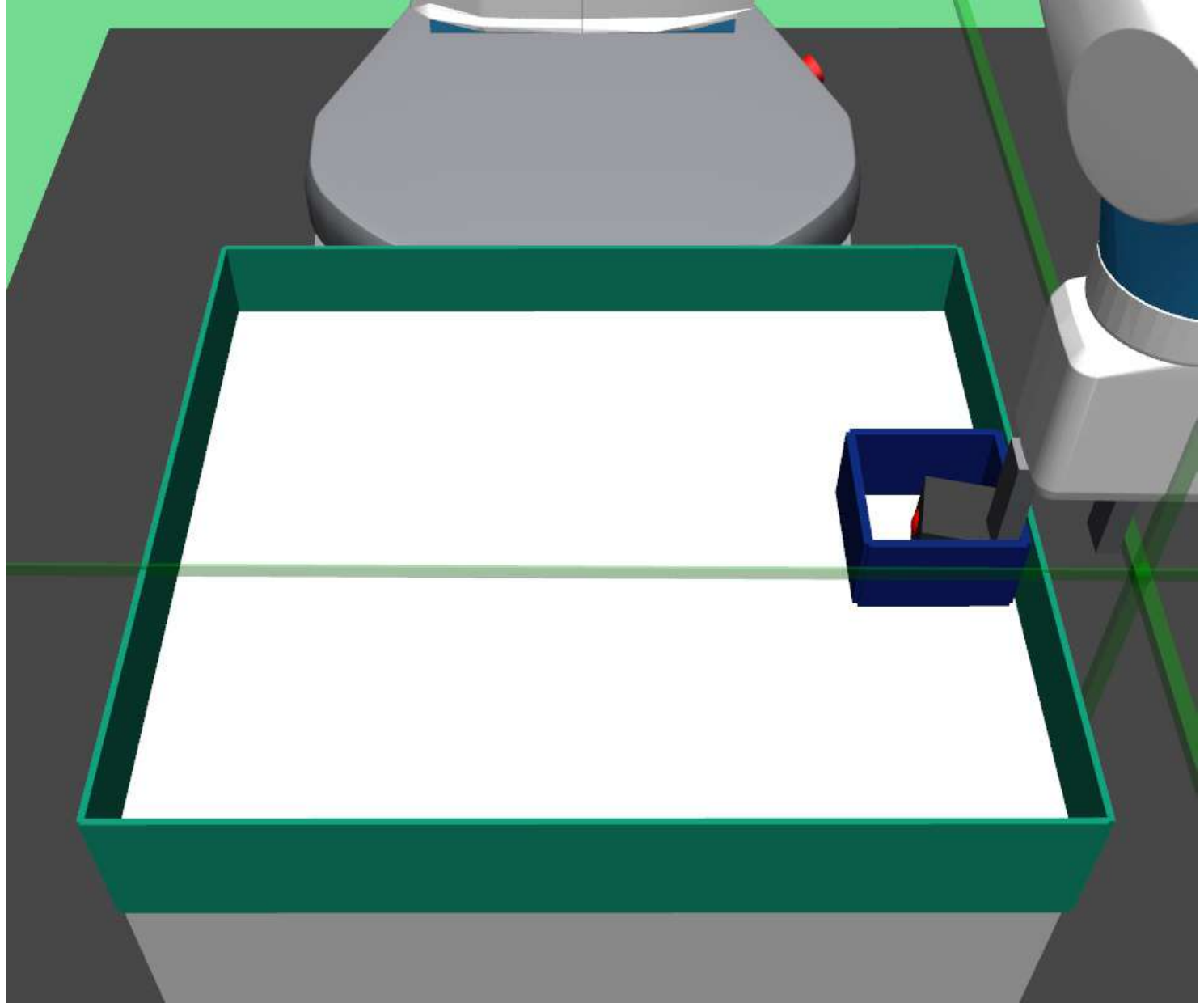}
\includegraphics[height=1.8cm,width=2.1cm]{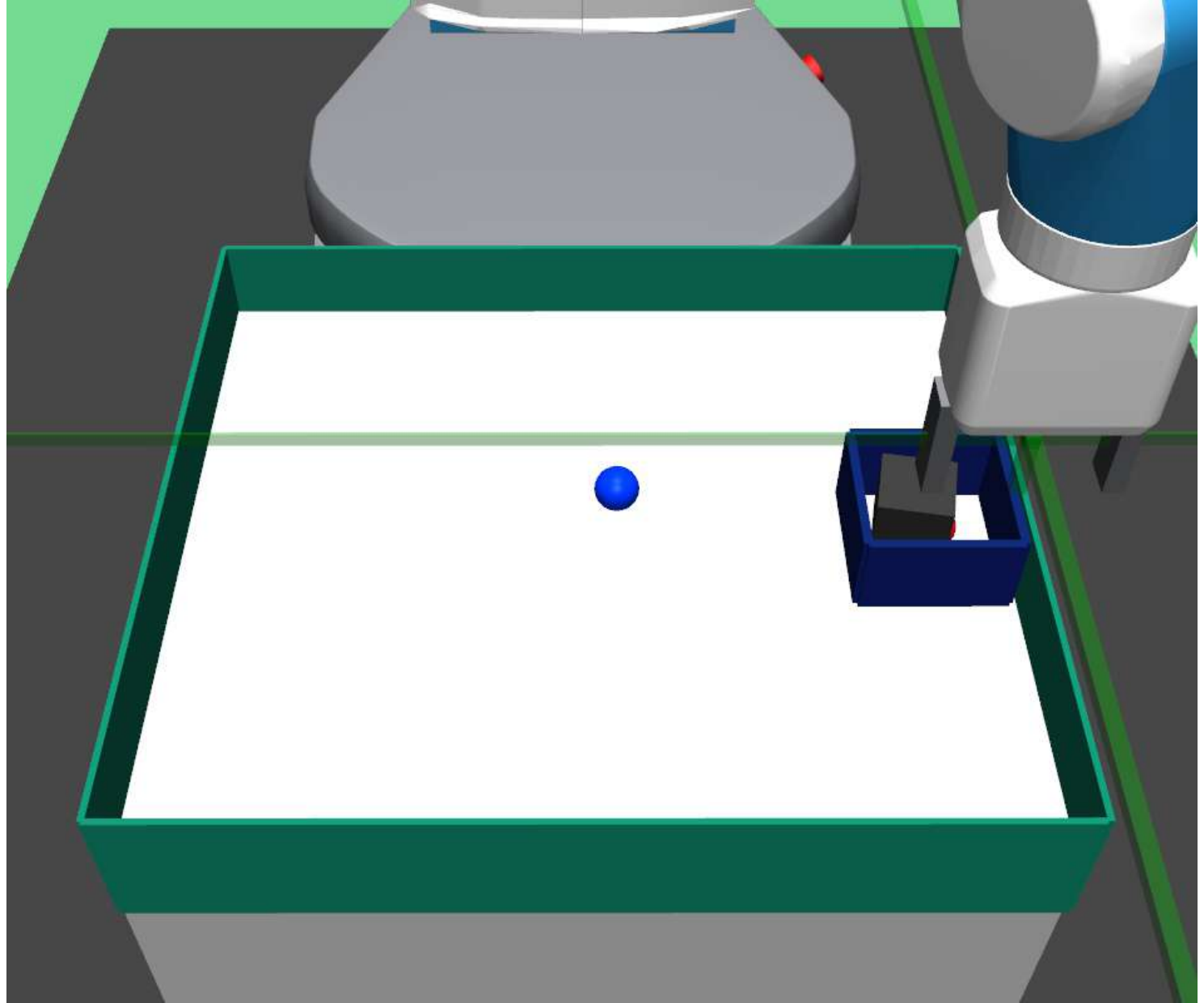}
\includegraphics[height=1.8cm,width=2.1cm]{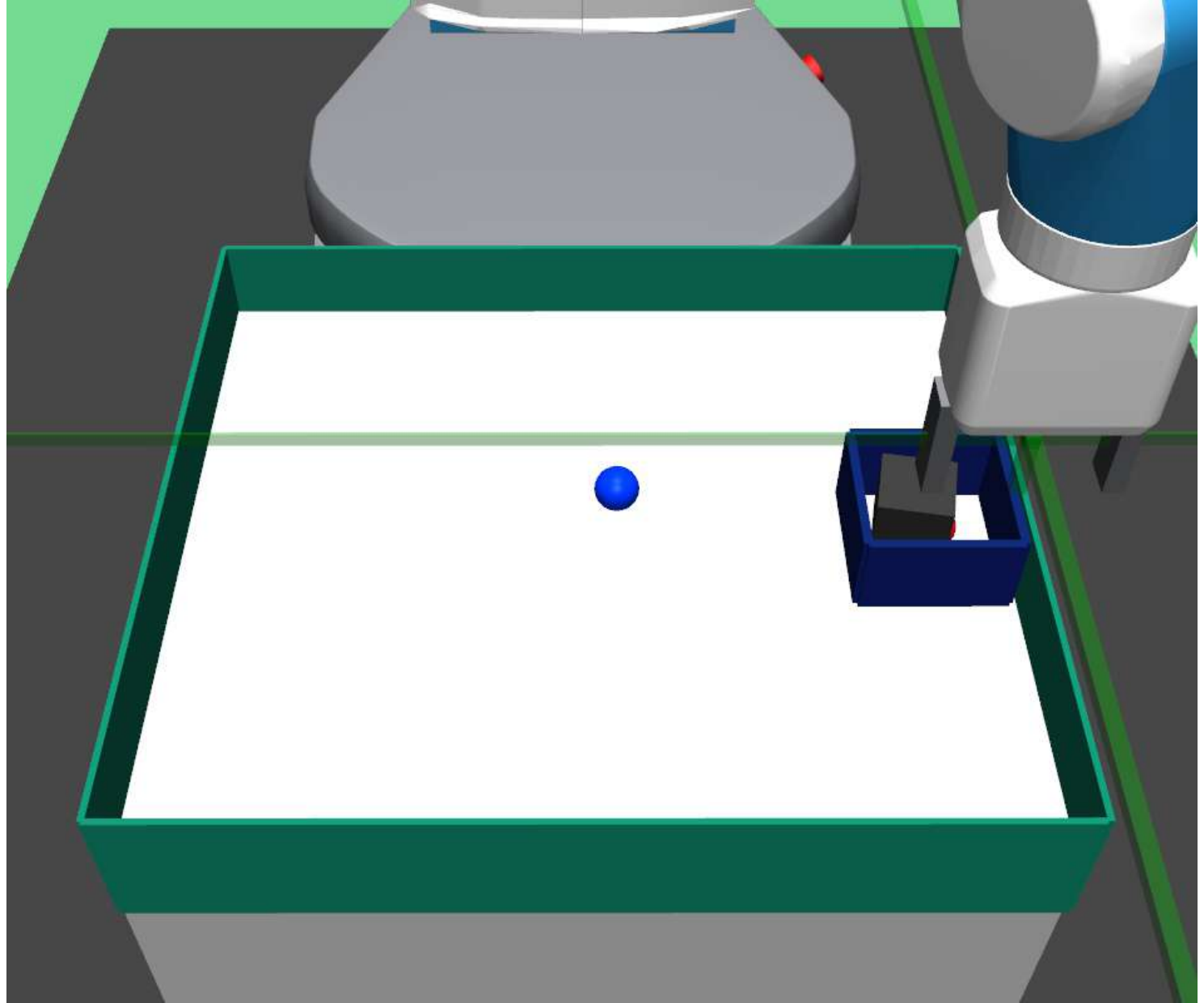}
\includegraphics[height=1.8cm,width=2.1cm]{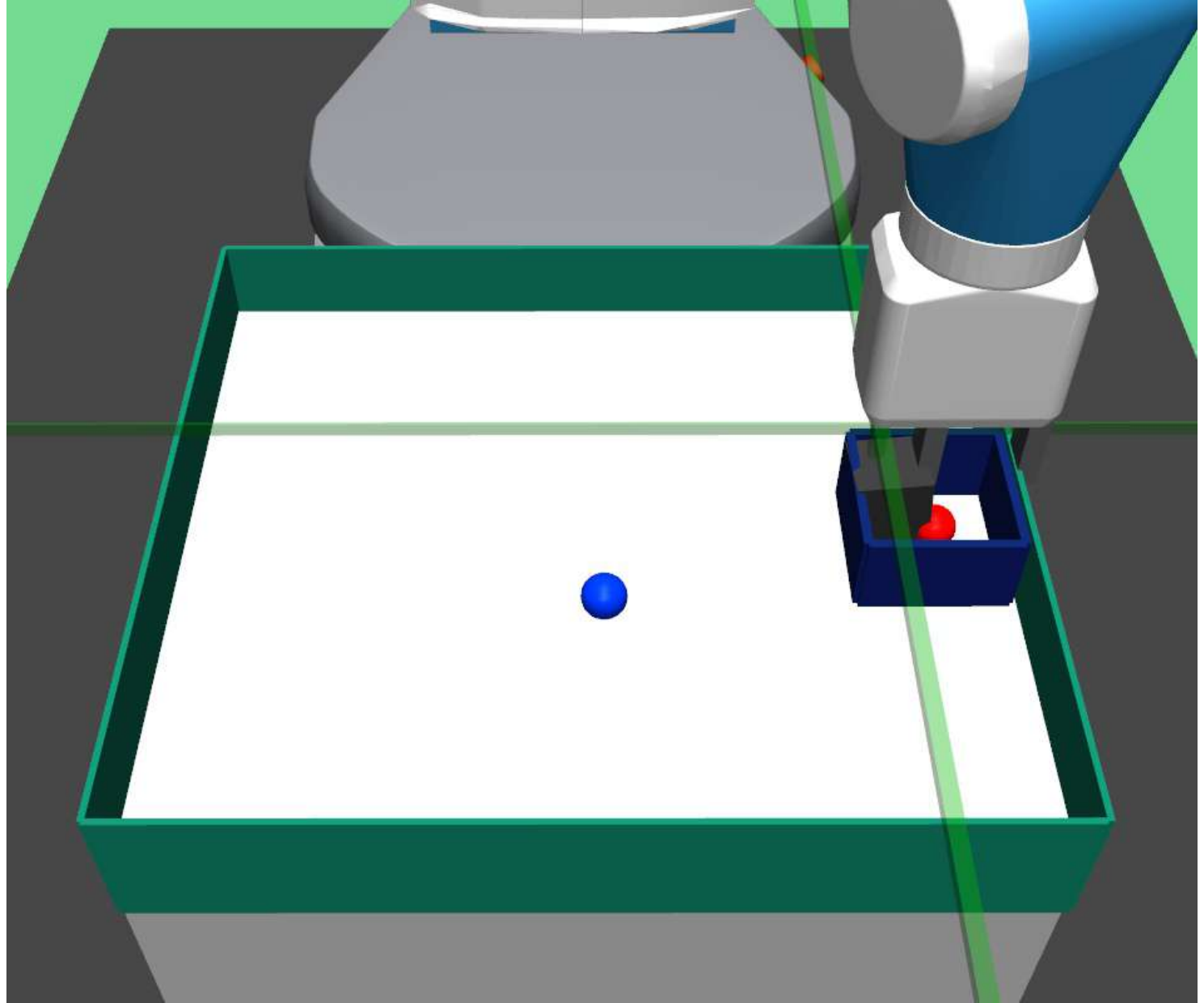}
\caption{The visualization is a successful attempt at performing bin task}
\label{fig:bin_viz_success_1_ablation}
\end{figure}

%% file: figures_tex/hollow_success_visualization_1.tex
\begin{figure}[H]
\vspace{5pt}
\centering
\captionsetup{font=footnotesize,labelfont=scriptsize,textfont=scriptsize}
\includegraphics[height=1.8cm,width=2.1cm]{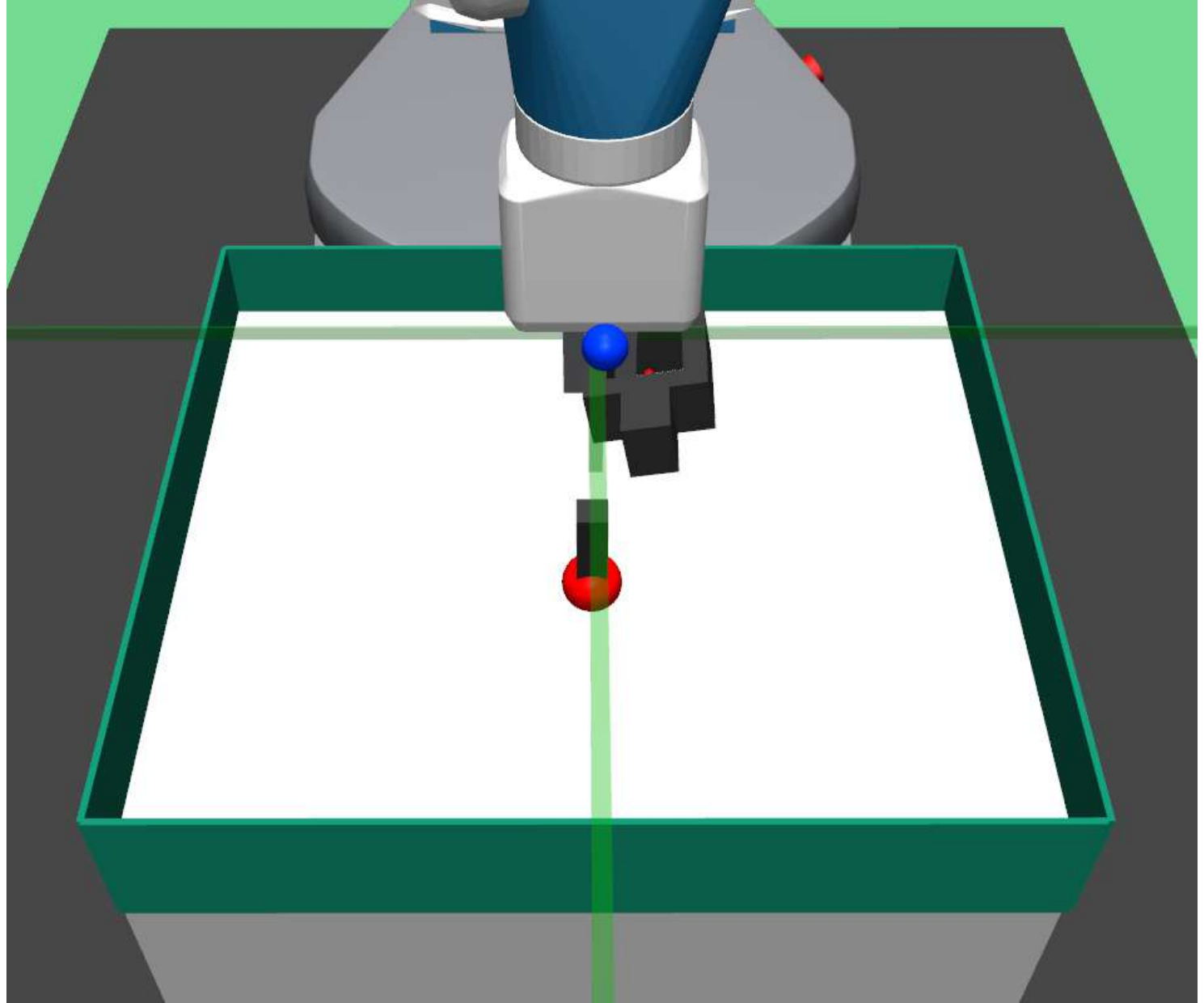}
\includegraphics[height=1.8cm,width=2.1cm]{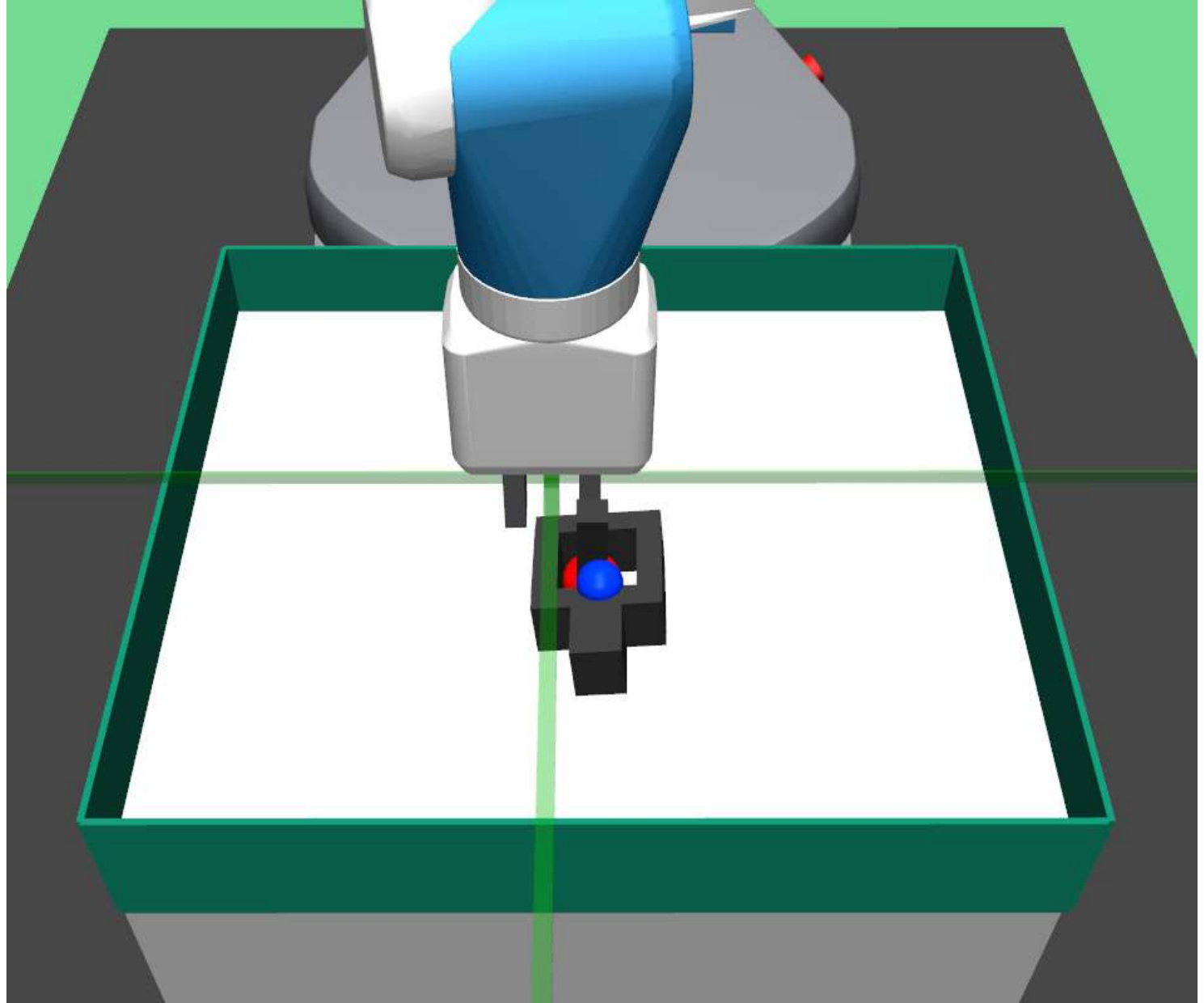}
\includegraphics[height=1.8cm,width=2.1cm]{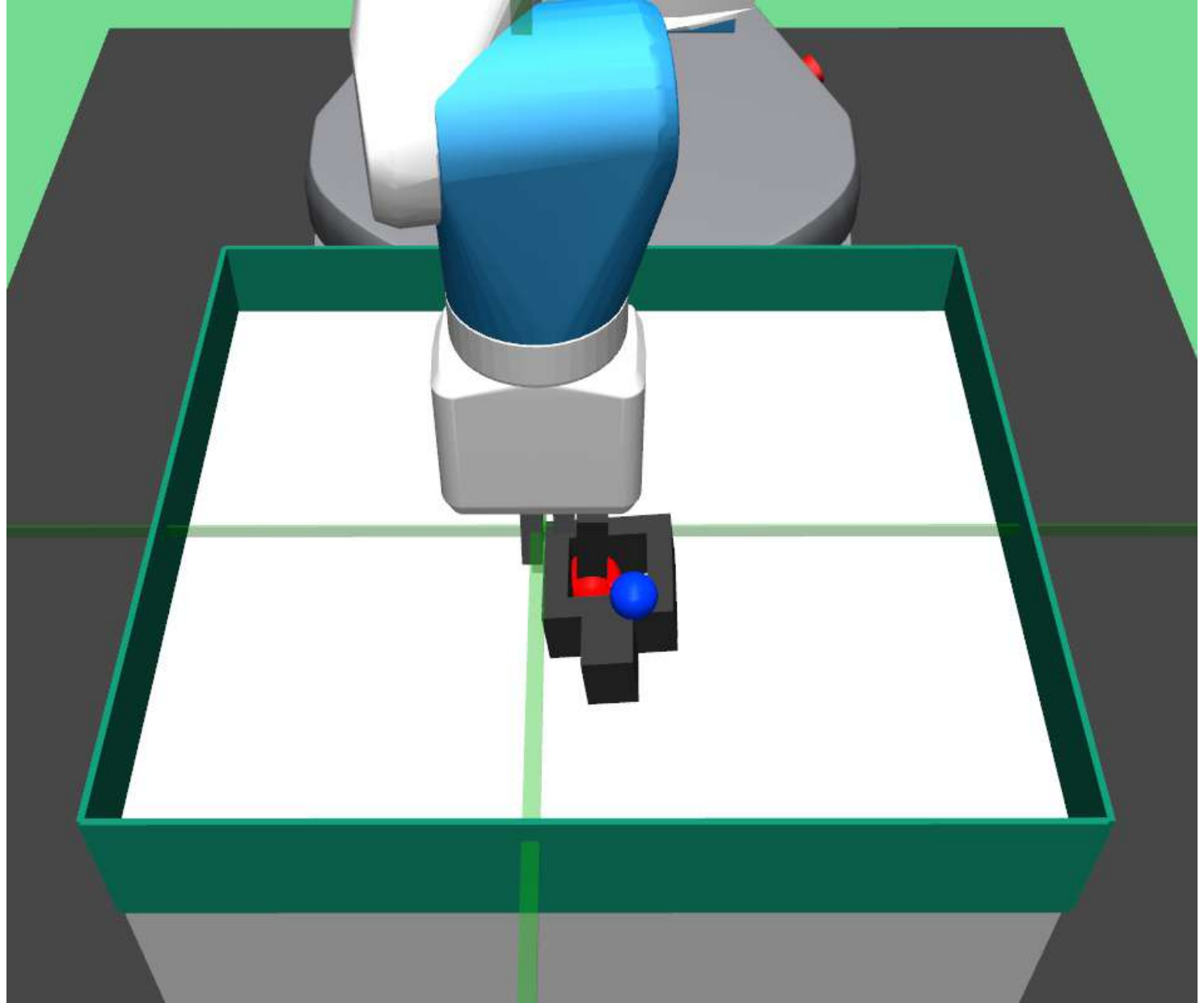}
\includegraphics[height=1.8cm,width=2.1cm]{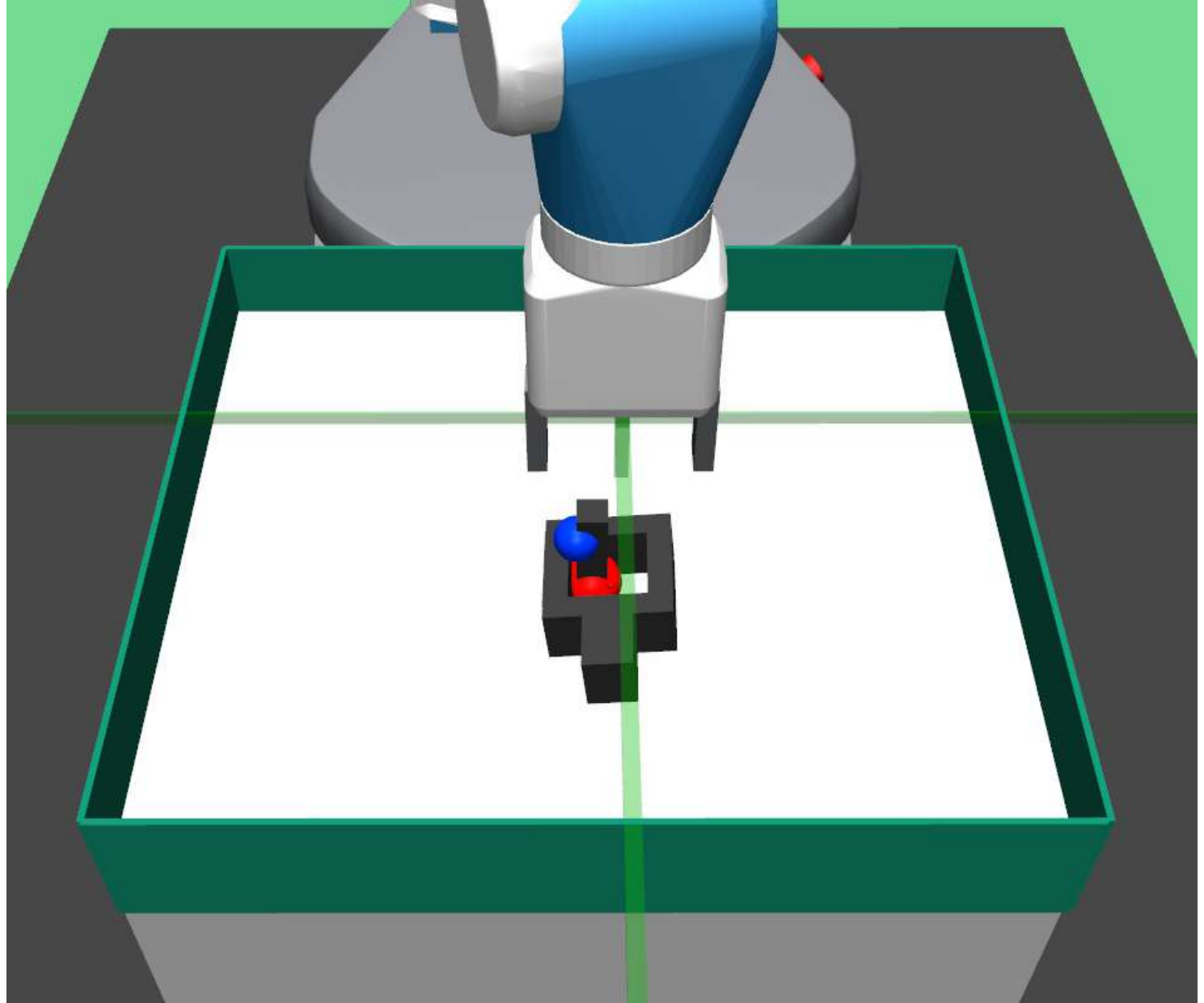}
\includegraphics[height=1.8cm,width=2.1cm]{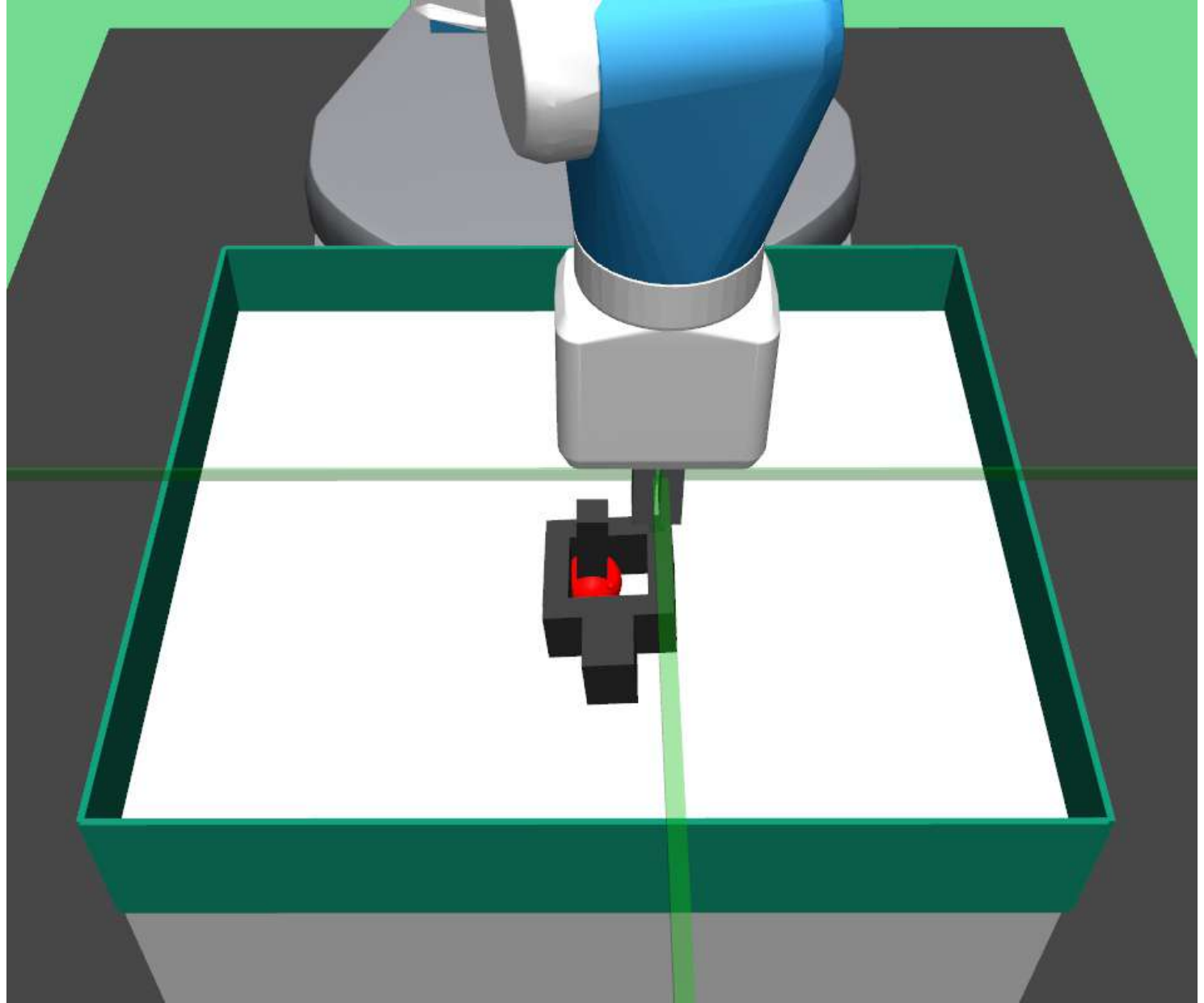}
\includegraphics[height=1.8cm,width=2.1cm]{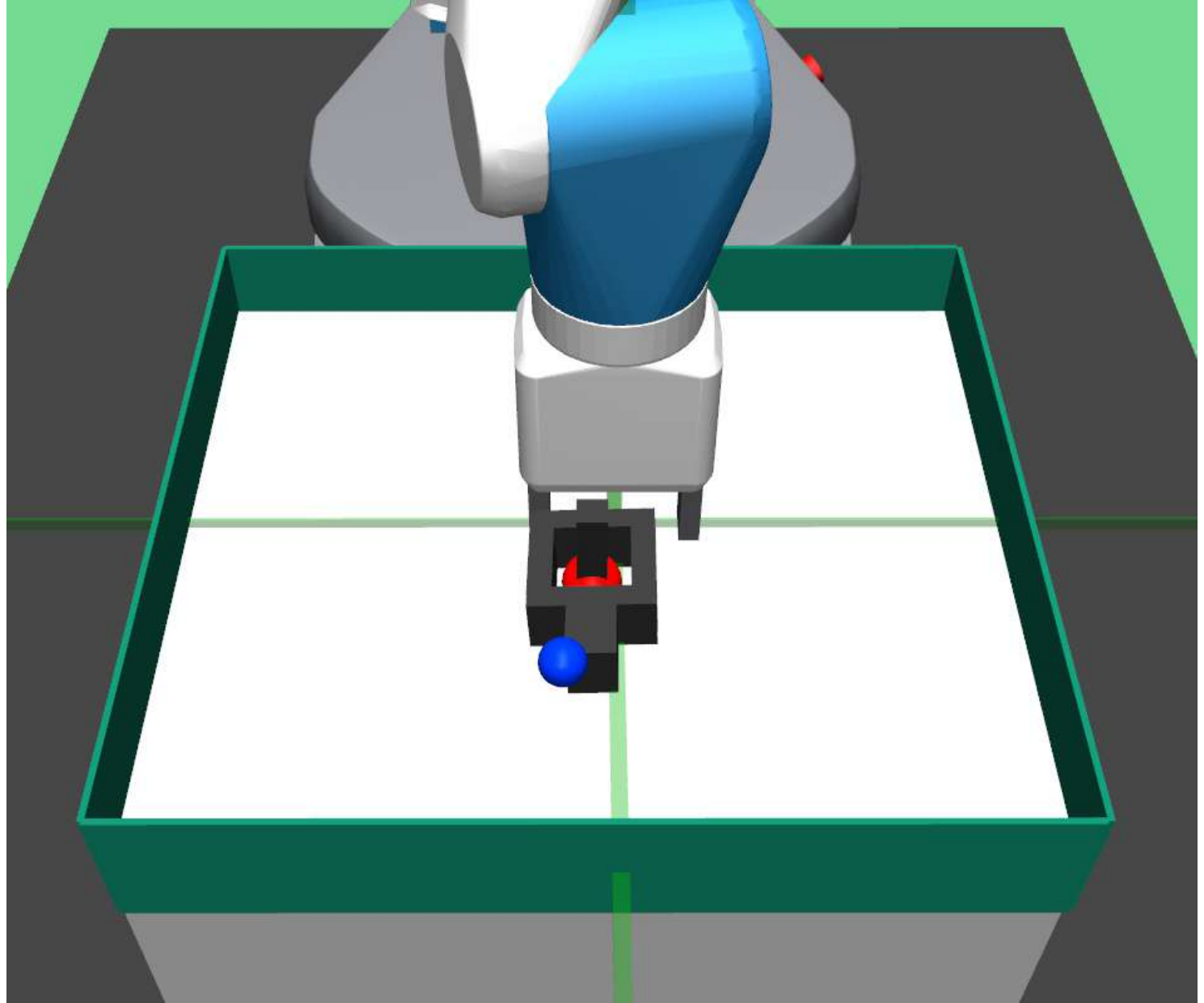}
\caption{\textbf{Successful visualization}: The visualization is a successful attempt at performing hollow task}
\label{fig:hollow_viz_success_1_ablation}
\end{figure}

%% file: figures_tex/rope_success_visualization_1.tex
\begin{figure}[H]
\vspace{5pt}
\centering
\captionsetup{font=footnotesize,labelfont=scriptsize,textfont=scriptsize}
\includegraphics[height=1.8cm,width=2.1cm]{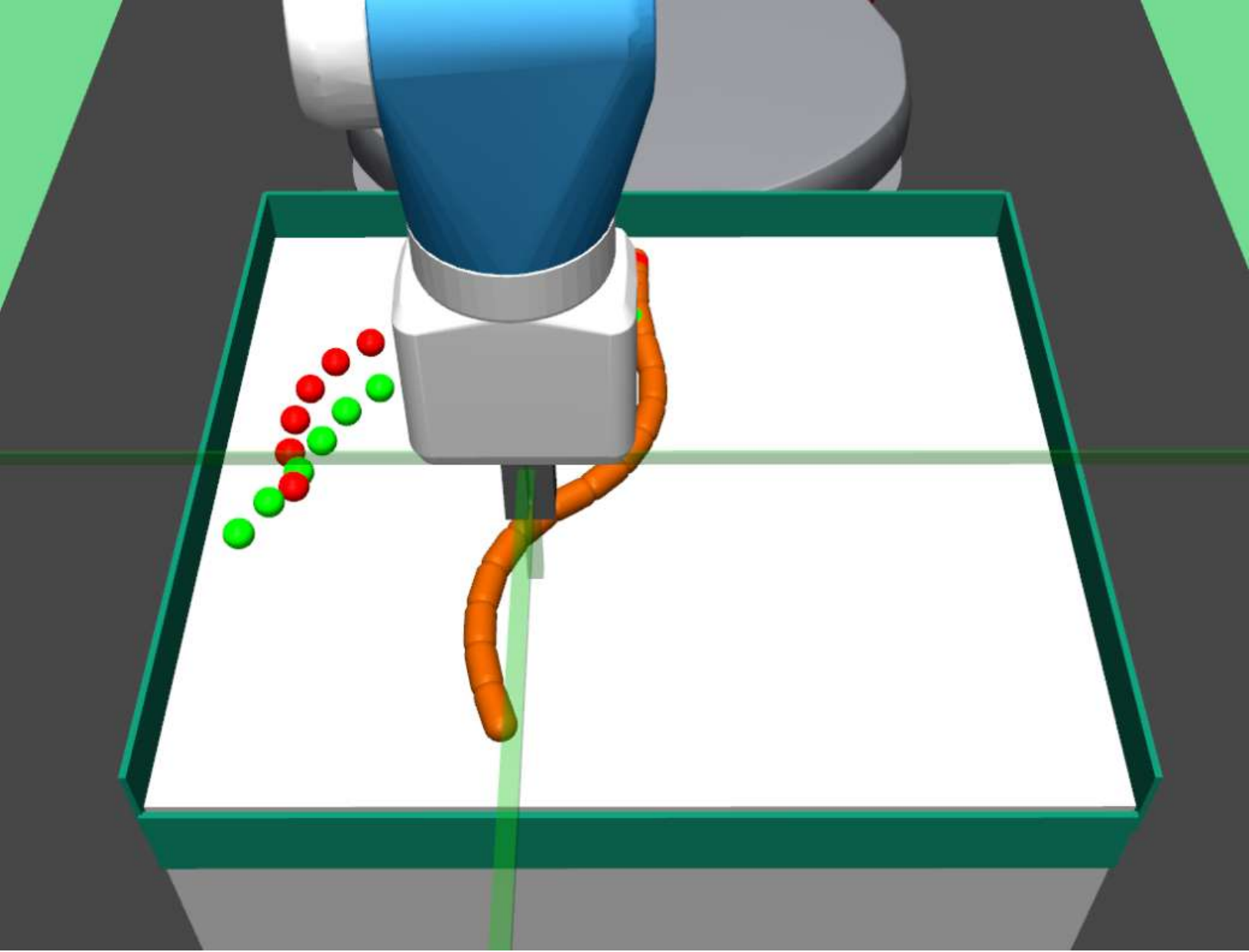}
\includegraphics[height=1.8cm,width=2.1cm]{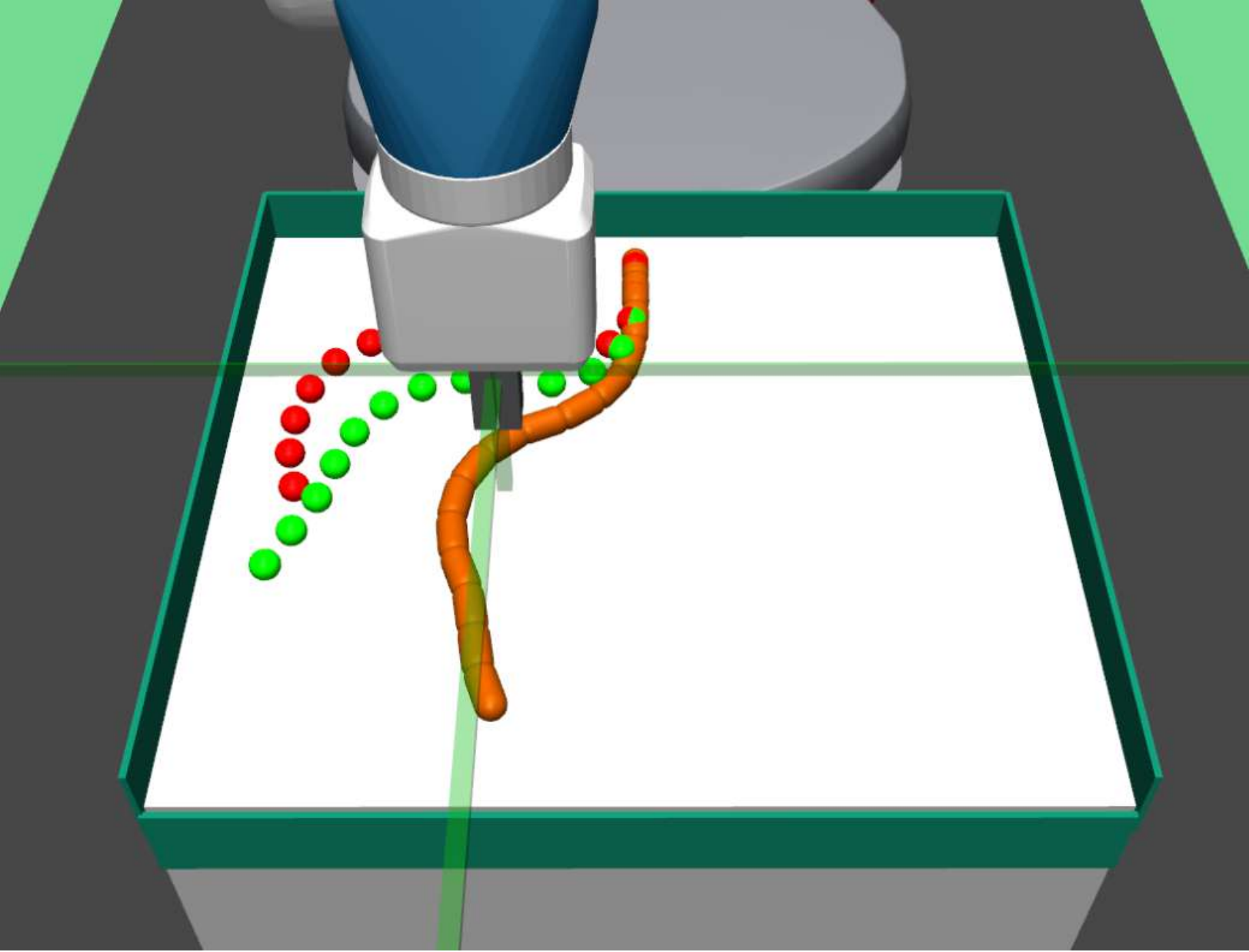}
\includegraphics[height=1.8cm,width=2.1cm]{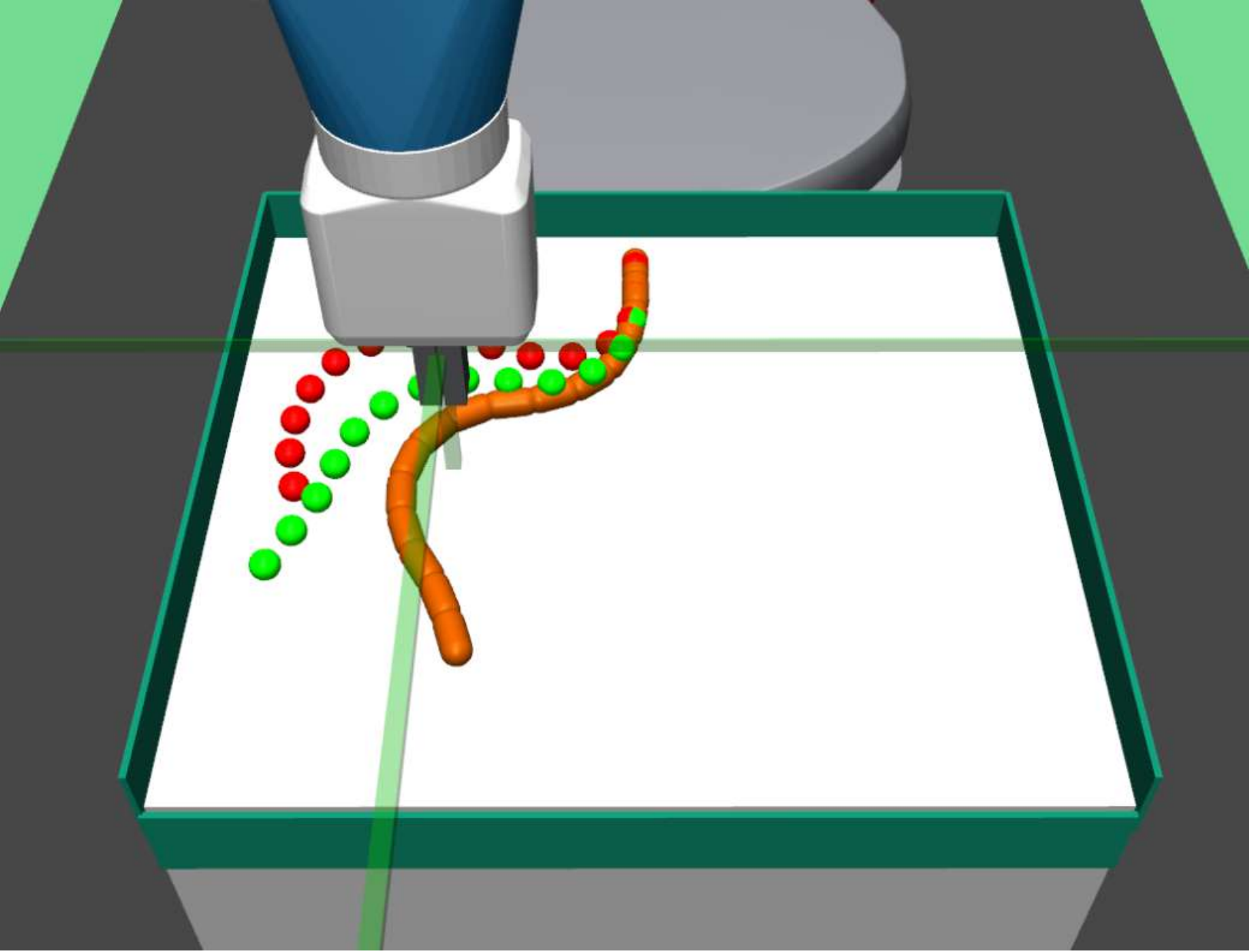}
\includegraphics[height=1.8cm,width=2.1cm]{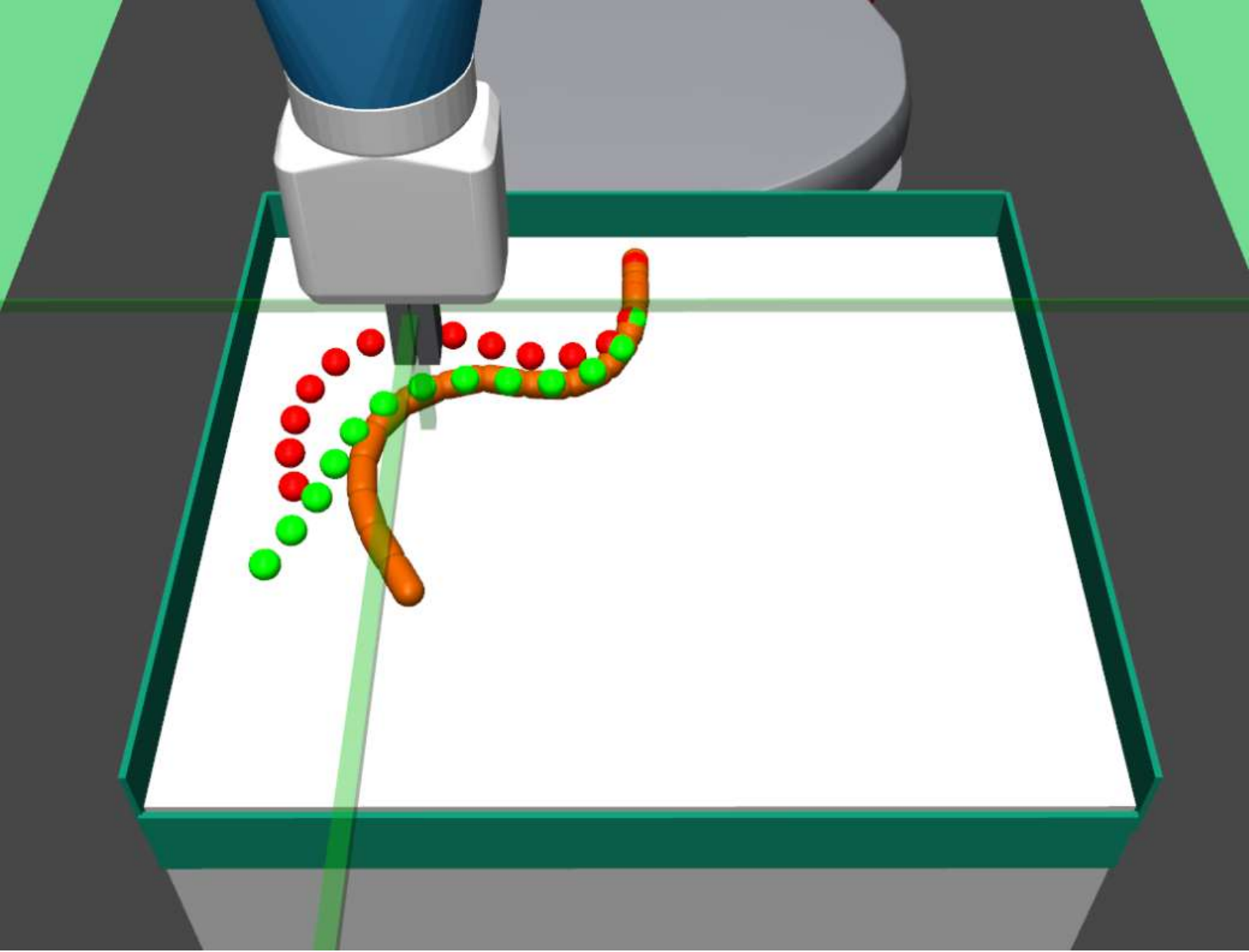}
\includegraphics[height=1.8cm,width=2.1cm]{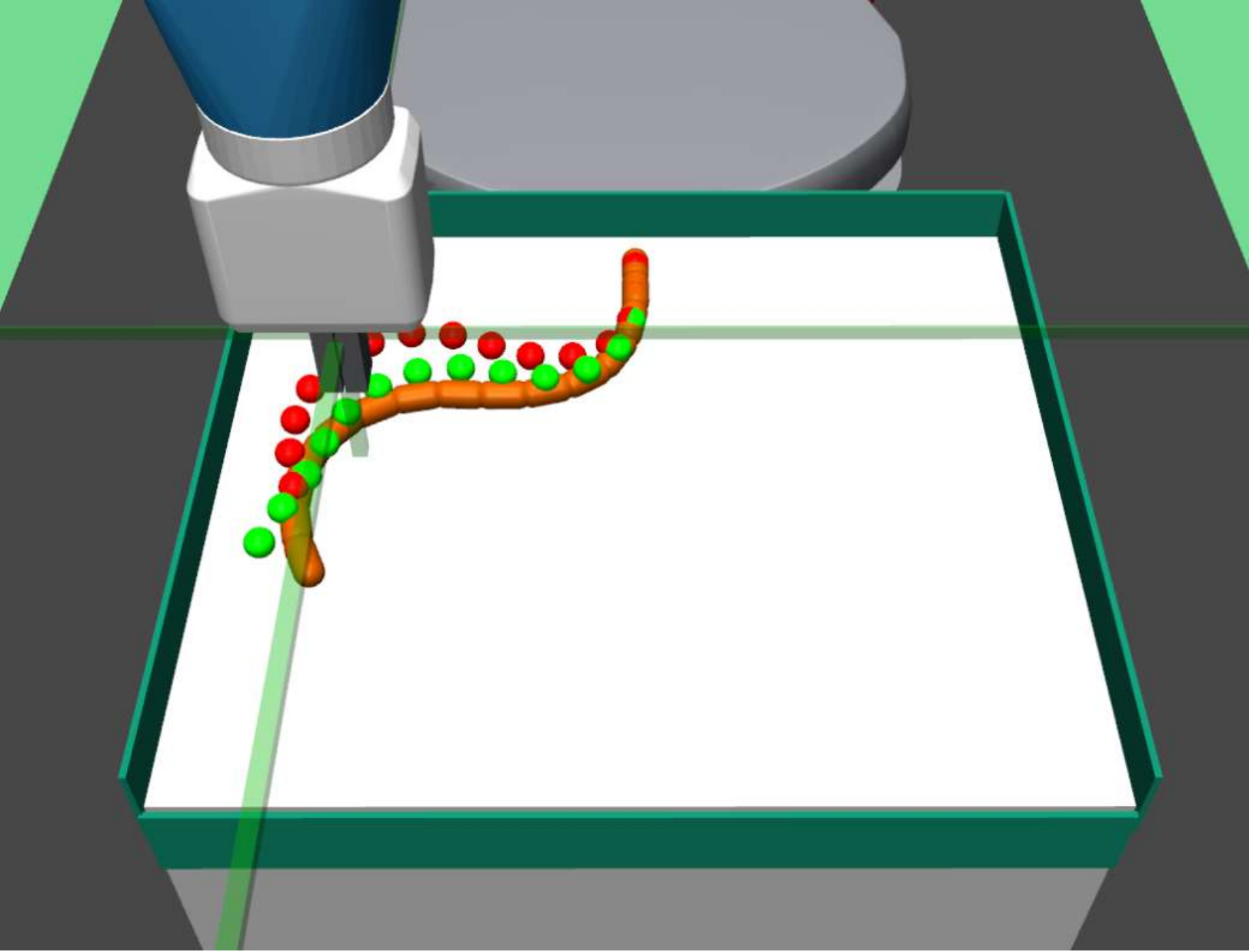}
\includegraphics[height=1.8cm,width=2.1cm]{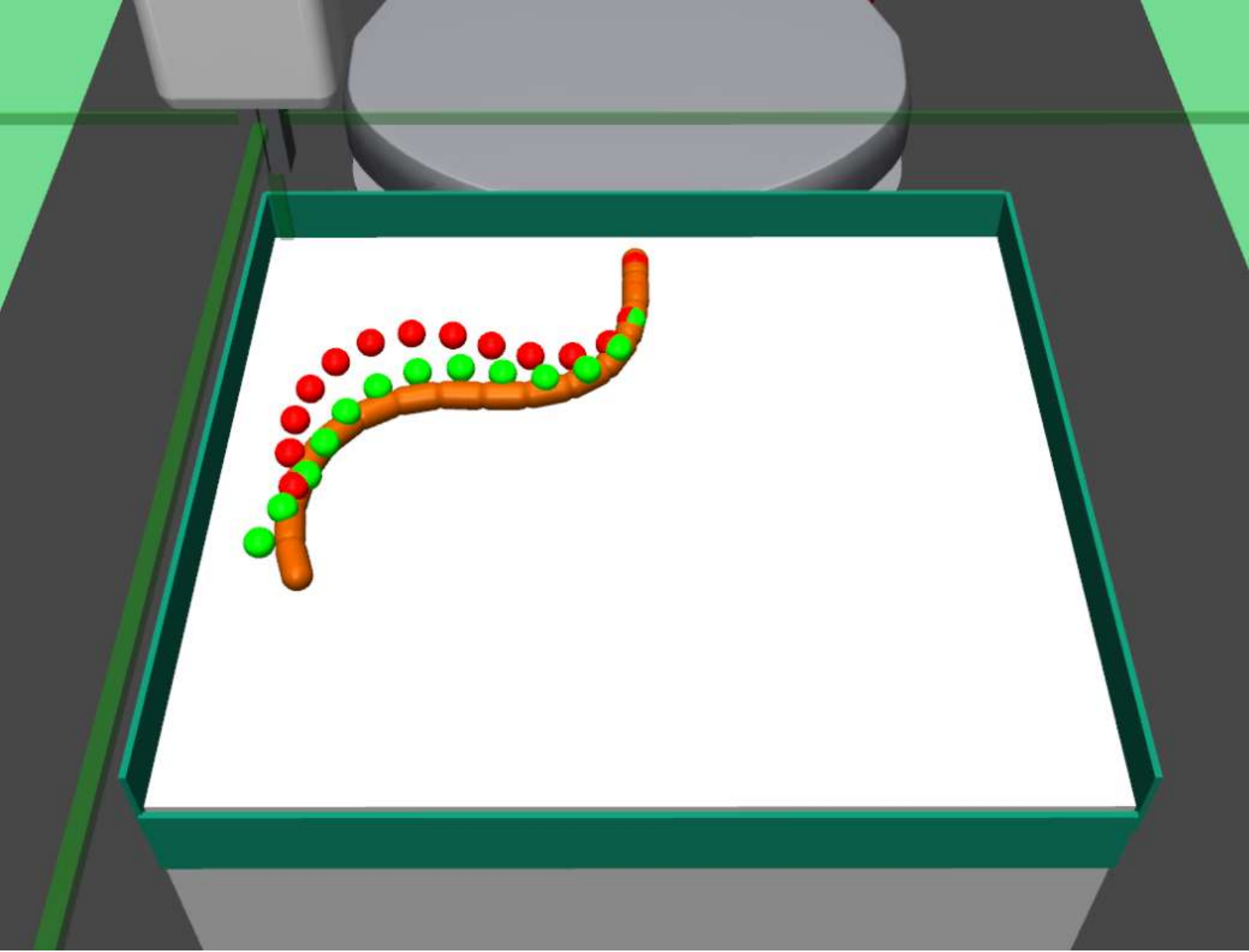}
\caption{The visualization is a successful attempt at performing rope navigation task}
\label{fig:rope_viz_success_1_ablation}
\end{figure}

%% file: figures_tex/kitchen_success_visualization_1.tex
\begin{figure}[H]
\vspace{5pt}
\centering
\captionsetup{font=footnotesize,labelfont=scriptsize,textfont=scriptsize}
\includegraphics[height=1.8cm,width=2.1cm]{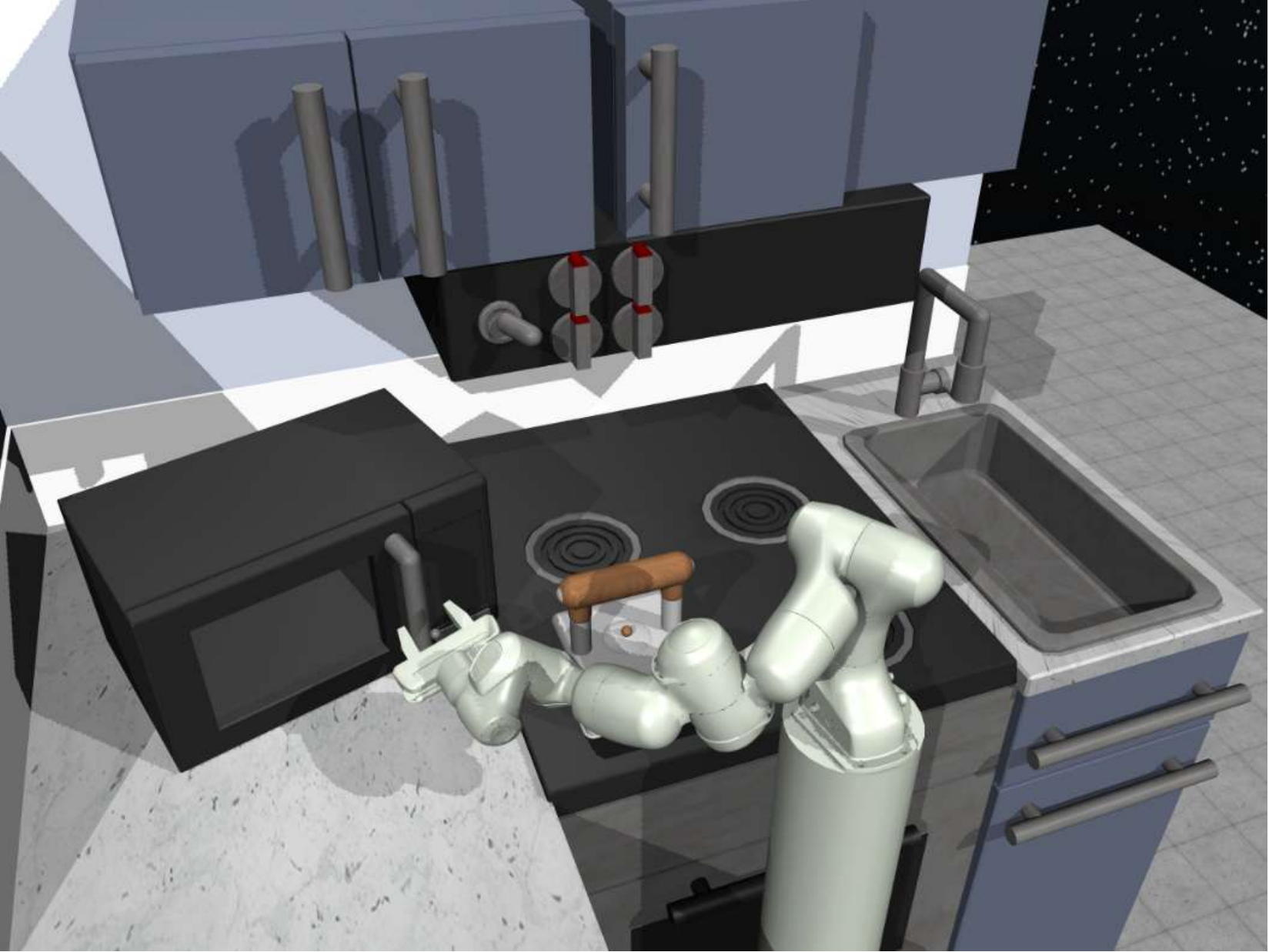}
\includegraphics[height=1.8cm,width=2.1cm]{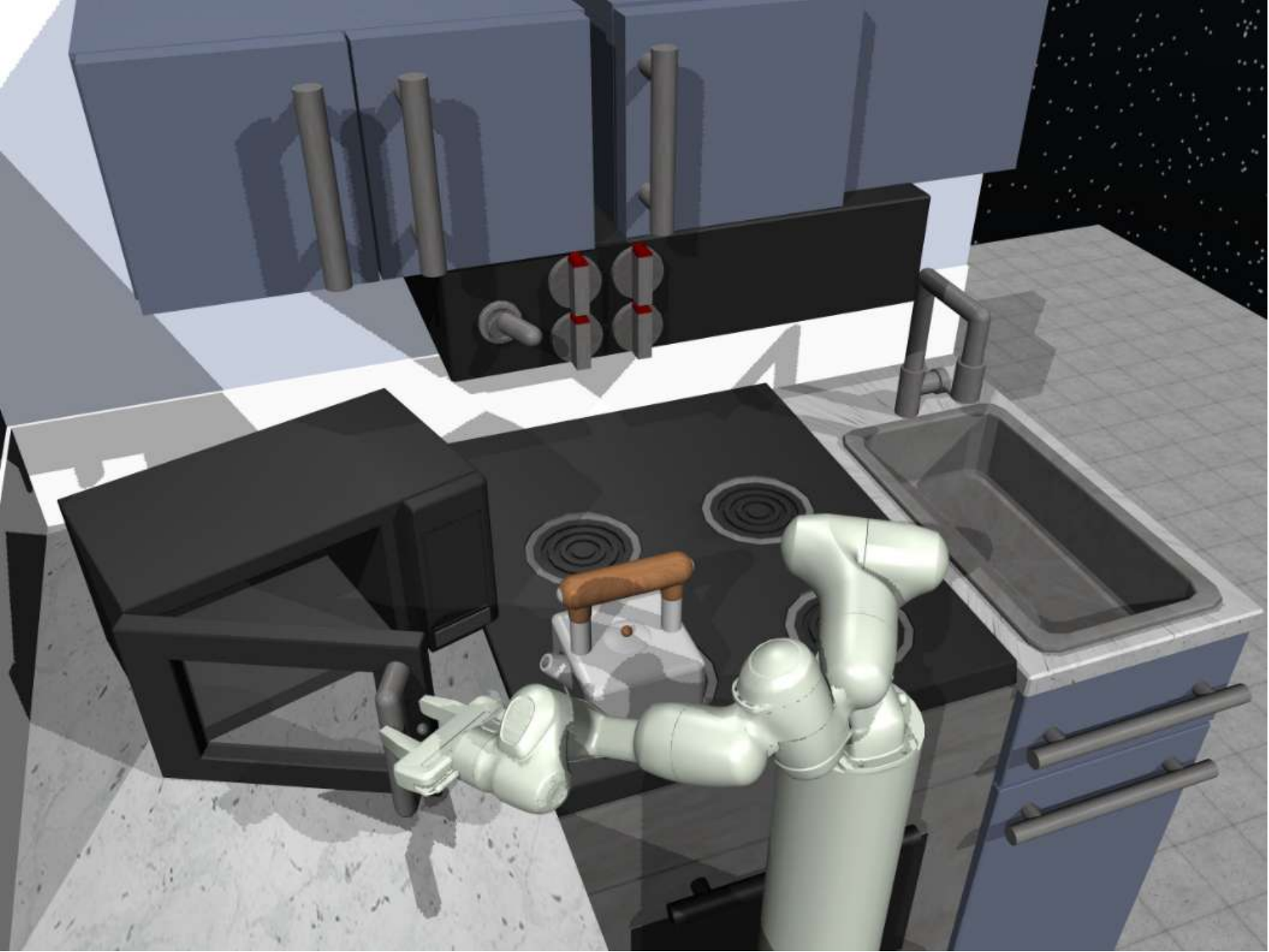}
\includegraphics[height=1.8cm,width=2.1cm]{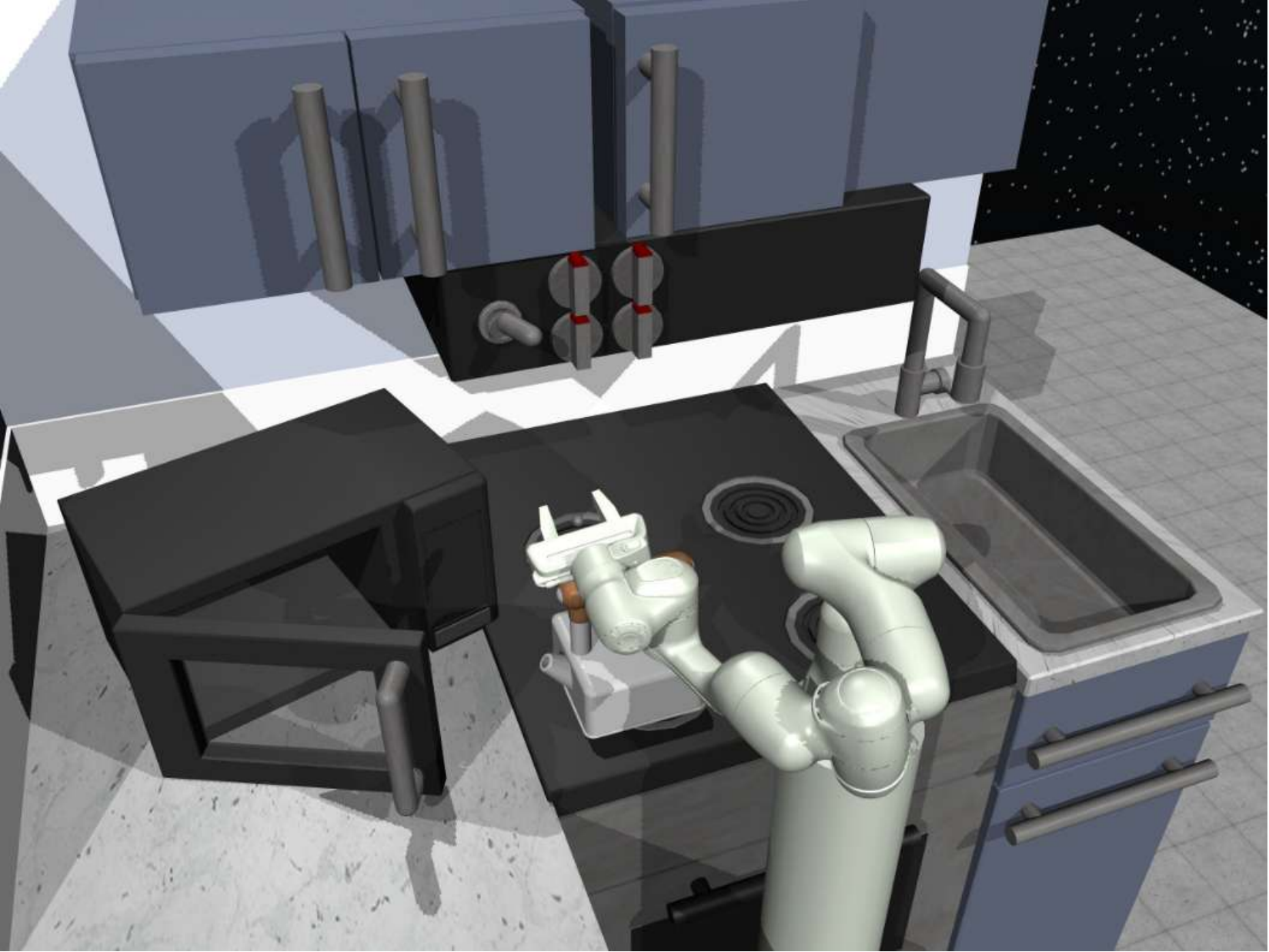}
\includegraphics[height=1.8cm,width=2.1cm]{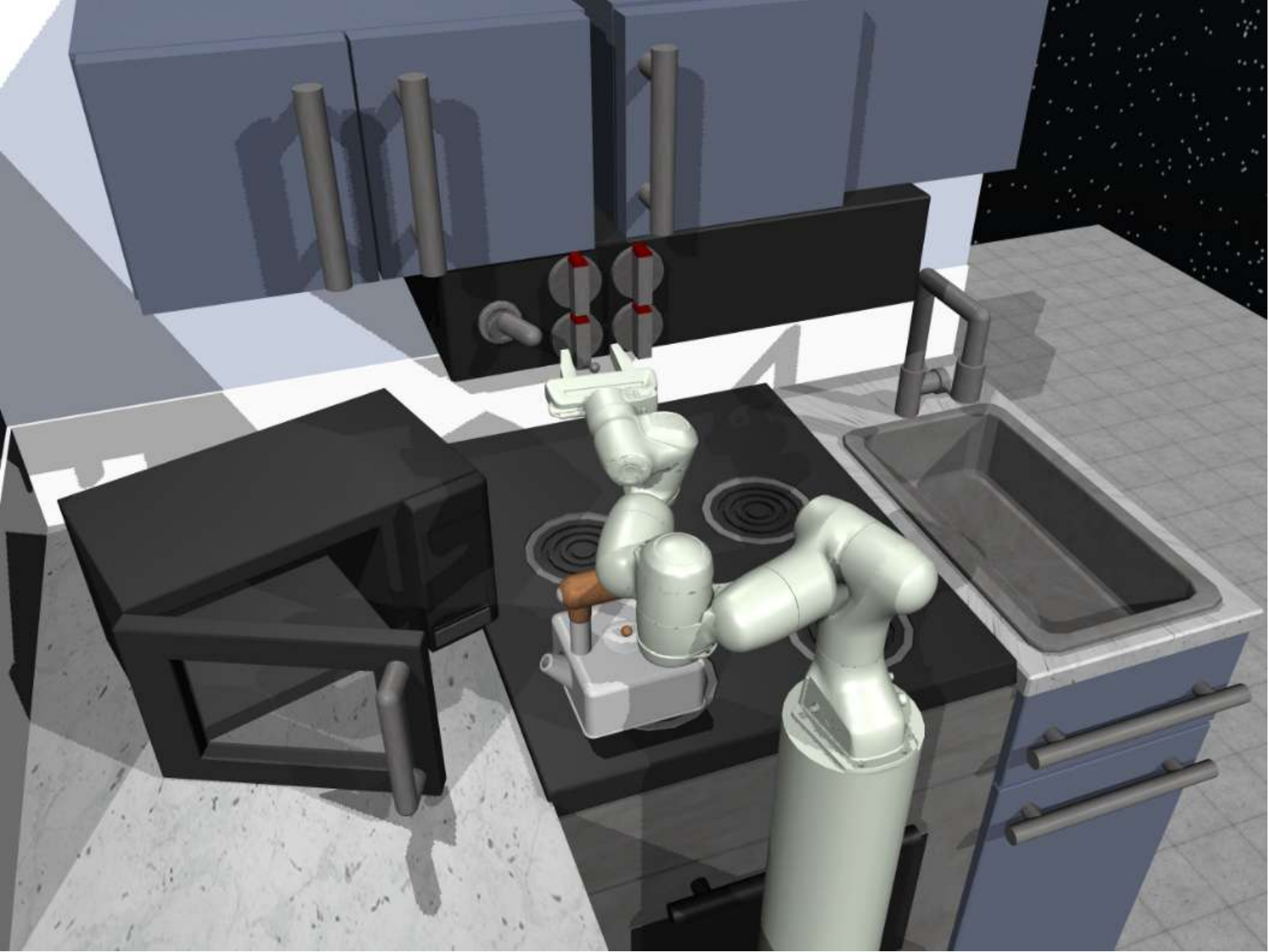}
\includegraphics[height=1.8cm,width=2.1cm]{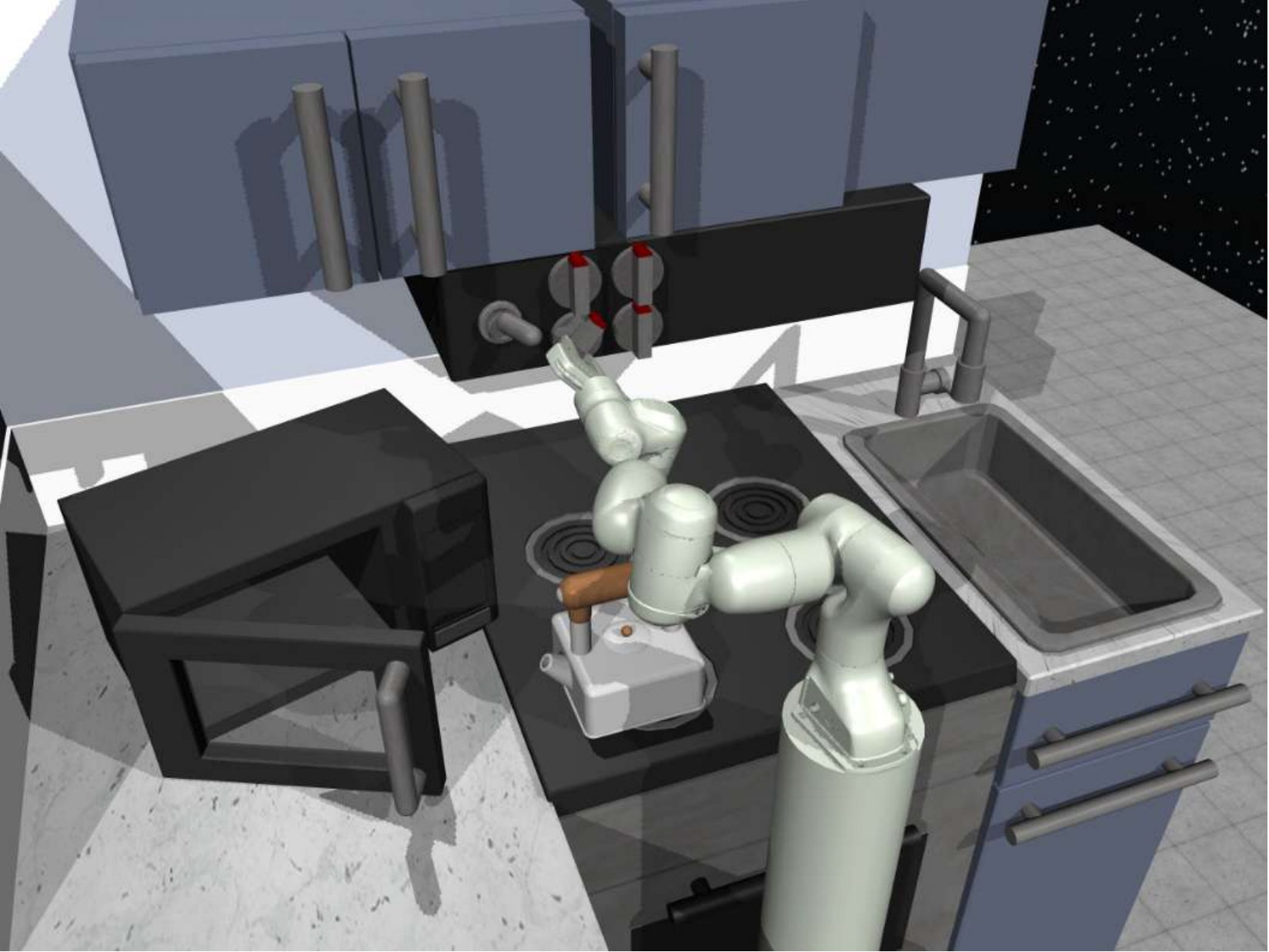}
\includegraphics[height=1.8cm,width=2.1cm]{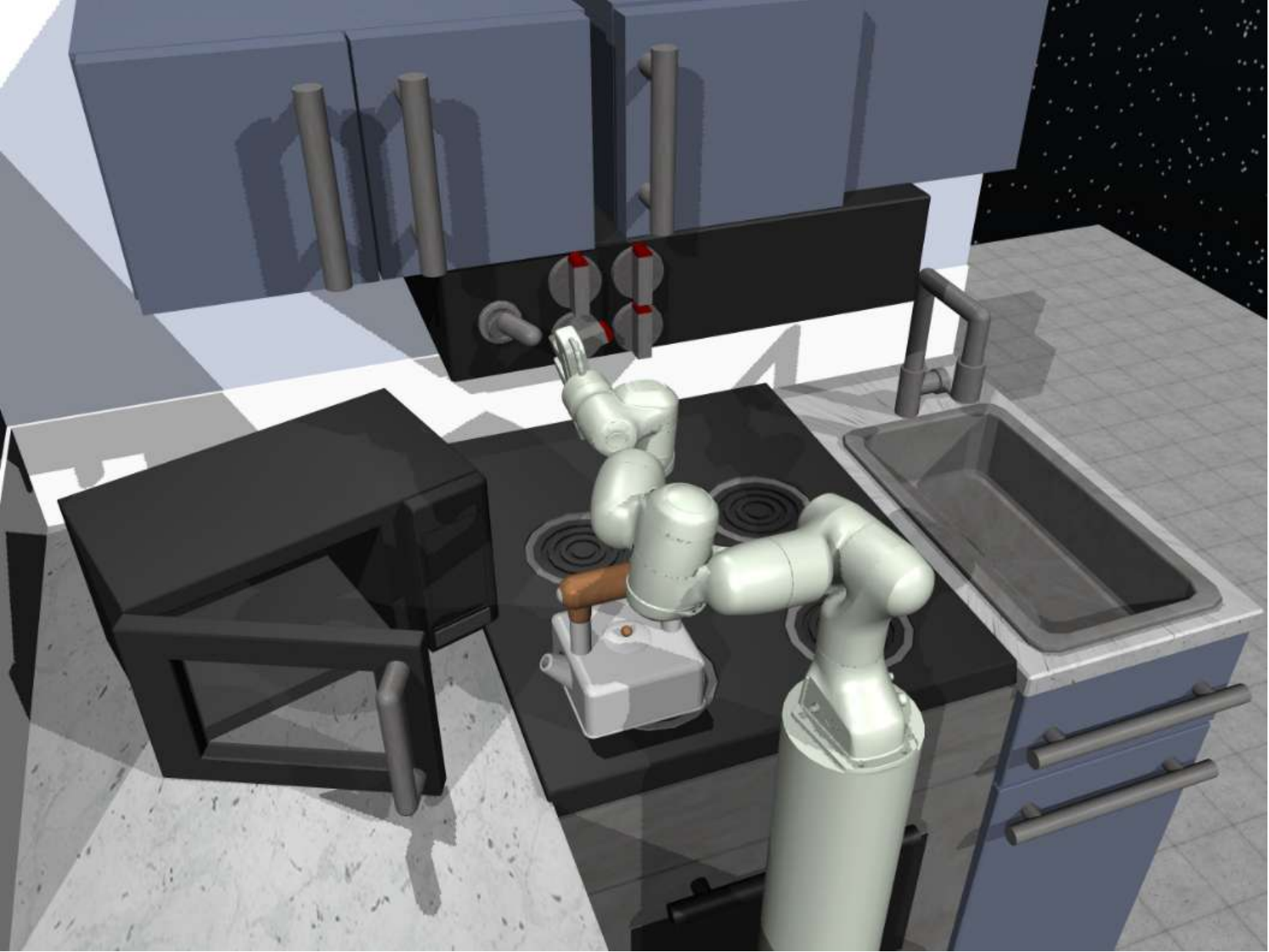}
\caption{The visualization is a successful attempt at performing kitchen navigation task}
\label{fig:kitchen_viz_success_1_ablation}
\end{figure}